\documentclass[journal]{IEEEtran}
\usepackage[utf8]{inputenc}
\usepackage{mathtools,lipsum,cuted}
\usepackage{amsmath,amsfonts}
\usepackage{enumitem}
\usepackage{algorithmic}
\usepackage{array}
\usepackage{subfigure}
\usepackage{stfloats}
\usepackage{url}
\usepackage{bm}
\usepackage{verbatim}
\usepackage{graphicx}
\usepackage{booktabs}
\usepackage{booktabs}
\usepackage{enumerate}
\renewcommand\eqref[1]{(\autoref{#1})}
\usepackage{mathtools,lipsum,cuted}
\usepackage{amsmath}     
\usepackage{amssymb}                       
\usepackage{mathrsfs}    
\usepackage{upgreek}
\usepackage{bm}     %bold for Greek symbols
\usepackage{authblk}
\usepackage{setspace}
\usepackage{graphicx}
\usepackage{amsfonts,amssymb}
\usepackage{caption}
\graphicspath{ {./figures/} }
\graphicspath{ {./Author/} }
\usepackage{amsmath}
\usepackage{cuted}
\usepackage{lineno}
\usepackage{colortbl}
\usepackage{bbding}
\usepackage{multirow}
\usepackage{makecell}
\usepackage{blindtext}
\usepackage{xcolor}
\usepackage{color}
\usepackage{setspace}
\usepackage{makecell}
\usepackage{threeparttable}
\usepackage{hyperref}
\usepackage[section]{placeins}
\usepackage{hhline}
\usepackage{threeparttable}
\usepackage{tikz}
\usepackage{diagbox}
\usepackage{algorithm,algorithmic}

\hypersetup{
	colorlinks=true,
	linkcolor=cyan,
	filecolor=blue,      
	urlcolor=black,
	citecolor=green,
}
\newcommand{\subsubsubsection}[1]{\paragraph{#1}\mbox{}}
\DeclareMathAlphabet\mathbfcal{OMS}{cmsy}{b}{n}
\def\BibTeX{{\rm B\kern-.05em{\sc i\kern-.025em b}\kern-.08em
    T\kern-.1667em\lower.7ex\hbox{E}\kern-.125emX}}
\usepackage{balance}

\begin{document}
\title{Language-Informed Hyperspectral Image Synthesis for Imbalanced-Small Sample Classification via Semi-Supervised Conditional Diffusion Model}
\author{Yimin Zhu, 
        Lincoln Linlin Xu, ~\IEEEmembership{Member,~IEEE}
\thanks{This work
was supported by the Natural Sciences and Engineering Research Council of
Canada (NSERC) under Grant RGPIN-2019-06744. (Corresponding author:
Lincoln Linlin Xu)}

\thanks{The authors are with the Department of Geomatics Engineering, University
of Calgary, Calgary, AB T2N 1N4, Canada (e-mail: yimin.zhu@ucalgary.ca;
lincoln.xu@ucalgary.ca)
}}

\markboth{Journal of \LaTeX\ Class Files,~Vol.~18, No.~9, September~2020}%
{How to Use the IEEEtran \LaTeX \ Templates}

\maketitle

\begin{abstract}
Data augmentation effectively addresses the imbalanced-small sample data  (ISSD) problem in hyperspectral image classification (HSIC). Although most methodologies extend features in the latent space, few leverage text-driven generation to create realistic and diverse samples for balancing limited annotated data. This paper proposes Txt2HSI-LDM(VAE), a novel language-informed hyperspectral image synthesis method to address the ISSD problem in HSIC. The proposed approach uses a denoising diffusion model, which iteratively removes Gaussian noise to generate hyperspectral samples conditioned on textual descriptions. Firstly, to address the high-dimensionality of hyperspectral data, a universal variational autoencoder (VAE) is designed to map the data into a low-dimensional latent space, which provides stable feature representation and significantly reduces the inference complexity of the diffusion model. Secondly, a semi-supervised diffusion model is designed to fully take advantage of unlabeled data. Random polygon spatial clipping (RPSC) and uncertainty estimation of latent feature (LF-UE) without parameters are used to simulate the varying degrees of mixing of training data. Thirdly, the VAE decodes HSI from latent space generated by the diffusion model with the language conditions as input, contributing to more realistic and diverse samples that match the textual descriptions. In our experiments, we fully evaluate synthetic samples' effectiveness from the aspects of statistical characteristics and data distribution in 2D-PCA space. Additionally, the key component, visual–linguistic cross-attention, is visualized on the pixel level to prove that our proposed model can capture the spatial layout and geometry property of the generated hyperspectral images. {Experiments carried out on two widely used HSI datasets demonstrate that the proposed Txt2HSI-LDM(VAE) outperforms the classical backbone models, state-of-the-art CNNs, semi-supervised methods, visual–linguistic feature fusion method, as well as Diffusion-model-based methods.}

\end{abstract}


\IEEEpeerreviewmaketitle

\section{Introduction}

\IEEEPARstart{H}yperspectral images (HSI) offer rich spatial-spectral information that can help to distinguish various spectrally-similar objects to support various environmental applications \cite{PAOLETTI2019279}. Among various hyperspectral processing tasks, HSI classification (HSIC) is a critical one \cite{Melgani2004, chenys2016, zhuxx2017, zhangmm2020, Liulq2022, yangjq2024}, which can support various key application, such as atmospheric environment monitoring \cite{Ghamisi2017, Alakian2024, Rouet-Leduc2024}, geologic mapping \cite{HAJAJ2024101218, Acosta2019}, and agriculture disaster response \cite{AKHYAR2024112067, ZHENG2021112636}. 

The extraction of spatial-spectral features plays an extremely important role in the field of HSIC. Traditionally, machine learning-based feature extraction methods and morphology-based approaches have been employed \cite{Melgani2004, Gualtieri2000}. Techniques such as Conditional Random Field (CRF) and Markov Random Field (MRF) are widely used to integrate spatial context information, enhancing HSIC performance \cite{Lifan2015}. However, these feature extractors are mostly based on feature engineering, limiting their adaptability to complex data characteristics required for improved HSIC. The data-driven feature learning and neural network (NN)-based approaches, represented by convolutional neural networks (CNNs) \cite{Hu2015, zhongzl2018, zhu2018, Qin2019, lianghb2022, zhuym2023} and transformer models \cite{Hexin2021, HongSpectralFormer2022, Sunle2022, zhongzlSSTN2022, Roy2023}, have significantly improved spatial-spectral feature extraction capabilities, thereby greatly boosting HSIC performance. 

To leverage unlabeled data for enhanced feature extraction, some self-supervised and unsupervised learning methods have been proposed. These methods employ an encoder-decoder network trained in a self-supervised manner for HSI reconstruction, such as autoencoder-based methods \cite{Luxq2017, Meish2019, yaochao2023, Linjy2023, hongspectralgpt2024}, and transfer learning using visual pre-training methods \cite{jiaolc2017, Hexin2021, wangdi2023}. Another group of hyperspectral processing methods is deep clustering, represented by graph-based approaches \cite{ding2022unsupervised, ding2025slcgc} through node definition and learning the strength of edge connections to cluster the spatial-spectral features. These models can leverage the label-agnostic multi-scale spatial-spectral structure information for enhanced HSIC.

Some unsupervised feature learning methods based on denoising diffusion probabilistic models (DDPMs) have emerged as powerful generative models with superior performance in image generation and restoration tasks \cite{Imagen2022, song2020denoising, ho2020denoising}. Due to the flexibility of DDPM's architecture and the accuracy of log-likelihood calculation, DDPM can implicitly capture both high-level and low-level visual concepts and model the complicated spatial-spectral relationship via explicit probability criteria \cite{song2019}. In remote sensing, change detection \cite{bandara2024ddpmcd}, classification \cite{chenning2023, chenjx2023, zhoujy2024, Tanxy2024, zhangjj2024} and super-resolution \cite{mengqy2023, Lisl2023, hekaiqi2024} tasks have fully explored the characteristic of DDPM. In HSIC, researchers in \cite{chenning2023,zhoujy2024} utilized a pre-trained diffusion model to extract the spatial-spectral diffusion features on denoising U-Net blocks for enhanced HSIC. 

%However, the approach of selecting timesteps manually is subjective, requires a lot of experimental verification, and cannot fully explore the complex spectral-spatial relationship, reducing the model generality. Another work \cite{zhoujy2024} proposed to use all of the timesteps, which undoubtedly led to redundancy.

%However, the aforementioned methods  require a large amount of training data to obtain refined land cover classification maps. 

However, these unsupervised DDPM-based approaches are limited to feature extractors, and cannot be used generate new data to better address limited training sample problems, especially for small classes that have very limited training samples. The long-tailed and sparse distribution of training data tends to bias the model toward the majority classes \cite{Johnson2019, wangxue2020, liuwei2021}. In some specific classification scenes, minority classes are more important from the application perspective, and thereby require a higher classification accuracy than the majority classes which might be less important \cite{xibb2023}. This imbalanced-small sample data (ISSD) problem also leads to a lack of sample diversity, limiting the generalization ability of the algorithm. Hence, it is critical to design new DDPM approaches that can efficiently generate new data and expand training data to deal with the ISSD HSIC problem.

Some data generation approaches have been successfully introduced to address the issue of ISSD in HSIC \cite{wangxy2020, han2024latent, shao2024diffult}. Among these, conditional generative adversarial networks (CGANs) are one of the most representative algorithms, which have the capability of generating realistic samples within the class-conditional distribution of real-world data \cite{fajardo2021oversampling, douzas2018effective}. CGANs employ data augmentation and minority-class oversampling strategies to mitigate ISSD challenges In HSIC \cite{dam2020mixture, roy2021generative}. However, a notable limitation of this framework lies in the unstable training mechanism of GANs, which may lead to model collapse, and cause the quality degradation of generated samples \cite{lala2018evaluation}. Although recent studies have attempted to integrate the minority-class generation and classifier optimization into a unified architecture, the joint optimization of the compound loss function remains a complex task. As \cite{xibb2023} reported, due to the high dimensionality of HSI, the class-conditional information might get overshadowed in this approach. 

Recently, conditional diffusion models utilizing variational autoencoders (VAE), known as latent diffusion models (LDMs), have successfully incorporated textual descriptions into the data generation pipeline, significantly enhancing generation performance. Some studies examined the impact of imbalanced class distribution within several long-tailed datasets on diffusion-based generative frameworks \cite{qin2023class, lomurno2024stable}, demonstrating their unique advantages in addressing minority classes. 

Hence, motivated by the success of text-guided LDMs in data simulation, especially for addressing minority classes, we propose a new language-informed HSIC framework named Txt2HSI-LDM(VAE), which integrates the LDM architecture with linguistic guidance to address the imbalance-small sample data (ISSD) problem in HSIC. Unlike CNNs and other generative models, Txt2HSI-LDM(VAE) embeds language information directly into the framework, ensuring the generated samples align well with both the underlying statistical distribution and and the textual descriptions, and enabling the visual fidelity of generated samples while mitigating the cost of extensive manual annotation.

In more detail, the main contributions of this article are summarized below.

1) \textbf{New Hyperspectral Image Synthesis Framework:} a new hyperspectral image synthesis framework based on a language-informed conditional diffusion model Txt2HSI-LDM(VAE) is proposed to solve the ISSD problem in HSIC. Txt2HSI-LDM(VAE) is a three-stage framework. The first stage uses a VAE to obtain stable representations in low-dimensional latent space, which accelerates diffusion model training and reduces the model's parameters. The second stage utilizes LDM to generate realistic samples under language guidance. The Txt2HSI-LDM(VAE) framework can better address the minority class issue by generating more balanced data. The third stage combines the synthetic data and real labeled data for enhanced ISSD HSIC.

2) \textbf{Reduction of Manual Annotations:} The proposed approach can reduce the burden of making manual annotations by using the fine-grained text description based on the distribution, color, shape, and adjacency relationship of the HSI classes. Moreover, even coarse-grained textual descriptions can improve classification performance. In addition, the language guidance is useful for generating new hyperspectral data in a language-guided manner for different purposes.

3) \textbf{Improved Model Generalization Capability:} The proposed approach can improve the model's generalization capability, because, at the second stage, we design a semi-supervised LDM approach to fully leverage the unlabeled data. Additionally, to better capture the underlying statistical distribution, random polygon spatial clipping (RPSC) and latent feature uncertainty estimation (LF-UE) are used to simulate the spatial-spectral heterogeneity and spectral mixing phenomena in real HSI data.

4) \textbf{Enhanced Classification Model:} A HSI classification approach is used that can make full use of both real and simulated data to address the ISSD problem in HSIC.

The remainder of the article is organized as follows. \autoref{Related work} briefly introduces the related works. \autoref{Proposed Txt2HSI-LDM(VAE)} describes our proposed method in detail. \autoref{Results} shows the experimental results. \autoref{conclusion} draws the conclusions and discussions.

\section{Related work} \label{Related work}

\subsection{Imbalanced-Small Sample Data (ISSD) HSIC}

Several prior works, including data augmentation, resampling (oversampling and undersampling) \cite{lijiaojiao2018, Lvqinzhe2021}, deep transfer learning \cite{Hexin2021, Scheibenreif2023}, and deep few-shot learning \cite{zhangyuxiang2024}, have been proposed to address the ISSD HSIC problem. For instance, random oversampling (ROS) and random undersampling (RUS) simply replicate the training samples which leads to data redundancy \cite{REN2012144}. Synthetic minority oversampling technique (SMOTE) may contribute to better classification performance on conventional classifiers, but this method uses a simple linear combination of one random sample and its nearest neighbors to generate synthetic minority samples, which fails to capture the complex manifold structure and is hard to satisfy the nonlinear spatial-spectral relationship in the HSI data \cite{SMOTE2002}.

Latent variable models, especially represented by the deep generative models (DGMs), introduce uncertainty into the low-dimensional discriminative latent space during the reconstruction process, possessing the ability to stimulate the complex heterogeneity of HSI data and generate samples for the minority class. Particularly, GANs replace the complexity of resampling by searching for a Nash equilibrium of the generator and discriminator through adversarial training. Specifically, class-informed mixture GANs are developed to ensure the generated samples align with the actual data distribution. For example, Dam et al. proposed a parallel mixture generator spectral 1D-GAN \cite{dam2020mixture} to generate class-dependent samples by fully exploiting the spectral feature. Auxiliary classifiers significantly improved the classification performance of imbalanced datasets. Zhu et al. proposed a 3D-GAN \cite{zhu2018} that took both spatial and PCA-reduced spectral features into account to handle 3D patch samples. The discriminator not only distinguished the real data from the synthetic fake data but also leveraged the softmax classifier to produce the classification maps.

The common deep transfer learning methods mainly use some pre-trained models in the natural images field, such as VGG16 and ResNet50 with frozen parameters, to fine-tune the downstream tasks in the remote sensing community. Thanks to these pre-trained models, they can obtain promising results with a small sample size and lower computational burden. Jiao et al. proposed a deep multiscale feature extraction algorithm, which fully took advantage of pre-trained VGG-16, to exploit the spatial structure information of hyperspectral images by performing a full convolution operation, and then realized adaptive spatial-spectral information fusion by employing weighted strategy \cite{jiaolc2017}. Wang et al. first built three-spectral band hyperspectral datasets through random band selection (RBS) procedure, letting VGG-16 learn the feature of homogeneous areas with distinguished semantic and geometric properties \cite{wangdi2023}. However, extracting features of HSI with models that are pre-trained on natural RGB images requires generalizing to hundred of bands from triple-band datasets, which is a chanllenging task.

\subsection{Image Synthesis for HSIC}
To estimate the data distribution for more realistic image synthesis and expand the small training data, scholars have developed several methods, which have achieved promising results. The commonly used algorithms include the mixture model (Gaussian Mixture Model, GMM), generative model (GAN, VAE), and DDPM.

Deep learning-based methods, with GAN and VAE being typical representatives, generate new samples of a given distribution. Specifically, this is achieved by training a generator to map random noise from the latent space to the data distribution. Instead of inferring the global distribution, scholars employed conditional generative models, which are designed to condition the output of the generator based on the hyperspectral classes. For instance, in \cite{Audebert2018}, Audebert et al. used conditional GAN to generate an arbitrarily large number of hyperspectral samples that matched the distribution of any dataset. Considering spectral mixing, this method can synthesize any combination of classes by interpolating between vectors in the latent space. Due to the high-dimensional HSI data with complex noise, VAE is more suitable for processing HSI by introducing the Gaussian noise into the encoded results in latent space. In \cite{xibb2023}, Xi et al. proposed a DGSSC method, which can achieve minority-class data augmentation in the latent variable space to address the problem of imbalanced data. However, the condition was only based on one-hot encoding of hyperspectral classes. Considering the complex situations in the real-world application and the semantic context, it is likely that the one-hot-like encoding in probabilistic form, such as pixel abundance, instead of binary vector, will work better. 

Very recently, DDPMs have demonstrated superior performance in image inpainting and synthesis compared to GANs and VAEs. For instance, in \cite{zhanglei2023}, Zhang et al. proposed R2H-CCD, an algorithm for HSI generation that leveraged the conditional DDPM to produce high-fidelity HSI from RGB image through cascaded generation. Liu et al. proposed a diverse hyperspectral remote sensing image generation method based on latent diffusion models (HyperLDM) \cite{liuliqin2023}, employing a conditional vector quantized generative adversarial network (VQGAN) to compact high-dimensional information for accelerating the diffusion process. Additionally, HyperLDM incorporated abundance maps derived from VQGAN to condition the HSI generation via LDM. By doing so, the model effectively accounted for the spectral mixing of multiple materials within HSI pixels. However, a critical limitation of HyperLDM is its reliance on an accurately calibrated spectral library during the transformation of abundance maps into HSI using the linear mixture model (LMM).

However, the above methods, whether conditional GANs conditional VAEs, or latent conditional DDPMs, still require substantial quantities of paired remote sensing image observations (e.g., RGB image and corresponding HSI). This poses a significant challenge in practical applications, due to the limited availability of such paired datasets arising from satellite revisit period variations and heterogeneous sensor specifications across different observation platforms.

\subsection{Image–Text Pairing information in HSIC}
Hyperspectral image analysis demands intensive manual labeling, while, language representations are inherently more accessible straightforward for humans to express and acquire. Recent advancements in AI methods, such as contrastive language image pre-training (CLIP) \cite{CLIP2021} has demonstrated the capability to model the bidirectional relationships between these two modalities (image and text). Moreover, these methods present exceptional transferability across diverse downstream visual and language tasks, making them promising candidates for addressing the data scarcity challenges in HSI-related applications \cite{wahabzada2016plant, Huanglb2024}.

Zhang et al. proposed a domain generalization network (LDGNet)  \cite{zhang-LDGNet2023} for HSIC. This method learned visual-linguistic alignment features from source datasets and transferred cross-domain invariant knowledge to target domains. Similarly, Dang et al. \cite{Dangyy2024} developed LIVEnet for cross-scene HSI classification, which enhanced the information exchange between linguistic and visual features while exploiting interclass correlation prevalent in imbalanced sample scenarios. Moreover, Cao et al. designed a three-aspect text description (color, label, shape) to extract linguistic prior information, enabling shared spatial and spectral features in a unified semantic space \cite{Caomx2024}. Given these successes, however, the text-guided approaches have not been fully explored for addressing the ISSD problem in HSIC. 

%However, a critical limitation of these aforementioned methods lies in their inability to quantitatively evaluate pixel-level text-image correlations. In other words, these methods fail to assess the relative importance of different textual components in a specific text description, which is essential for understanding the contribution of individual linguistic features to the classification process.

\begin{figure*}[h]
    \centering
    \includegraphics[width=0.99\textwidth]{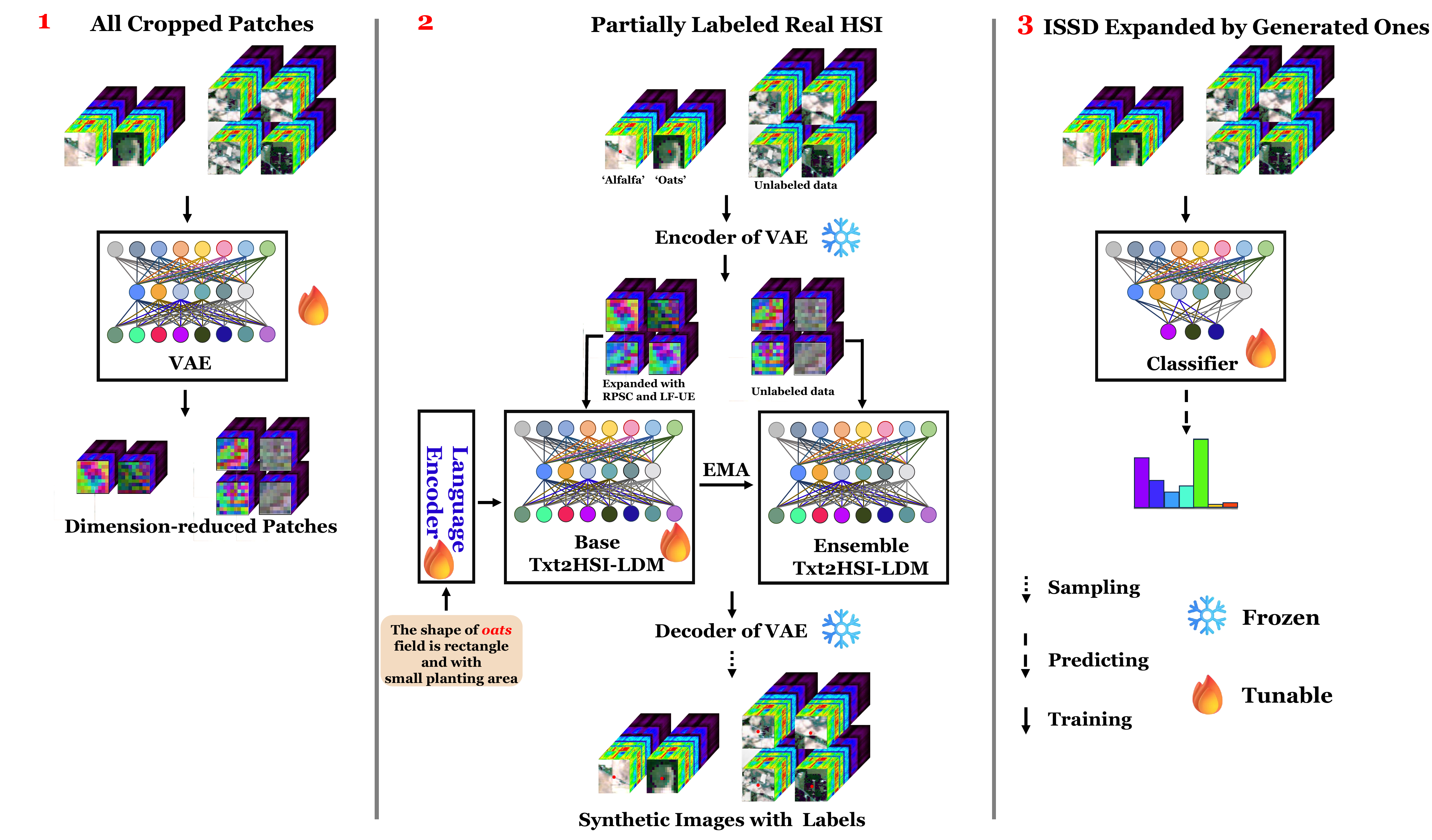}
    \caption{{An overview of Txt2HSI-LDM(VAE). First, a VAE is trained on all the cropped patch HSI data, and the dimension-reduced data is used for the second stage. Second, a language-informed conditional diffusion model, Txt2HSI-LDM is trained on limited labeled data and unlabeled data in a semi-supervised way to generate synthetic images given random language descriptions. Finally, the classifier \cite{zhuym2023} is trained or fine-tuned on ISSD data expanded by generated images with labels. EMA means exponential moving average to transfer parameters from the base model to the ensemble model. The language encoder uses the pre-trained parameter of the CLIP model named 'ViT-B-32.pt' \cite{CLIP2021}, and we fine-tune it together with the diffusion model.}}
    \label{overview}
\end{figure*}

\subsection{Latent Conditional Diffusion Model} \label{LDM_section}

Diffusion models \cite{ho2020denoising} are probabilistic models designed to learn data distributions \(p(x)\) by gradual denoising a normally distributed variable, which corresponds to learning the reverse process of a fixed Markov Chain of total length \(T\).

However, for high-dimensional distributions, diffusion model training (see \autoref{ddpm_training}) and sampling (see \autoref{ddpm_sample}) require massive computational resources.
\begin{align}
    \mathcal{L}_{DM} = \mathbb{E}_{x, \epsilon \sim \mathcal{N}(0,1), t} [\lVert \epsilon - \epsilon_{\theta}(x_t, t)  \rVert _{2} ^{2}]
\label{ddpm_training}
\end{align}
\begin{align}
    q(x_{t-1} | x_{t}, t) = \mathcal{N}(x_{t-1}; \frac{1}{\sqrt{\alpha_{t}}}(x_{t} - \frac{\beta_{t}}{\sqrt{1-\bar{\alpha}_{t}}} \epsilon_{\theta} (x_{t}, t)), \beta_{t} \mathbf{I})
\label{ddpm_sample}
\end{align}
where, \(\alpha_{t} \coloneqq  1-\beta_{t}\), \(\bar{\alpha}_{t} \coloneqq  \prod \limits_{s=1}^t \alpha_{s}\), \(\epsilon \sim \mathcal{N}(0, \mathbf{I})\), \(\epsilon_{\theta}\) represents the diffusion model with parameter \(\theta\).

To enhance the efficiency of the training and sampling process, Rombach et al., the authors of Stable Diffusion \cite{stablediffusion2022} proposed latent diffusion models (LDMs), which performed the diffusion in the latent space of a pre-trained VAE. Specifically, given a frozen encoder \(\mathcal{E}\): \(\mathbb{R}^{n} \rightarrow \mathbb{R}^{k}\) and decoder \(\mathcal{D}\): \( \mathbb{R}^{k} \rightarrow \mathbb{R}^{n} \), train a diffusion model of representations \(z = \mathcal{E}(x_0) \), where \(x_0\) represents the original data. The noisy samples can be created by:
\begin{align}
    q(z_{t} | \mathcal{E}(x_0), t) = \mathcal{N} (z_{t}; \sqrt{\bar \alpha_{t}} \mathcal{E}(x_0), (1-\bar \alpha_{t}) \mathbf{I})
\end{align}
The corresponding objective can be written as:
\begin{align}
    \mathcal{L}_{LDM} = \mathbb{E}_{z \sim \mathcal{E}(x), \epsilon \sim \mathcal{N}(0,1), t} [\lVert \epsilon - \epsilon_{\theta}(z_{t}, t)  \rVert _{2} ^{2}]
\end{align}

At inference time, new images can then be generated by sampling a clean representation \(z\) from the diffusion model and subsequently decoding it to an image with the learned decoder \(x = \mathcal{D}(z)\)

In principle, diffusion models have the capability to formulate conditional distributions in the form of \(p(z|c)\). This can be implemented with a conditional denoising autoencoder \(\epsilon_{\theta}(z_{t}, t, c)\), which paves the way to control the synthesis process through inputs \(c\), such as text \cite{khanna2024diffusionsat}, semantic maps \cite{Tang2024}, and other image translation tasks \cite{Sat2Scene2024}.

\section{Proposed Txt2HSI-LDM(VAE)} \label{Proposed Txt2HSI-LDM(VAE)}

\subsection{Overview of proposed Txt2HSI-LDM(VAE)}

{In this section, we introduce the Txt2HSI-LDM(VAE) framework, which integrates the strengths of VAE (for dimension reduction) and language-informed conditional diffusion models, Txt2HSI-LDM for short, to address the ISSD problem in HSIC by generating realistic hyperspectral image synthesis. The proposed method is specifically designed to enhance classification performance, particularly when the labeled dataset is imbalanced and small. As shown in \autoref{overview}, the framework consists of three main phases: (1) VAE pretraining, (2) Txt2HSI-LDM training, and (3) Txt2HSI-LDM inference and classification. It is worth noting that, in the second phase, to expand and enrich the training data, Txt2HSI-LDM is trained in a semi-supervised fashion using abundant unlabeled data. In order to further enhance the diversity of the sample, two methods without parameters, random polygon spatial clipping (RPSC) and uncertainty estimation of latent feature (LF-UE), are used to stimulate the diversity of hyperspectral images. This architecture enables us to harness the information compression capabilities of VAE and the generative potential of conditional diffusion models to address ISSD challenges.}

\subsection{Latent Space Representation via VAE}
\begin{figure}[h]
    \centering
    \includegraphics[width=0.49\textwidth]{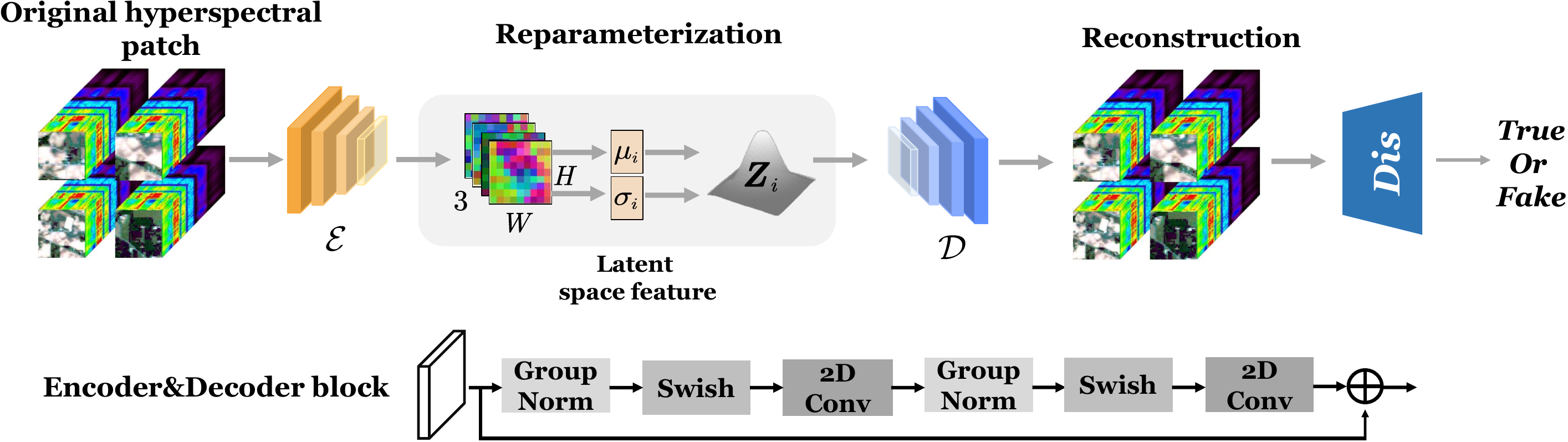}
    \caption{{Illustration of the variational autoencoder (VAE), which consists of an encoder network, a decoder network, a reparameterization part, and a discriminator network. The detailed architectures of the encoder and decoder blocks are also depicted.}}
    \label{vae_architecture}
\end{figure}

As depicted in \autoref{LDM_section}, the first stage of the proposed model Txt2HSI-LDM(VAE) is implemented with VAE to obtain compressed and stable feature representations in latent space. The main components include four parts: an encoder network, a reparameterization process, a decoder network, and a discriminator network (see \autoref{vae_architecture} for detailed architectures of the encoder and decoder networks.). Given a hyperspectral instance \( \mathcal{H} \in \mathbb{R}^{C \times H \times W}\), where \(H\) and \(W\) are the height and width of hyperspectral patch, \(C\) is the number of spectral band. The encoder of VAE projects the \( \mathcal{H} \) into \( z \in \mathbb{R}^{4 \times H \times W}\), which is modeled by \( \boldsymbol{q}_{\mathcal{E}_{\theta}} (\boldsymbol{Z}|\boldsymbol{\mathcal{H}})\). The reconstructed hyperspectral instance \(\mathcal{H}^{'}\) can be obtained by feeding the reparameterized \(\boldsymbol{Z}\) to the decoder part, modeled by \(\boldsymbol{p}_{\mathcal{D}_{\theta}} (\boldsymbol{\mathcal{H}} | \boldsymbol{Z})\).
Reparameterization is a process that samples a Gaussian distribution with mean \(\mu(Z)\) and \(\Sigma(Z)\) variance of latent feature \(Z\).

We optimize the VAE under the framework of GANs to make the reconstruction of HSI more realistic. The discriminator is the same as in our previous work \cite{zhu2025sDiffCRN}. The VAE is optimized by minimizing the following objective function:
\begin{align}
    \begin{aligned}
        &\underset{\theta_{\mathcal{E}}, \theta_{\mathcal{D}}}{\min} \underset{\theta_{\mathcal{G}}}{\max}  (\mathcal{L}_{vae} + \mathcal{L}_{adv}) \\ 
        \mathcal{L}_{vae} &= \frac{1}{HW} \underset{i=1}{\overset{H}{\varSigma}} \underset{j=1}{\overset{W}{\varSigma}} \lVert \mathcal{H}'_{ij} - \mathcal{H}_{ij} \rVert_{2}^{2} 
        \\ &+ \lambda_{KL}\frac{1}{N} \underset{i=1}{\overset{HW}{\varSigma}}  D_{KL} (q_{\mathcal{E}_{z} }(z_{i}|\mathcal{H}_{i}) || p(z))\\
        \mathcal{L}_{adv} &= \mathbb{E}_{\mathcal{H} \sim p(\mathcal{H})} \log \mathbf{Dis}(\mathcal{H}) \\
        &+ \mathbb{E}_{\mathcal{H} \sim p(\mathcal{H})} \log [1-\mathbf{Dis}(\mathcal{H}')]
    \end{aligned}
\end{align}
where \(\theta_{\mathcal{E}}\), \(\theta_{\mathcal{D}}\), and \(\theta_{\mathcal{G}}\) denote parameters in encoder \(\mathcal{E}\), decoder \(\mathcal{D}\), and discriminator \(\mathbf{Dis}\), respectively. \(q_{\mathcal{E}_z}({z_{i}} | \mathcal{H}_{i})\) is the variational distribution in the VAE.

\begin{figure}[h]
    \centering
    \includegraphics[width=0.49\textwidth]{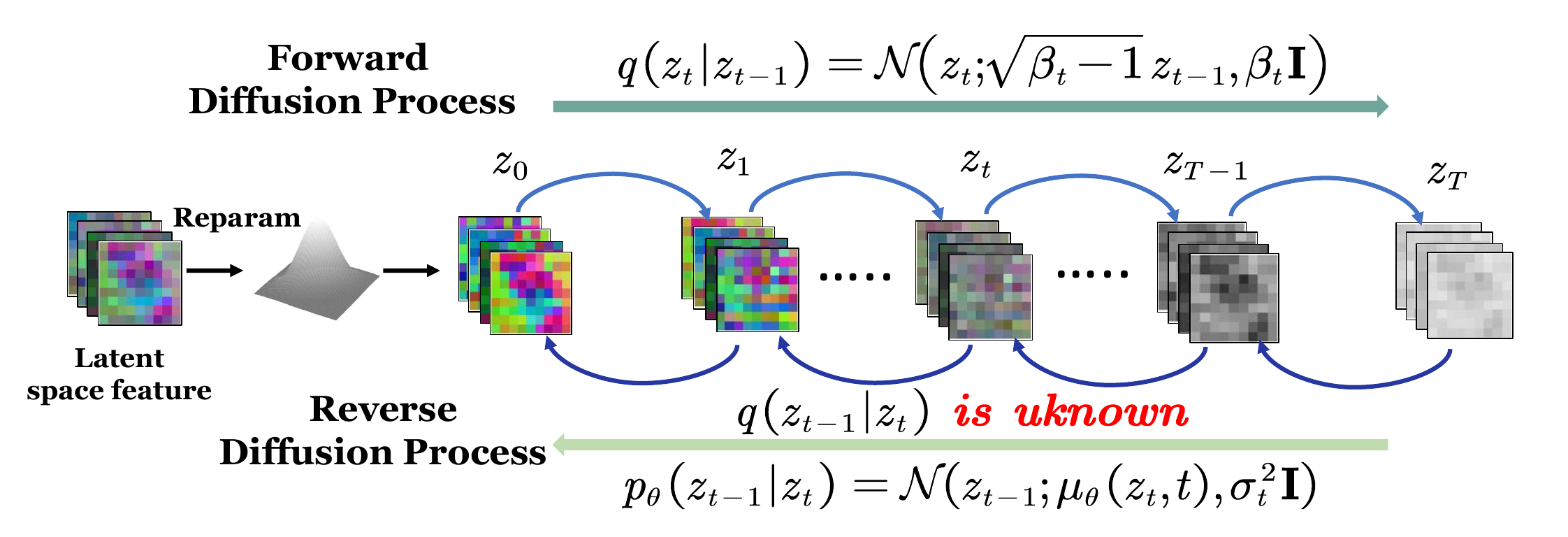}
    \caption{{Latent diffusion forward and backward process.
$q(z_{\boldsymbol{t}}|z_{\boldsymbol{t-1}})$, $p_{\theta}(z_{\boldsymbol{t-1}}|z_{\boldsymbol{t}})$ represent the noise-adding forward process and denoising backward process, respectively. The essential question is to estimate the conditional probability $q(z_{\boldsymbol{t-1}}|z_{\boldsymbol{t}})$. \(z_T\) is nearly the pure Gaussian noise.}}
    \label{diffusion_process}
\end{figure}

% \begin{figure*}[h]
%     \centering
%     \includegraphics[width=0.99\textwidth]{figures/framework_overview.pdf}
%     \caption{Overview of the proposed method Txt2HSI-LDM(VAE) . The encoder \(\mathcal{E}\) of VAE maps the hyperspectral patch data into a latent space where the diffusion process operates. The VAE is universal and only trained once before training the diffusion model. The diffusion model is conditioned on the fine-grained language \(c_{l}\) and timestep \(c_{t}\). The generation process starts with Gaussian noise and iteratively denoises it to obtain the latent representation \(z\). The decoder \(\mathcal{D}\) then takes the latent code \(z\) as input and produces generated hyperspectral image instances, which match the description of the text. The language encoder uses the pre-trained parameter of the CLIP model named 'ViT-B-32.pt' \cite{CLIP2021}, and we fine-tune it together with the diffusion model. Details of VAE are shown in \autoref{vae_architecture}.}
%     \label{framework_overview}
% \end{figure*}
\subsection{Preliminary of Diffusion Model}

The diffusion model is a latent variables model that contains a forward diffusion process and a reverse diffusion process. Both forward and reverse processes can be defined as a parameterized Markov chain, and the reverse process can estimate the added noise in the forward diffusion process for generation \cite{ho2020denoising}.

\subsubsection{\textbf{Forward Diffusion Process}}
We set the latent space feature learned by the encoder \(\mathcal{E}\) of VAE before adding the Gaussian noise as \(z_{0}\). The forward process is inspired by non-equilibrium thermodynamics which can be viewed as a Markov chain that the noise is gradually added to \(z_{0}\) to produce latent variables \(z_{1}\) through \(z_{T}\) in \(T\) timesteps with fixed
linear variance schedule \(\beta_{t}\). At timestep t, the noisy feature \(z_{t}\) can be represented as follows:
\begin{align}
    &q(z_{1}, z_{2}, ..., z_{T}|z_{0}) = \underset{t=1}{\overset{T}{\Pi}} q(z_{t}|z_{t-1}) \label{EQ6}\\
    &q(z_{t}|z_{t-1}) = \mathcal{N}(z_{t}; \sqrt{\beta_{t} - 1} z_{t-1}, \beta_{t} \mathbf{I}) \label{EQ7}
\end{align}

Through the reparameterization trick,  \(z_{t}\) can be sampled directly at an arbitrary time step \(t\) in closed form, the marginal distribution of intermediate latent variables \(z_{t}\) given \(z_{0}\) can be derived as:
\begin{align}
    &q(z_{t} | z_{0}) = \mathcal{N} (z_{t}; \sqrt{\bar{\alpha}_{t}} z_{0}, (1 - \bar{\alpha}_{t}) \mathbf{I})\\
    &z_{t} = \sqrt{\bar{\alpha}_{t}}z_{0} + (1 - \bar{\alpha}_{t}) \epsilon \label{EQ9}
\end{align}

where \(\alpha_{t} = 1-\beta_{t}\), \(\bar{\alpha}_{t} = \underset{s=1}{\overset{t}{\Pi}} a_{s}\), \(\epsilon \sim \mathcal{N}(0, \mathbf{I})\). In \autoref{EQ7}, the magnitude of the added Gaussian noise decreases as the value of \(\alpha_{t}\) increases, which also means that the informative of \(z_{0}\) decreases as the time step \(t\) increases in \autoref{EQ9}. {When \(t=T\), \(z_T\) will become pure Gaussian noise.}

\subsubsection{\textbf{Reverse Diffusion Process}}
In the reverse process, \autoref{diffusion_process}, the estimation of \(q(z_{t-1}|z_{t})\) is intractable and depends on the whole data distribution, and a denoising neural network \(\epsilon _{\theta}\) is used to estimate it, where \(\theta\) represents the network parameters, and then, {we generate the real sample from Gaussian sample \(z_{T} \sim \mathcal{N}(0,\mathbf{I})\) with conditional text constraint \(c\), motivated by classifier-free diffusion guidance \cite{ho2022classifier}}.  According to the property of Markov chain, \autoref{EQ6}, and \autoref{EQ7}, the reverse distribution \(p_{\theta}\) can be formulated as follows:
\begin{align}
    p_{\theta}(z_{0:T}|c) &= p(z_{T}) \underset{t=1}{\overset{T}{\Pi}} p_{\theta} (z_{t-1} | z_{t},c) \\
    p_{\theta}(z_{t-1} | z_{t}, c) &= \mathcal{N} (z_{t-1}; \mu_{\theta}(z_{t},t, c),  \Sigma_{\theta} (z_{t},t, c))
\end{align}
We set \(\Sigma_{\theta} (z_{t}, t, c) = \sigma_{t}^{2} \mathbf{I} = \frac{1 - \bar{\alpha}_{t-1}}{1 - \bar{\alpha}_{t}} \beta_{t}\mathbf{I}\) to untrained time dependent constants. For \(\mu_{\theta}(z_{t}, t, c)\), the conditional distribution of \(p_{\theta}(z_{t-1}|z_{t},c)\) can be reparameterized as follows:
\begin{align}
    \mu_{\theta}(z_t,t, c) = \frac{1} {\sqrt{\alpha_{t}}} (z_{t} - \frac{\beta_{t}}{\sqrt{1 - \bar{\alpha}_{t}}} \epsilon _{\theta}(z_{t},t, c))
\end{align}
Finally, the objective function of the diffusion model can be constructed by \autoref{EQ13}, and the detailed solutions and derivation process can be referred to \cite{ho2020denoising}:
\begin{align}
    \mathcal{L}_{DM}(\theta) = \mathbb{E}_{z_0, c, \epsilon} [ \lVert \epsilon - \epsilon_{\theta}(\sqrt{\bar {\alpha}_{t}} z_{0} + \sqrt{1 - \bar{\alpha}_{t}} \epsilon, t, c)\rVert _{2}^{2}] \label{EQ13}
\end{align}
\subsubsection{\textbf{Sampling Process}}
In the inference process, which is also known as the sampling process, a new latent feature \(z_0\) can be generated from either Gaussian noise or a noisy image \(z_{t}\) by iteratively sampling \(z_{t-1}\) until \(t=1\) according to the \autoref{EQ14}. We adopt adopted denoising diffusion implicit models (DDIM) \cite{song2021DDIM} to accelerate the sampling process.
\begin{align}
    \begin{aligned}
        z_{t-1} = \sqrt{\alpha_{t-1}} (\frac{z_{t} -                            \sqrt{1-\alpha_t} \bar{\epsilon}_{\theta}}{\sqrt{\alpha_t}})
                    + \sqrt{1 - \alpha_{t-1} - \sigma_{t}^{2}}                         \bar{\epsilon}_{\theta} + \sigma_{t}                            \epsilon_{t}
    \end{aligned} \label{EQ14}
\end{align}
where \(\epsilon_{t} \sim \mathcal{N}(0, \mathbf{I})\). Following classifier-free guidance diffusion \cite{ho2022classifierfreediffusionguidance}, we set \(\bar{\epsilon}_{\theta} = (1+\omega) \epsilon_{\theta}(z_t,t,c) - \omega \epsilon_{\theta}(z_t, t, c=\oslash)\). \(c = \oslash\) is done by randomly dropping out \(c\) during training and replacing it with a learned “null” embedding \(\oslash\). Generally, training with classifier-free guidance requires two models: an unconditional generation model and a conditional generation model. However, these two models can be unified into a single model by probabilistically omitting the language condition during training. During inference, the final result can be achieved by linear extrapolation between the conditional and unconditional generations. This allows for adjustment of the generation effect to balance the fidelity and diversity of the generated samples by changing the guidance coefficient \(\omega\).

\subsection{Conditional Denoising Diffusion in the Latent Space} \label{text_despription}
The hyperspectral data usually have low spatial resolution, follow a long-tail distribution, and have high-dimension properties. To speed up the generation process of the diffusion model and reduce the number of parameters, we perform diffusion in the latent space to cope with the high dimensionality and uneven range of the hyperspectral data. Simultaneously, if we use the whole-image level hyperspectral data as the input of the VAE encoder, predictably, the step-by-step spatial downsampling operations will lose the detail of images, which leads to the negative effects on the generated samples, as well as the classification results. So, we crop 
hperspectral data into patches and encode these into latent space using a universal VAE, which only reduces the spectral dimension and keeps the spatial size.  Then, we train a conditional latent diffusion model in this space to generate latent features with conditional fine-grained text descriptions as input. Finally, we decode the latent features into HSI using the VAE decoder. An overview of our proposed model, Txt2HSI-LDM(VAE), is shown in \autoref{overview}. 
% \subsection{Framework of of Txt2HSI-LDM(VAE)}

\begin{figure}[]
    \centering
    \includegraphics[width=0.49\textwidth]{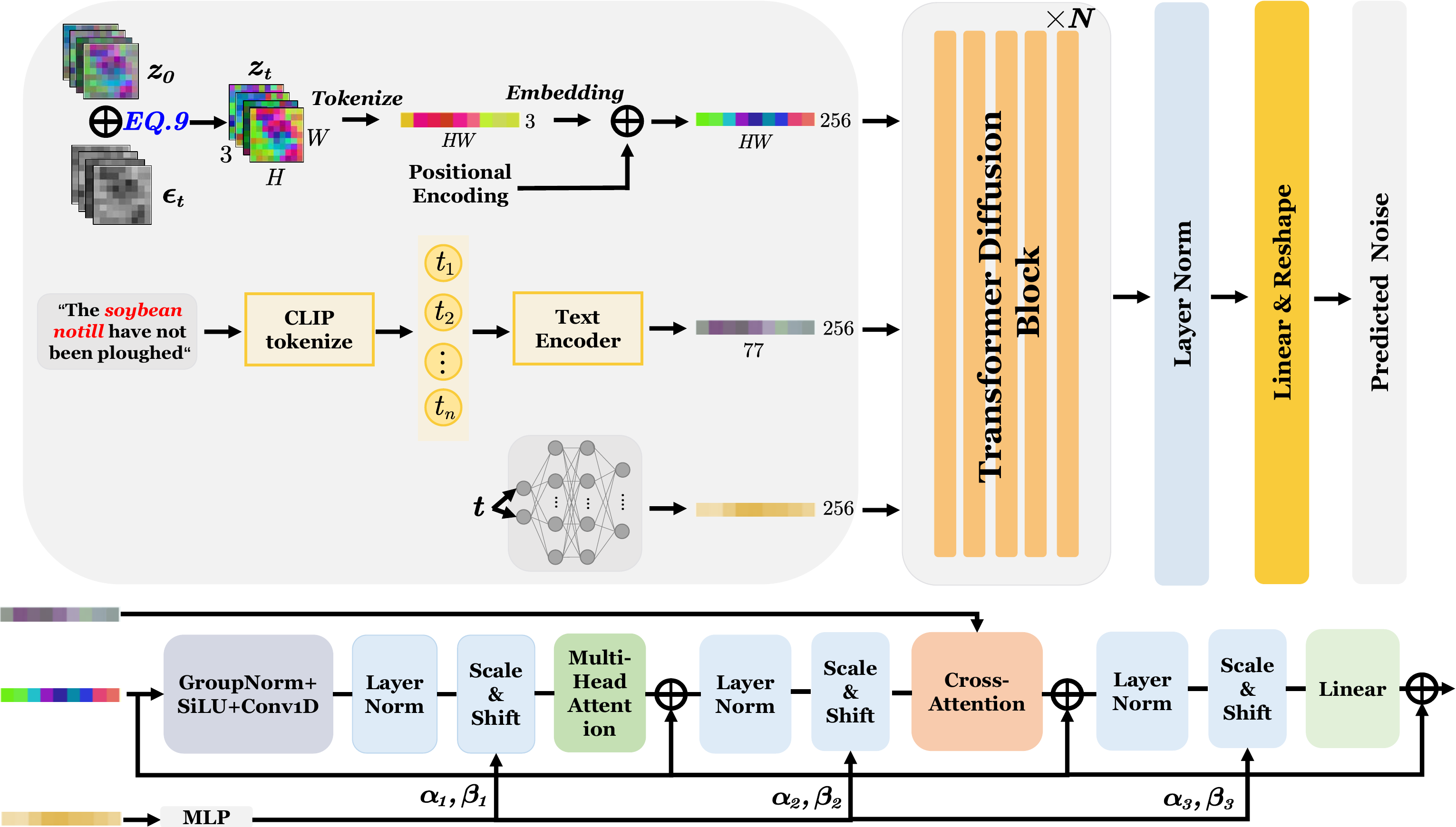}
    \caption{Illustration of the Transformer Diffusion Network (Upper) and details of blocks (Bottom).}
    \label{Transformer Diffusion Network}
\end{figure}

\subsubsection{\textbf{Noise Schedule}}
A small enough \(a_{T}\) guarantees a convergent to a normal distribution as the sampling process start value. The total time step \(T\) is set to 500. Inspired by \cite{dhariwal2021diffusion}, we take a linear schedule with minimum 0.0001 and maximum 0.02, where \(\bar{\alpha}_{t} = 4.04 \times 10^{-5}\). As we can see from \autoref{EQ9}, when \(t=T\), \(z_{T}\) hardly contain the information of \(z_0\) and can be seen as normal distribution. So, we can start the denoising process with a randomly sampled noise from the normal distribution.

\subsubsection{\textbf{Overview of Proposed Language-informed Diffusion Model}}
The detail of architecture can be seen in \autoref{Transformer Diffusion Network}. Txt2HSI-LDM(VAE) has three kinds of input, i.e., noised latent feature \(z_{t}\), corresponding time step \(t\) and text description \(l\). For unifying the dimension of the input tensor, and adapting to the input of the Transformer Diffusion Block (TDB), we set all the dimension of embedded features \(d_{emb}\) (including latent feature, embedded textual description, and timesteps) before feeding into the TDB as 512. Similar to the Transformer model, for both text and latent feature tokens, we use an embedding layer with positional encoding to transform them into the embedding space with a dimension of \(d_{emb}\), which can help to inject the relative or absolute position information among different tokens in the sequence. For time step embedding, we use multilinear layers to project discrete-time information into continuous space with a dimension of \(d_{emb}\). Let \(c_{l} \in \mathbb{R}^{n \times d_{emb}}\) and \(z_{t}^{emb} \in \mathbb{R}^{HW \times d_{emb}}\) and \(t_{emb} \in \mathbb{R}^{d_{emb}}\) denote the obtained text embedding, latent feature embedding, and time embedding, respectively. \(H\) and \(W\) represent the spatial size of hyperspectral patch data, and \(n\) represents the total length of the text tokens, here \(n=77\). TDB is mainly built on Multi-Head Attention (MHA), Cross-Attention, and skip-connection. The \(t_{emb}\) is used to rescale the feature with scale and shift before the skip connection. We project \(t_{emb}\) to \(\alpha_{i}, \beta_{i}, i=1,2,3\) as shown in \autoref{shift_scle}. The input feature will be mapped multiplied by \(\alpha_{i}\) and added by \(\beta_{i}\).
\begin{align}
    \alpha_{i}, \beta_{i} = \text{Chunk}(\text{MLP}(t_{emb}))
    \label{shift_scle}
\end{align}
where \(\text{Chunk}\) is splitting the feature into two parts with equal size along the channel dimension. \(\text{MLP}\) is implemented as a SiLU activation followed by a linear layer.
\begin{figure}[]
    \centering
    \includegraphics[scale=0.3]{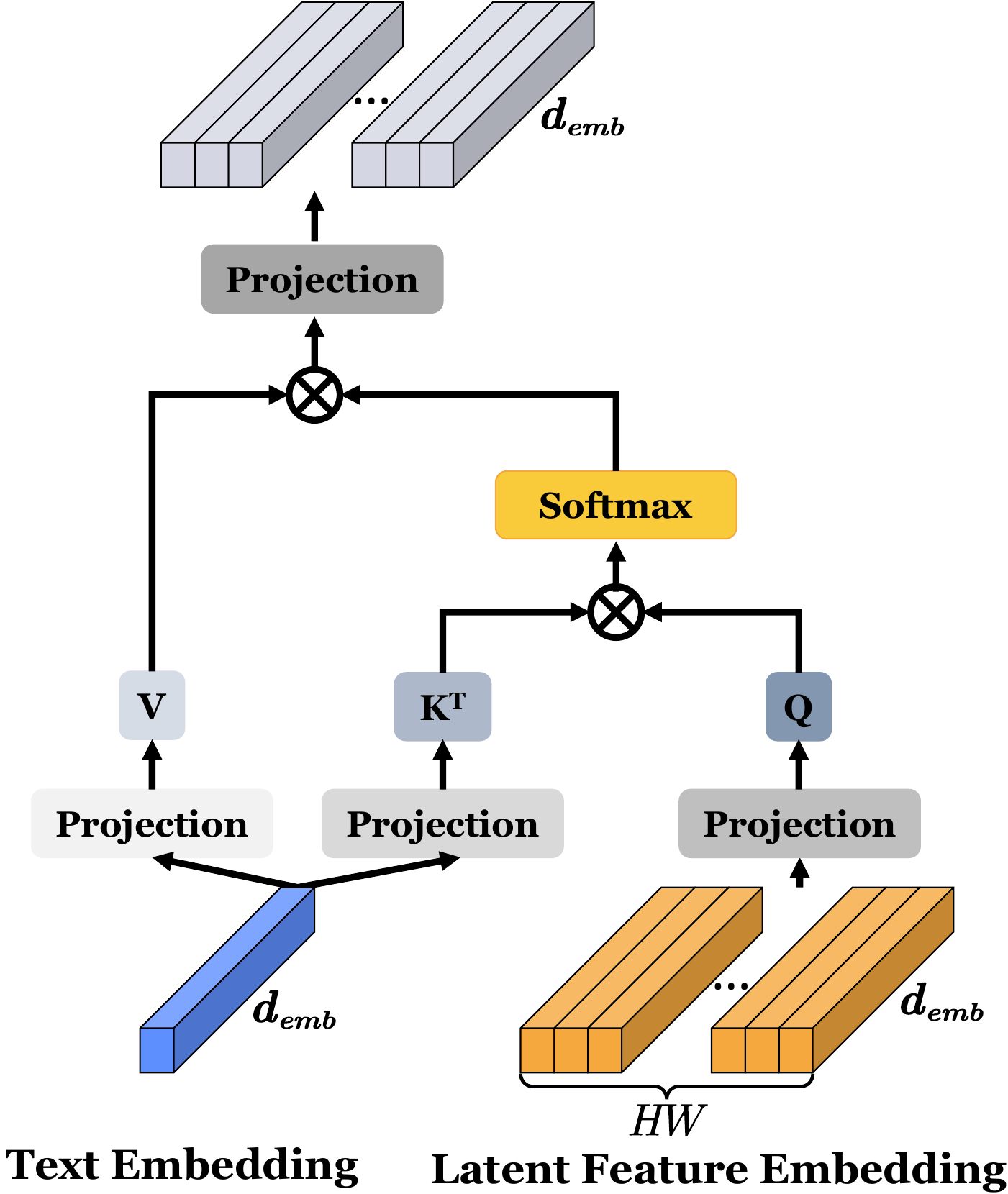}
    \caption{Illustration of the Cross-Attention}
    \label{Cross-Attention}
\end{figure}

\begin{figure}[]
    \centering
    \includegraphics[width=0.49\textwidth]{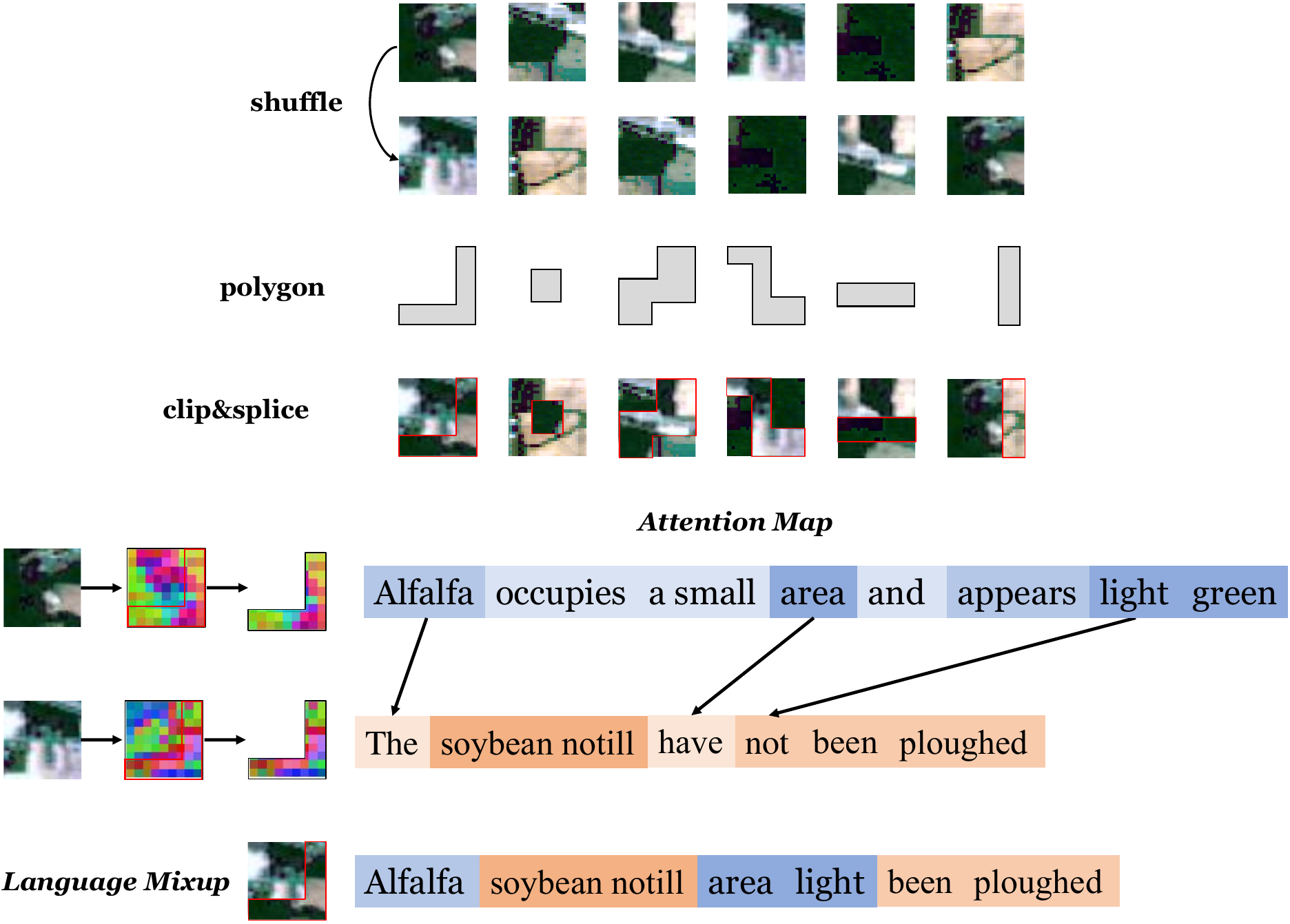}
    \caption{Illustration of the random polygon spatial clipping (RPSC) and language mixup. The deeper color means the higher attention value.}
    \label{RPSC}
\end{figure}

\begin{figure*}[]
    \centering
    \includegraphics[width=0.99\textwidth]{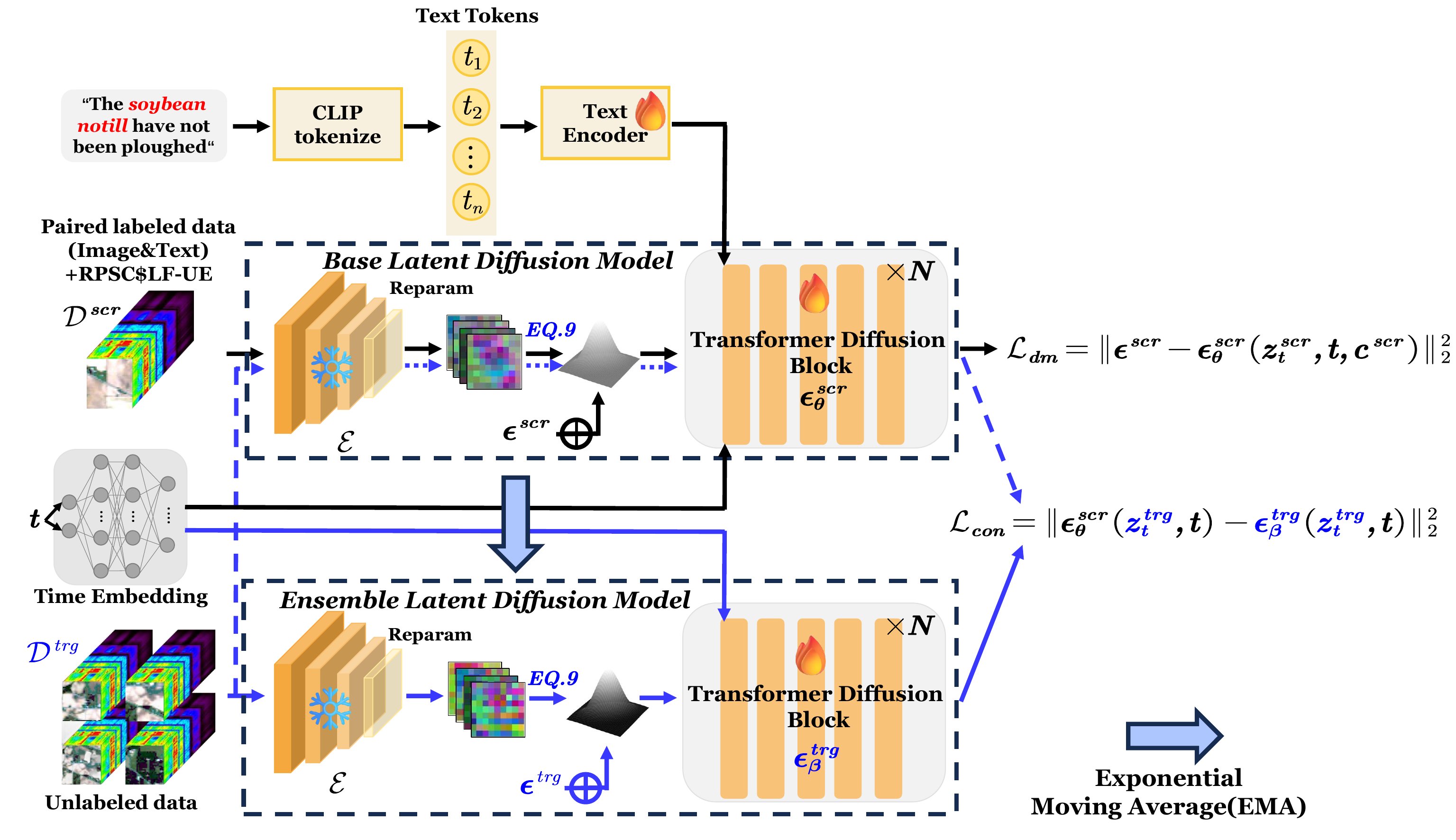}
    \caption{Detail of the semi-supervised learning part. In the upper part, the text descriptions are first converted into tokens, and then the vectors are encoded by the on-hand CLIP pre-trained weight "ViT-B-32.pt". On the bottom part, a bunch of unlabeled data are respectively fed into Base Latent diffusion Model \(\epsilon_{\theta}^{scr}\) and Ensemble Latent Diffusion Model \(\epsilon_{\beta}^{trg}\).}
    \label{semi_framwork}
\end{figure*}

\subsubsection{\textbf{Text Encoder}} \label{txt}
First we need to justify the text description rules used in this paper. We design two different kinds of descriptions, i.e., coarse-grained and fine-grained. The coarse-grained description is constructed by the template of '\textbf{A hyperspectral image of + \textit{class name}}', such as, "\textbf{A hyperspectral image of trees}". While the fine-grained text description is manually designed based on the distribution, color, shape, and adjacency relationship \cite{zhang-LDGNet2023}, for example, "\textbf{\textit{Alfalfa} occupies a small area and appears light green}", "\textbf{The \textit{shadows }is next to the buildings and appear black color}". {This solution is widely used in multi-source (image and language model) as well as large language model inspired remote sensing foundation model, such as RSGPT \cite{HU2025272}, LIVEnet \cite{Dangyy2024}. In RSGPT, the language for large-scale remote sensing scene images is manually designed by experts, containing the scene theme, shape, color, quantity, and position (relative and absolute). The first paper that discussed the language generated by machines should also consider these five aspects \cite{7891049}. These two studies focus on the large-scale remote sensing scene captioning, which is not suitable for patch-level hyperspectral image classification. Although using large language models (LLMs) can generate detailed descriptions, remote sensing texts often contain technical terms (such as "NDVI anomaly area"), which can not be decoded by LLMs, such as DeepSeek, ChatGPT, which are pretrained on natural languages. We believe that the powerful performance of RSGPT \cite{HU2025272} is based on the refined manual annotation, costing a total of 35 person days, involving 5 remote sensing experts. Another example \cite{das2025hyperspectralimagelandcover} can support that we prefer to rely on manually crafted text descriptions. In \cite{das2025hyperspectralimagelandcover},  researchers built hyperspectral image land cover captioning datasets by automatic generation and manual refinement, but the generated texts were more likely to be general characteristics, for example, "\textbf{\textit{Alfalfa: Dense leafy canopy with soft tones across wide patches}}". So, manual annotations make sure that we can also take into account the general topological relations between ground cover types while keeping the whole image's spatial features.}

The text encoder is a 512-wide language-model transformer \cite{CLIP2021} with three layers and eight attention heads. Firstly, the text tokens is obtained by a lower-cased byte pair encoding (BPE). The maximum sequence of text length is limited to 77, and we use 0 as the placeholder to fill the empty tokens, because the number of words in the input text description is usually smaller than 77. This linguistic feature with positional encoding is then normalized by layer and linearly projected into the latent space.

\subsubsection{\textbf{Visual–Linguistic Cross Attention}}
Given the property that the attention mechanism can realize the multi-modality data fusion \cite{Guoxin2024, Huyuan2023}, hence, we design the cross-attention to fuse to information of text embedding feature and latent feature embedding to ensure its supervision of language on the generation process. The detail of cross-attention is shown in \autoref{Cross-Attention}. The value of cross-attention is computed by the following equation:
\begin{align}
    \text{CA} = \text{Attention}(\mathcal{Q}_{img}, \mathcal{K}_{txt}, \mathcal{V}_{txt}) = \sigma(\frac{\mathcal{Q}_{img} \mathcal{K}_{txt}^T}{\sqrt{d_{emb}}}) \mathcal{V}_{txt}
\end{align}
where, \(\mathcal{Q}_{img} = \mathcal{F}_{img} \mathcal{W}^{Q}_{img}, \mathcal{K}_{txt} = \mathcal{F}_{txt} \mathcal{W}^{K}_{txt}, \mathcal{V}_{txt} = \mathcal{F}_{txt} \mathcal{W}^{V}_{txt}\) are queries, keys, and values, \(\mathcal{W}^{Q}_{img}, \mathcal{W}^{K}_{txt}, \mathcal{W}^{V}_{txt}\) represent the projection weight of \(\mathcal{Q}_{img}, \mathcal{K}_{txt}, \mathcal{V}_{txt}\). \(\mathcal{F}_{img}\) and \(\mathcal{F}_{txt}\) represent the feature of image and text. \(\sigma\) indicates \textit{softmax} function.

\subsubsection{\textbf{Semi-supervised Denoising Network and Loss Function}}
For better dealing with the Imbalanced-Small Sample Data (ISSD) problem in HSIC, and considering the sparse labeled data in the real world, one of the effective and promising methods is to fully learn knowledge from the unlabeled data and facilitate information exchange between the knowledge of unlabeled and labeled data in a way of knowledge distillation. Semi-supervised learning framework can realize this kind of purpose. Some related works and contents can be found in \cite{wangyx2021, Xuyh2024, Wanglg2024, Liuh2024}. 

Additionally, for better simulating the spectral mixing and heterogeneity and expanding the diversity of training data, we introduce two kinds of non-parameter methods, i.e., random polygon spatial clipping (RPSC) and uncertainty estimation of latent feature (LF-UE). The process of RPSC is depicted in \autoref{RPSC}, and the language description of the clipped sample is generated by mixing the attention value of data before clipping \cite{yoon-etal-2021-ssmix}. The generalized sample of latent space feature \(z\) obtained by LF-UE is formulated by the \autoref{EQ17}:
% \begin{align}
%     \begin{aligned}
%         &\mu(z) = \frac{1}{HW} \underset{h=1}{\overset{H}{\Sigma}} \underset{w=1}{\overset{W}{\Sigma}} z_{b,c,h,w}\\
%         &\sigma(z) = \frac{1}{HW} \underset{h=1}{\overset{H}{\Sigma}} \underset{w=1}{\overset{W}{\Sigma}} (z_{b,c,h,w} - \mu(z))^{2} \\
%         &\Sigma_{\mu}^{2} (z) = \frac{1}{B} \underset{b=1}{\overset{B}{\Sigma}} (\mu_{bc} (z) - \mathbb{E}_{b}(\mu_{bc}(z)))^{2}\\
%         &\Sigma_{\sigma}^{2} (z) = \frac{1}{B} \underset{b=1}{\overset{B}{\Sigma}} (\sigma_{bc}(z) - \mathbb{E}_{b}(\sigma_{bc}(z)))^2 \\
%         &\phi(z) = \mu(z) + \epsilon_{\mu} \Sigma_{\mu}(z), \epsilon_{\mu} \sim \mathcal{N}(0, \mathbf{I})\\
%         &\varphi(z) = \sigma(z) + \epsilon_{\sigma} \Sigma_{\sigma}(z), \epsilon_{\sigma} \sim \mathcal{N}(0, \mathbf{I}) \\
%         &\tilde{z} = \phi(z) + \varphi(z) \times \frac{z-\mu(z)}{\sigma(z)}
%     \end{aligned}\label{EQ17}
% \end{align} 

\begin{strip}
\begin{align}
    \begin{aligned}
        &\mu(z) = \frac{1}{HW} \sum_{h=1}^{H} \sum_{w=1}^{W} z_{b,c,h,w}
        &\sigma(z) = \frac{1}{HW} \sum_{h=1}^{H} \sum_{w=1}^{W} (z_{b,c,h,w} - \mu(z))^{2} \\
        &\Sigma_{\mu}^{2} (z) = \frac{1}{B} \sum_{b=1}^{B} (\mu_{bc} (z) - \mathbb{E}_{b}(\mu_{bc}(z)))^{2}
        &\Sigma_{\sigma}^{2} (z) = \frac{1}{B} \sum_{b=1}^{B} (\sigma_{bc}(z) - \mathbb{E}_{b}(\sigma_{bc}(z)))^2 \\
        &\phi(z) = \mu(z) + \epsilon_{\mu} \Sigma_{\mu}(z), \quad \epsilon_{\mu} \sim \mathcal{N}(0, \mathbf{I})
        &\varphi(z) = \sigma(z) + \epsilon_{\sigma} \Sigma_{\sigma}(z), \quad \epsilon_{\sigma} \sim \mathcal{N}(0, \mathbf{I}) \\
        &\tilde{z} = \phi(z) + \varphi(z) \times \frac{z-\mu(z)}{\sigma(z)}
    \end{aligned}\label{EQ17}
\end{align}
\end{strip}

As shown in  \autoref{semi_framwork}, \(\mathcal{D}^{scr}\) and \(\mathcal{D}^{trg}\) represent limited labeled patch data with corresponding fine-grained text descriptions and bunch of unlabeled patch data respectively. On the upper part of  \autoref{semi_framwork}, the CLIP tokenization returns the tokenized representation of given text input \(l\), for example, "The soybean notill fields have not been ploughed" will be converted into a two-dimensional tensor [\textbf{49406}, 518, 34606, 2534, 660,   720, 783, 1025, 33811, 538, \textbf{ 49407},0, 0, ..., 0] \(\in \mathbb{R}^{77}\). Next, the tensor will be encoded by an on-hand pre-trained text encoder \(\mathcal{T}_{\phi}\) as the condition of base latent diffusion \(\epsilon_{\theta}^{scr}\). On the bottom part of  \autoref{semi_framwork}, abundant unlabeled patch data randomly selected are fed into the unconditional ensemble latent diffusion model \(\epsilon_{\beta}^{trg}\) with parameter \(\beta\) before being encoded by VAE. Both \(\epsilon_{\theta}^{scr}\) and \(\epsilon_{\beta}^{trg}\) share the same architecture. During the training phase, the diffusion loss calculated on the labeled sample with the following equation \autoref{EQ13}:
\begin{align}
    \mathcal{L}_{dm}(\theta, \phi, z_{t}^{scr}, c^{scr}) = \rVert \epsilon^{scr} - \epsilon_{\theta}^{scr}(z_{t}^{scr}, t, c^{scr})\lVert_{2}^{2}
\end{align}
where \(c^{scr}=\mathcal{T}_{\phi}(\text{tokenize}(l))\), \(\phi\) is the weight of text encoder \(\mathcal{T}\), we optimize \(\phi\) together with \(\epsilon_{\theta}^{scr}\), while the unlabeled samples are fed into both the base latent diffusion model and the ensemble latent diffusion model to calculate the diffusion consistency loss with mean-squared error as: 
\begin{align}
        \mathcal{L}_{con}(\theta, z_{t}^{trg}) = \rVert \epsilon_{\theta}^{scr}(z_{t}^{trg}, t) - \epsilon_{\beta}^{trg}(z_{t}^{trg}, t)\lVert_{2}^{2}
\end{align}
At each iteration, the \(\epsilon_{\theta}^{scr}\) is updated with the gradients from both two losses. The \(\epsilon_{\beta}^{trg}\) does not participate in this process. Instead, its parameters are manually updated using the exponential moving average (EMA) of the historical parameters in the \(\epsilon_{\theta}^{scr}\) to realize knowledge distillation. At the  \(\tau\)th iteration, \(\beta^{\tau}\) be updated by:
\begin{align}\label{EQ19}
    \beta^{\tau} = \alpha \beta^{\tau - 1} + (1 - \alpha) \theta^{\tau}
\end{align}
where \(\alpha\) is a smoothing coefficient, we set 0.99. Our goal is to let the \(\epsilon_{\theta}^{scr}\) learn knowledge from the \(\epsilon_{\beta}^{trg}\) on unlabeled data. Therefore, the parameters in \(\epsilon_{\beta}^{trg}\) does not participant the optimization. The full objective function for training the base latent diffusion mode \(\epsilon_{\theta}^{scr}\) can be formulated as:
\begin{align} 
    \underset{\theta, \phi}{\min} (\mathcal{L}_{dm}(\theta, z_{t}^{scr}, c^{scr}) + \mathcal{L}_{con}(\theta, z_{t}^{trg}))
\end{align}
% \subsubsection{Deep Image Feature Encoder}
% \subsubsection{Overall Architecture}
\subsubsection{\textbf{Processes of Optimizing and Sampling}}
To optimize the diffusion model, we take a random hyperspectral patch \(\mathcal{H}\) with corresponding text description \(l\), time step \(t\), as well as Gaussian noise \(\epsilon\) and update the noise prediction model Txt2HSI-LDM(VAE)  at each iteration. For the sampling process, we use \autoref{EQ14} through the DDIM method to generate the latent space feature \(z_{0}\), and then the decoder \(\mathcal{D}\) of VAE is used to generate the synthesis image patch. 

We summarize the optimization and sampling procedures of the latent diffusion model in algorithms 
\autoref{algo1} and \autoref{algo2}.
\begin{figure*}[htbp]
\centering
\subfigure[]{\includegraphics[width=0.16\textwidth]{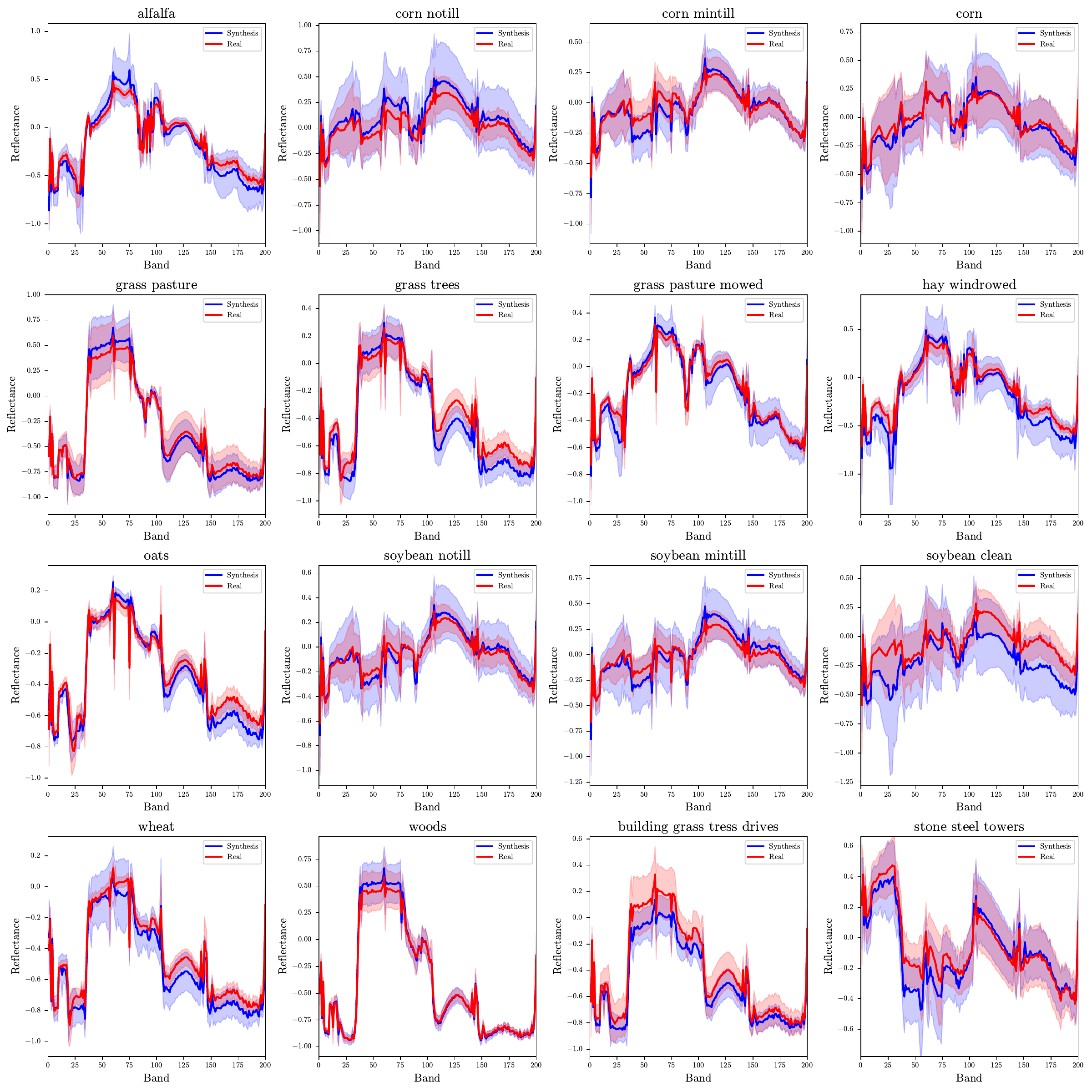}}
\subfigure[]
{\includegraphics[width=0.16\textwidth]{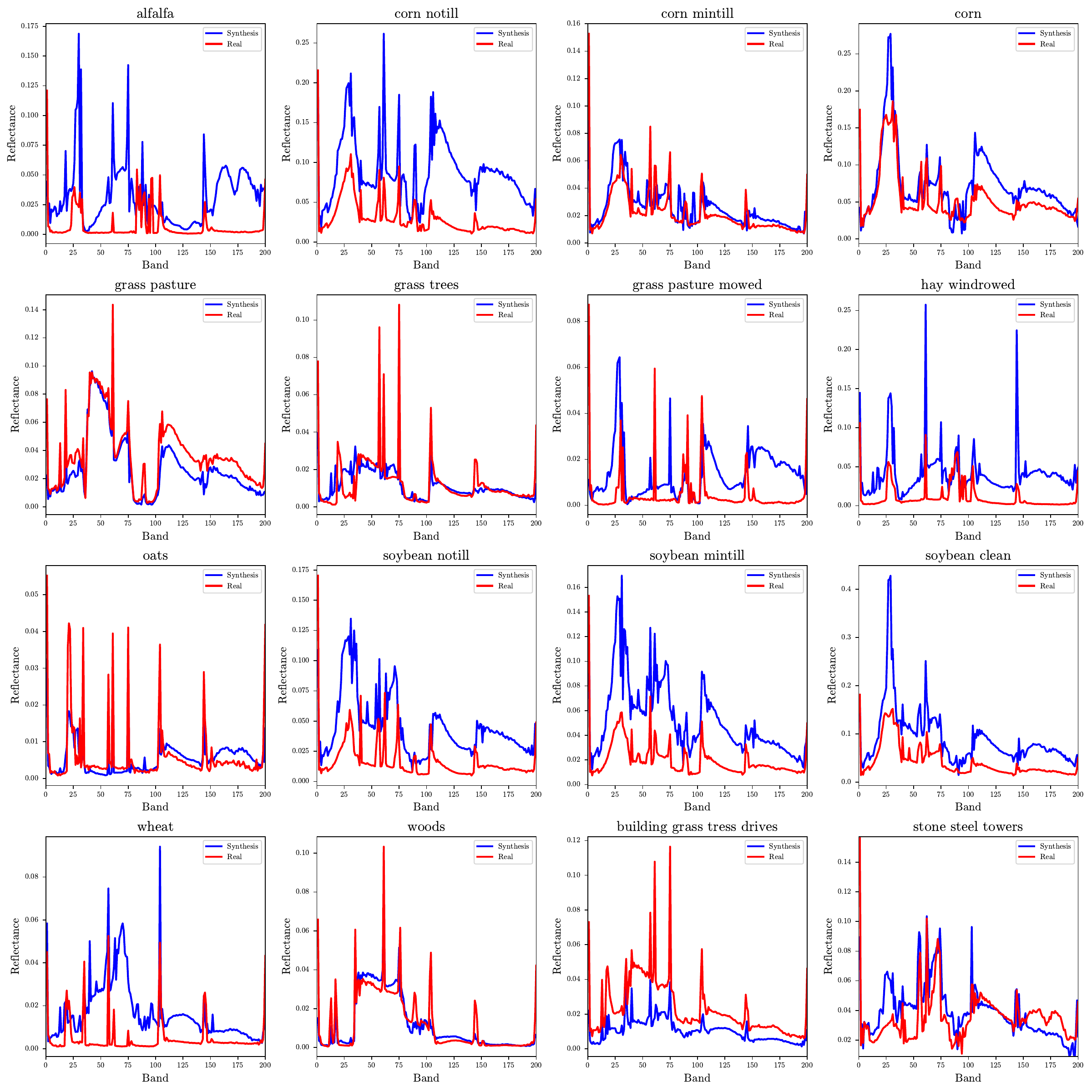}}
\subfigure[]{\includegraphics[width=0.16\textwidth]{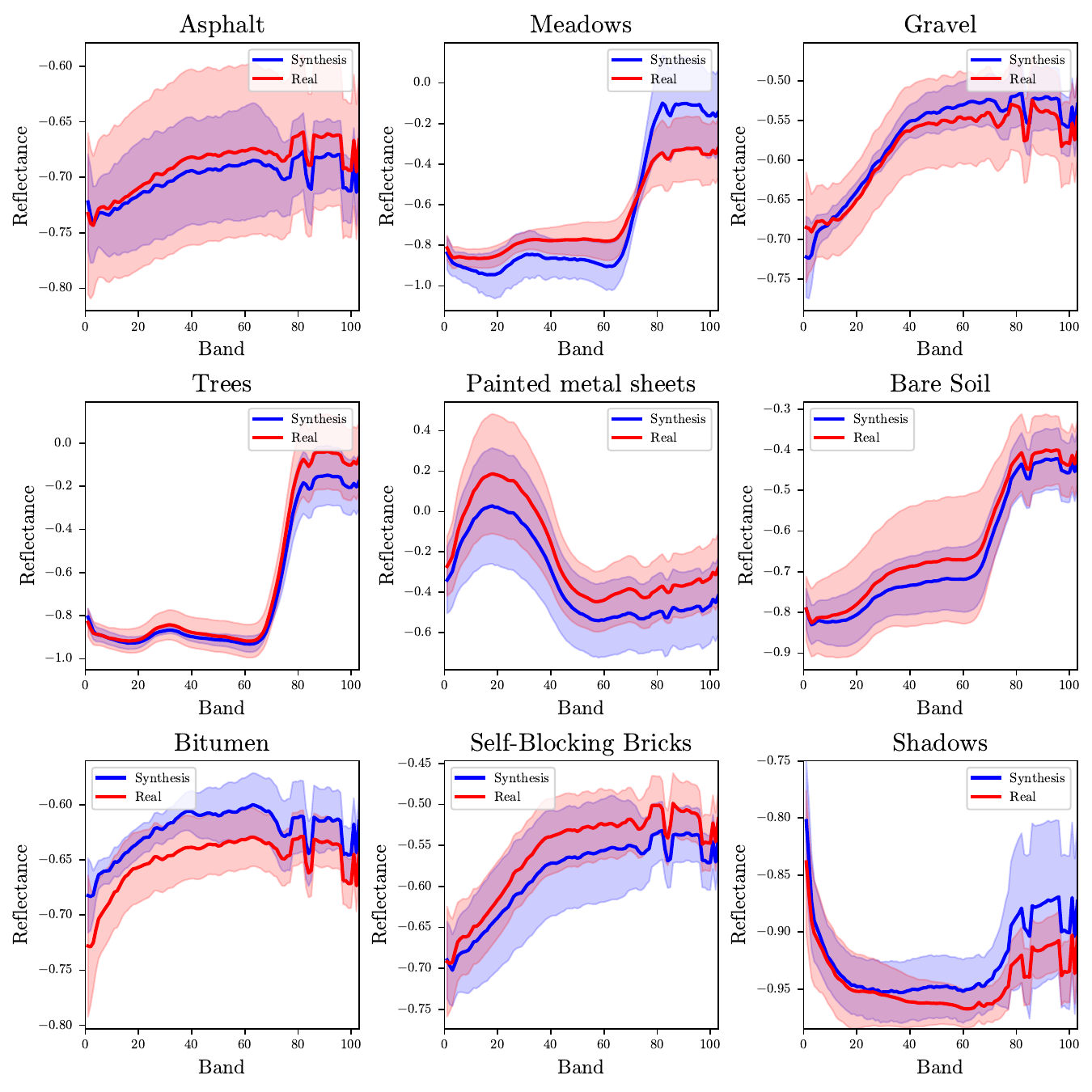}}
\subfigure[]
{\includegraphics[width=0.16\textwidth]{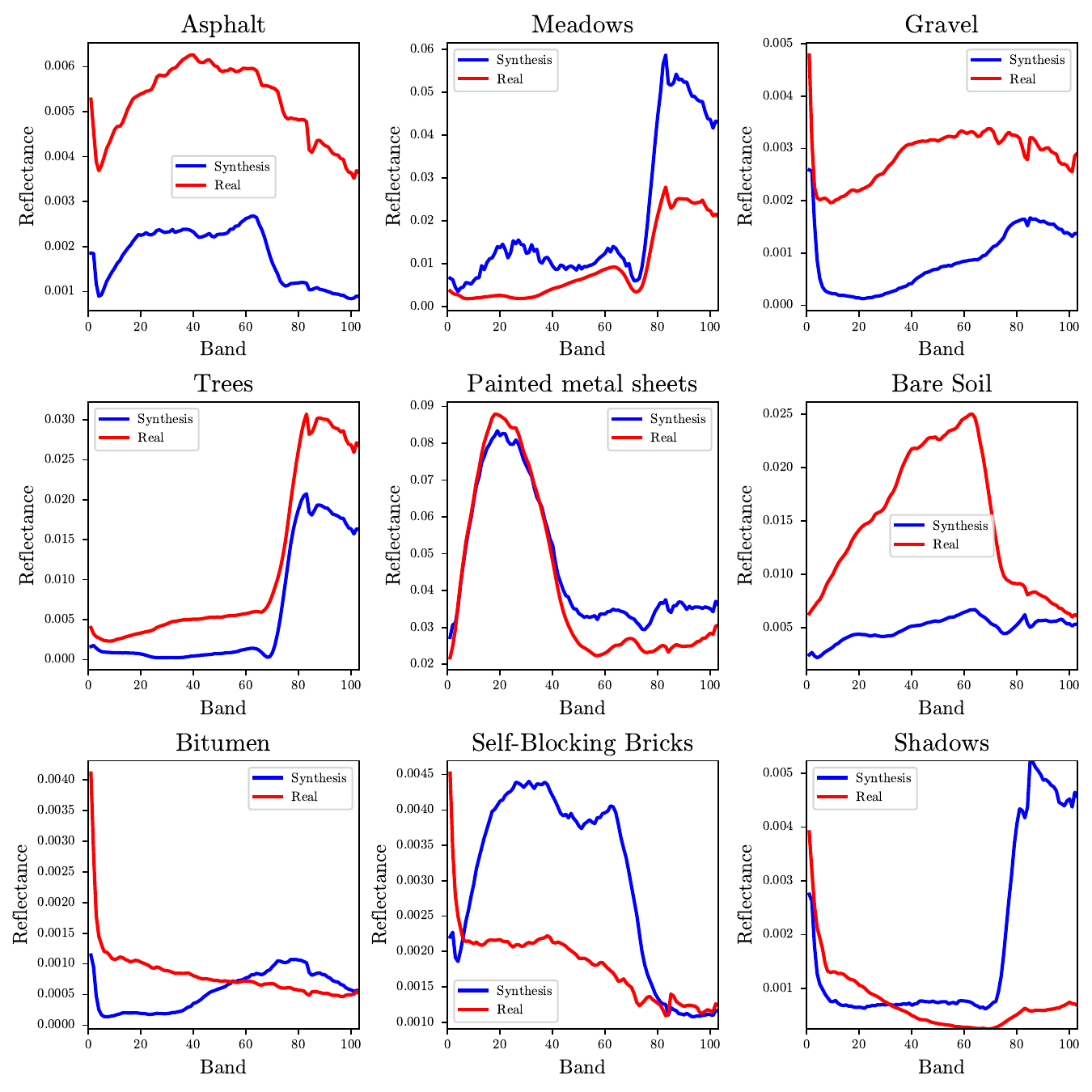}}
\subfigure[]{\includegraphics[width=0.16\textwidth]{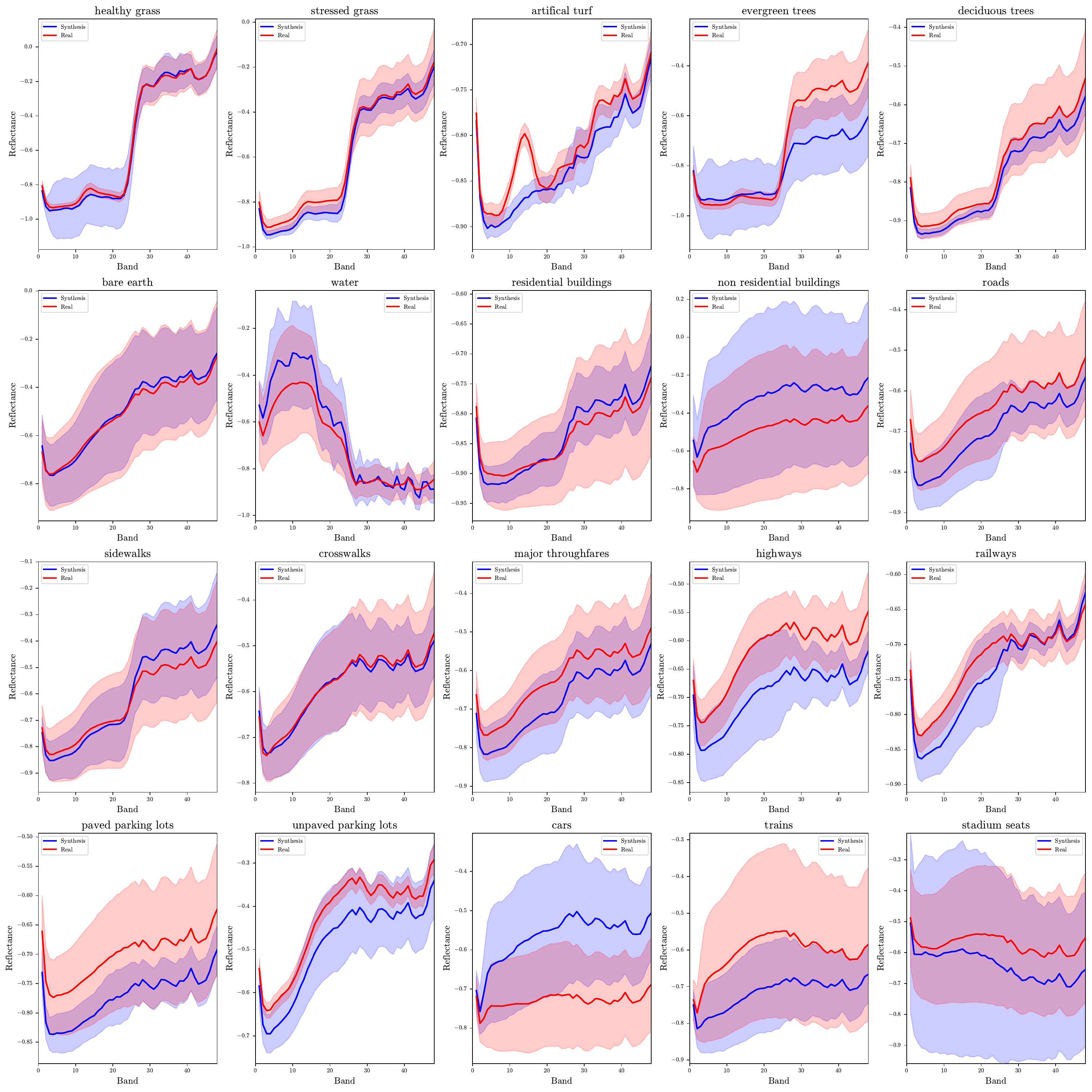}}
\subfigure[]
{\includegraphics[width=0.16\textwidth]{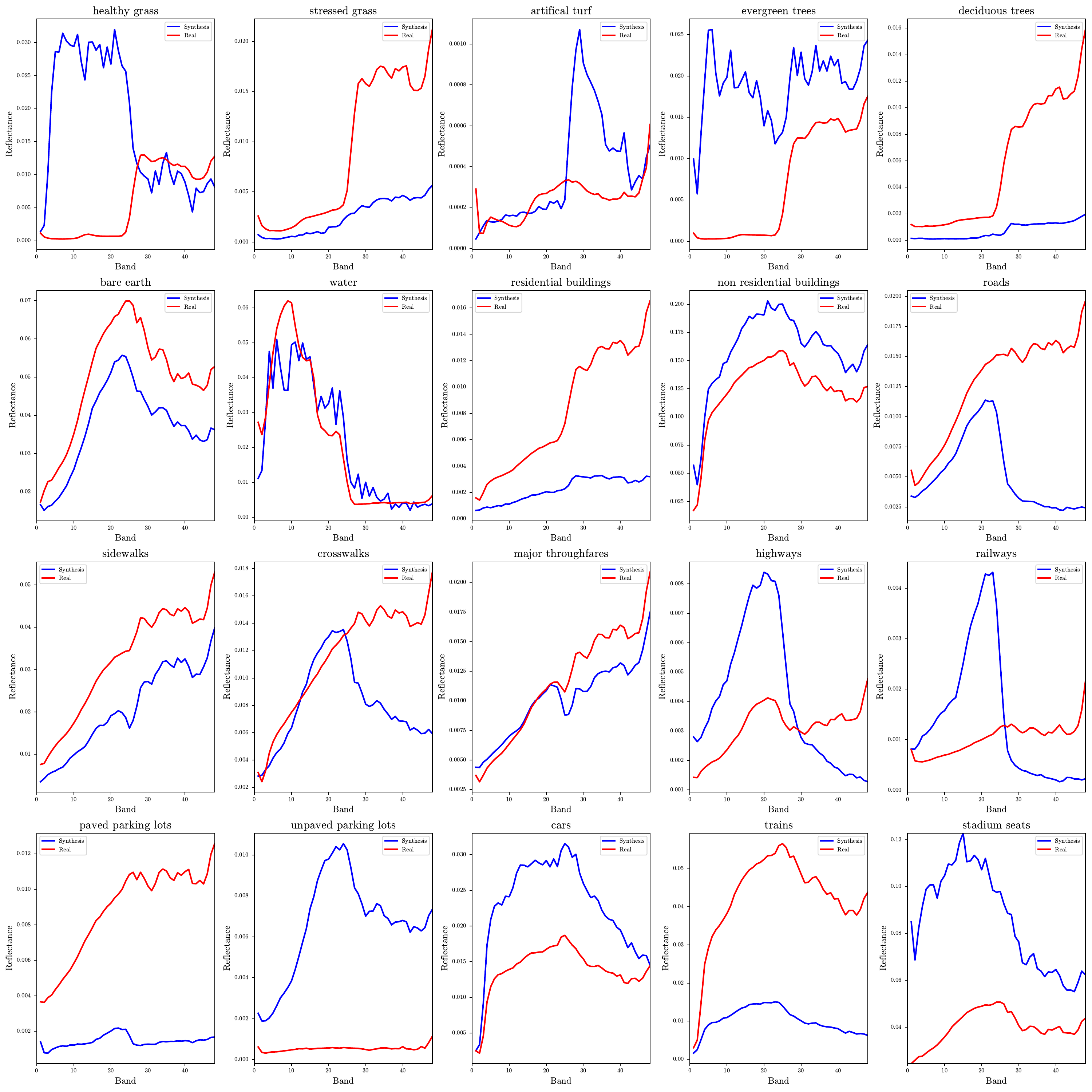}}

\caption{{Statistical characteristics comparison between the generated spectral curves and real ones on the Indian Pines, Pavia University dataset, and Houston18 dataset for different land cover and land use categories. The red solid line represents all the real labeled data, while the blue solid line represents the generated sample. The red and blue shadows represent the standard deviations (std) of real and generated samples. (a) error bar plot on Indian Pines. (b) std along the spectral dimension on Indian Pines. (c) error bar plot at Pavia University. (d) std along the spectral dimension at Pavia University. Zoom in for the best. (e) error bar plot at Houston18. (f) std along the spectral dimension at Houston18. Zoom in for the best.}}
\label{error_IN}
\end{figure*}
\begin{algorithm}
\caption{{Training of the Semi-supervised Latent Diffusion Model Txt2HSI-LDM(VAE)}}
    \begin{algorithmic}[1]  \label{algo1}
        \REQUIRE labeled data \(\mathcal{D}^{scr}\) with corresponding text             description \(l\), unlabeled data \(\mathcal{D}^{trg}\), learning rate \(\eta\), total epoch \(E\), initialized parameter \(\theta, \beta, \phi\) for \(\epsilon^{scr}, \epsilon^{trg}, \mathcal{T}\) respectively
        \ENSURE The trained model \(\epsilon^{scr}\) with parameter \(\theta\)
        \WHILE{$e<E$}
        \STATE Sample \(\mathcal{D}^{scr} \sim q(\mathcal{D}^{scr})\), \(\mathcal{D}^{trg} \sim q(\mathcal{D}^{trg})\),\(\epsilon^{scr} \sim \mathcal{N}(0, \mathbf{I})\), \(t \sim U({0,...,T})\), \(p \sim U(0,1)\)
        \STATE Feed \(\mathcal{D}^{scr}\) and \(\mathcal{D}^{trg}\) into VAE encoder and get \(z^{scr}\), \(z^{trg}\) respectively
        \STATE Encode the language using \(\mathcal{T}\) with parameter \(\phi\) and get \(c^{scr}\)
        \IF{$0 < p \leq 1/3$}
        \STATE Predict the noise with \(\epsilon_{\theta}^{scr} (z_{t}^{scr}, t, c^{scr})\)
        \ELSIF{$1/3 < p \leq 2/3$}
        \STATE get \(\tilde{z}^{scr}\) using \autoref{EQ17}
        \STATE Predict the noise with \(\epsilon_{\theta}^{scr} (\tilde{z}^{scr}, t, c^{scr})\)
        \ELSIF{$p > 2/3$}
        \STATE get mixed sample \(\tilde{z}^{scr}\) using RPSC.
        \STATE get corresponding language description \(\tilde{l}\) and embeded feature \(\tilde{c}^{scr}\)
        \STATE  Predict the noise with \(\epsilon_{\theta}^{scr} (\tilde{z}^{scr}, t, \tilde{c}^{scr})\)
        \ENDIF
        \STATE Calculate the gradient descent \(\nabla_{\theta} (\mathcal{L}_{dm}+ \mathcal{L}_{con})\)
        \STATE Update \(\theta\) with \(\theta = \theta - \eta \nabla_{\theta}\)
        \STATE Update \(\beta\) using \autoref{EQ19}
        \ENDWHILE
    \end{algorithmic}
\end{algorithm}

\begin{algorithm}
\caption{Sampling of Txt2HSI-LDM(VAE)}
    \begin{algorithmic}[1] \label{algo2}
        \REQUIRE text description \(l\), the frozen decoder \(\mathcal{D}\) of VAE, the frozen Ensemble Latent Diffusion Model \(\epsilon_{\beta}^{trg}\), guidance coefficient \(\omega\)
        \ENSURE Latent feature \(z_{0} \sim q(z_{0})\) and generated data \(\mathcal{H}^{gen}\)
        \STATE Sample Gaussian noise \(z_{T} \in \mathbb{R}^{4 \times H \times W}\)
         \FOR{$t = T, ..., 1$}
         \STATE Calculate \(\epsilon_{\beta}\) 
         \STATE Sample \(\epsilon_{t} \sim \mathcal{N}(0, \mathbf{I})\) if \(t > 1\) else \(\epsilon_{t} = 0\)
        \STATE Calculate \(z_{t-1}\) using \autoref{EQ14}
        \ENDFOR
        \STATE Generate synthesis hyperspectral patch data \(\mathcal{H}^{gen}\) using \(\mathcal{D}(z_{0})\) 
    \end{algorithmic}
\end{algorithm}

\begin{figure}[!t]
\centering

\subfigure[IN Real]{
\begin{minipage}{0.225\textwidth}
\centering
\begin{tikzpicture}
\node[inner sep=0pt] {\includegraphics[width=\linewidth]{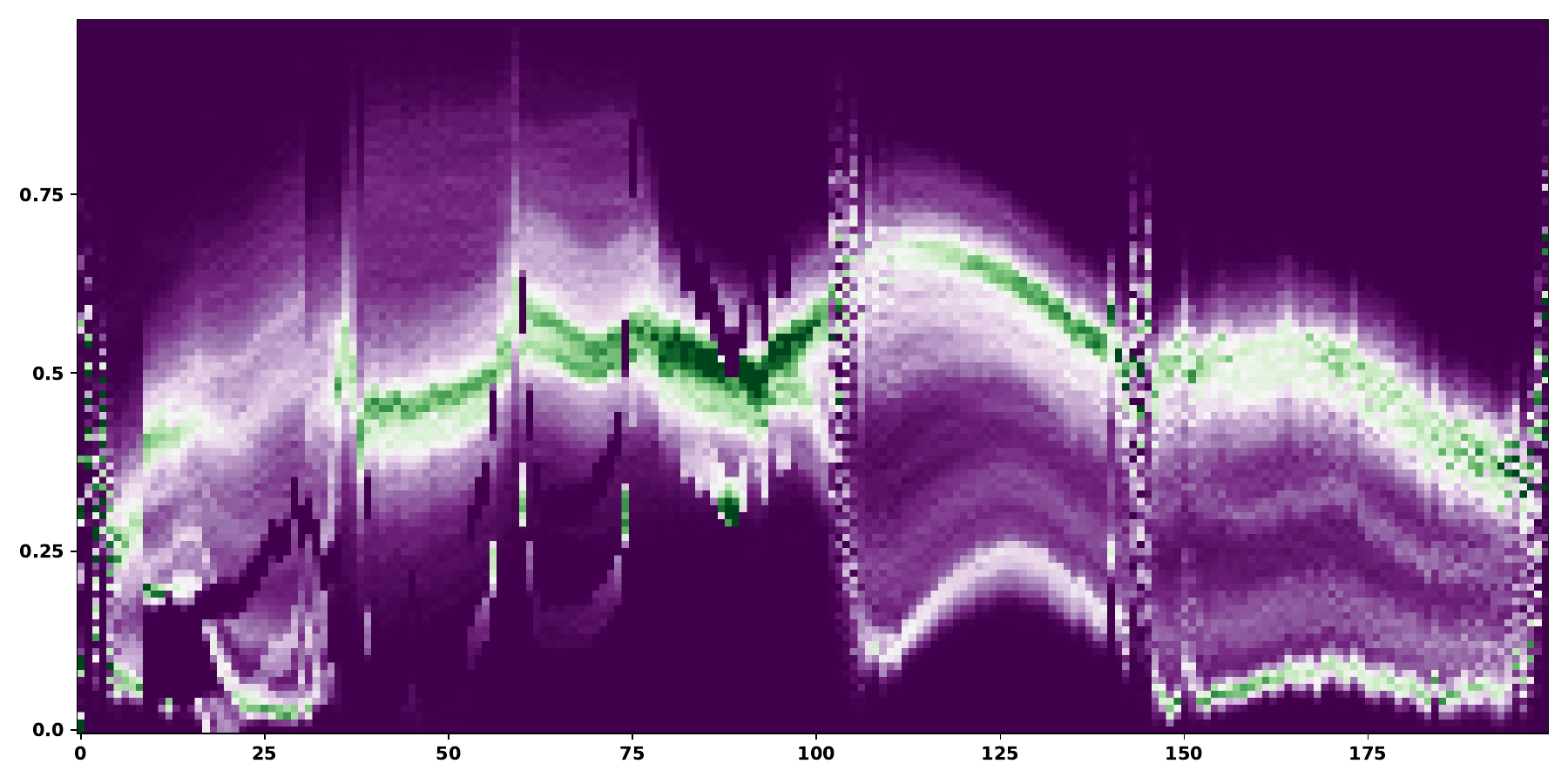}};
\draw[red, very thick, rounded corners] (-0.5,-0.5) rectangle (0.3,0.5);
\end{tikzpicture}
\end{minipage}
}
\hfill
\subfigure[Txt2HSI-LDM(VAE)]{
\begin{minipage}{0.225\textwidth}
\centering
\begin{tikzpicture}
\node[inner sep=0pt] {\includegraphics[width=\linewidth]{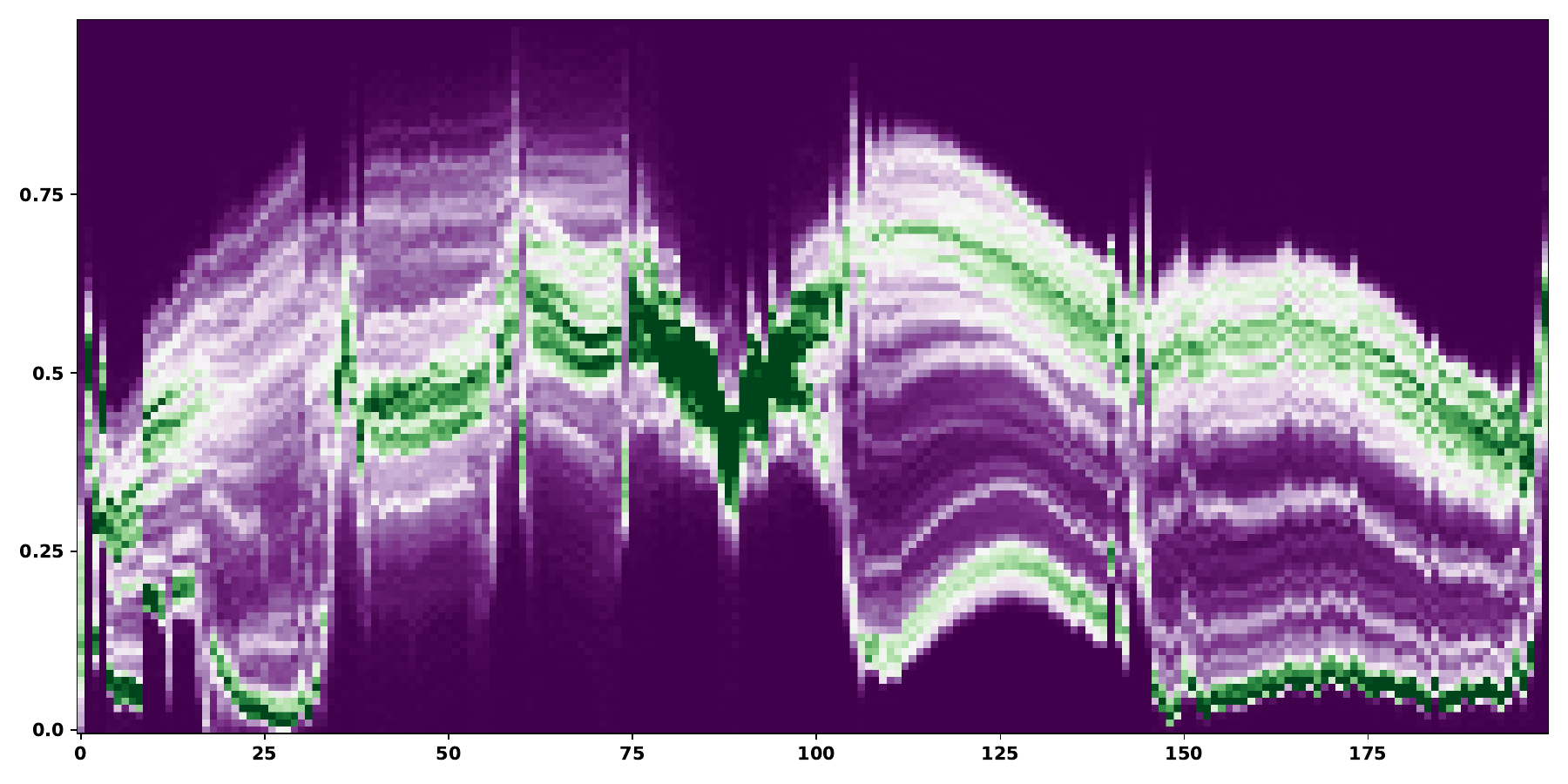}};
\draw[red, very thick, rounded corners](-0.5,-0.5) rectangle (0.3,0.5);
\end{tikzpicture}
\end{minipage}
}
\hfill
\subfigure[PU Real]{
\begin{minipage}{0.225\textwidth}
\centering
\begin{tikzpicture}
\node[inner sep=0pt] {\includegraphics[width=\linewidth]{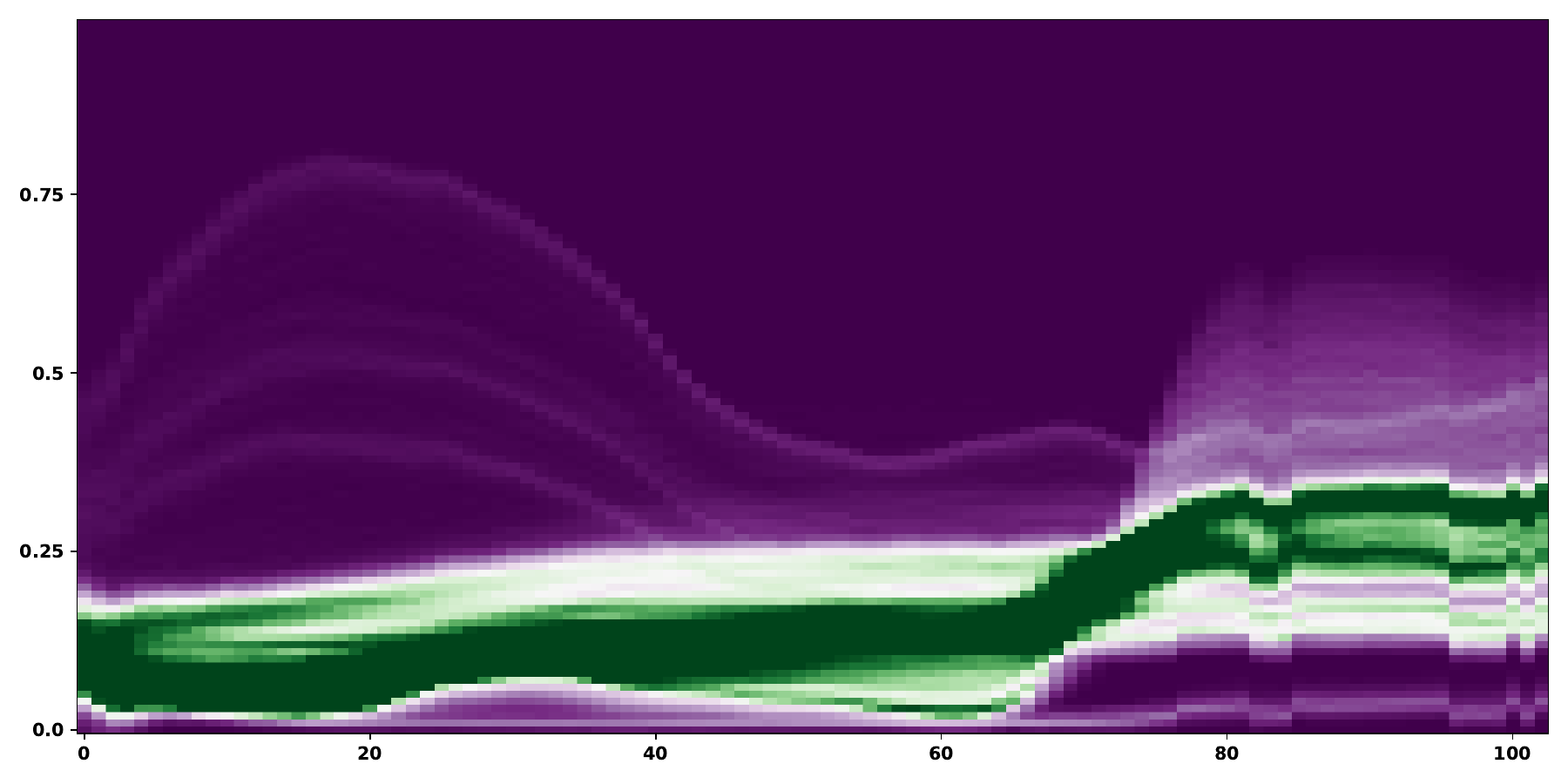}};
\draw[red, very thick, rounded corners] (-1.5,-0.5) rectangle (0.5,0.8);
\end{tikzpicture}
\end{minipage}
}
\hfill
\subfigure[Txt2HSI-LDM(VAE)]{
\begin{minipage}{0.225\textwidth}
\centering
\begin{tikzpicture}
\node[inner sep=0pt] {\includegraphics[width=\linewidth]{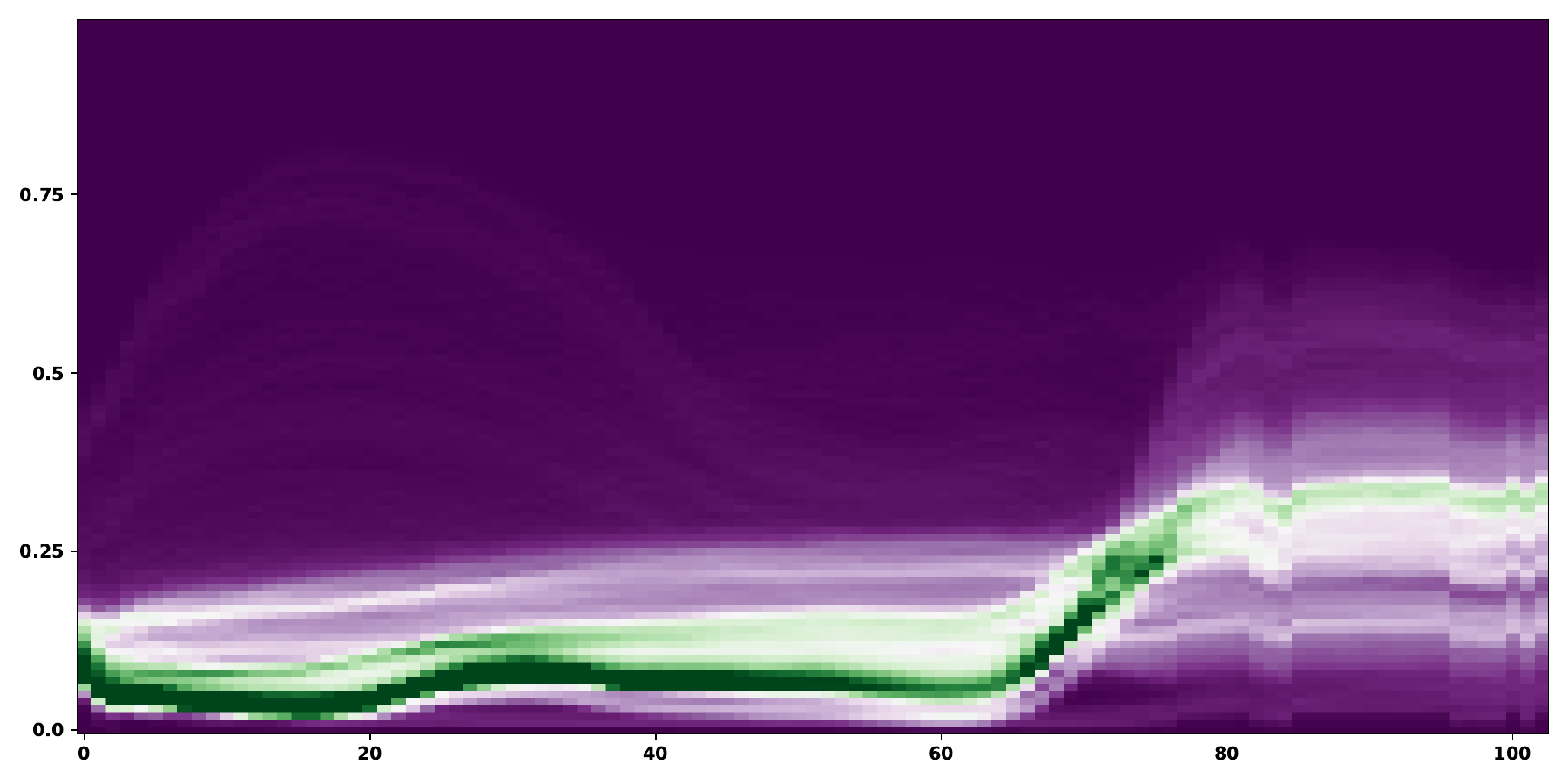}};
\draw[red, very thick, rounded corners](-1.5,-0.5) rectangle (0.5,0.8);
\end{tikzpicture}
\end{minipage}
}
\hfill
\subfigure[Houston Real]{
\begin{minipage}{0.225\textwidth}
\centering
\begin{tikzpicture}
\node[inner sep=0pt] {\includegraphics[width=\linewidth]{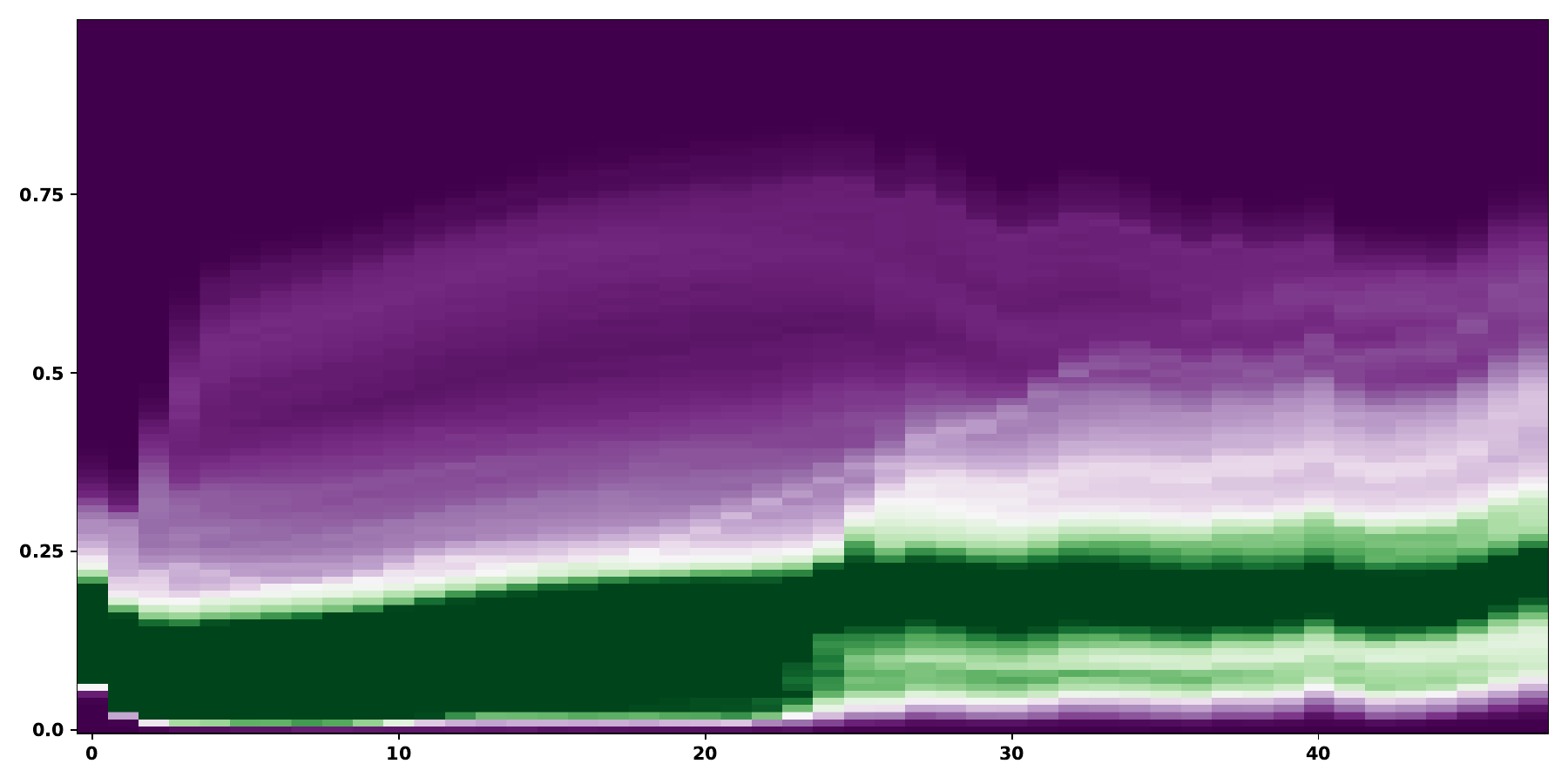}};
\draw[red, very thick, rounded corners](-1.5,-0.5) rectangle (0.5,0.8);
\end{tikzpicture}
\end{minipage}
}
\hfill
\subfigure[Txt2HSI-LDM(VAE)]{
\begin{minipage}{0.225\textwidth}
\centering
\begin{tikzpicture}
\node[inner sep=0pt] {\includegraphics[width=\linewidth]{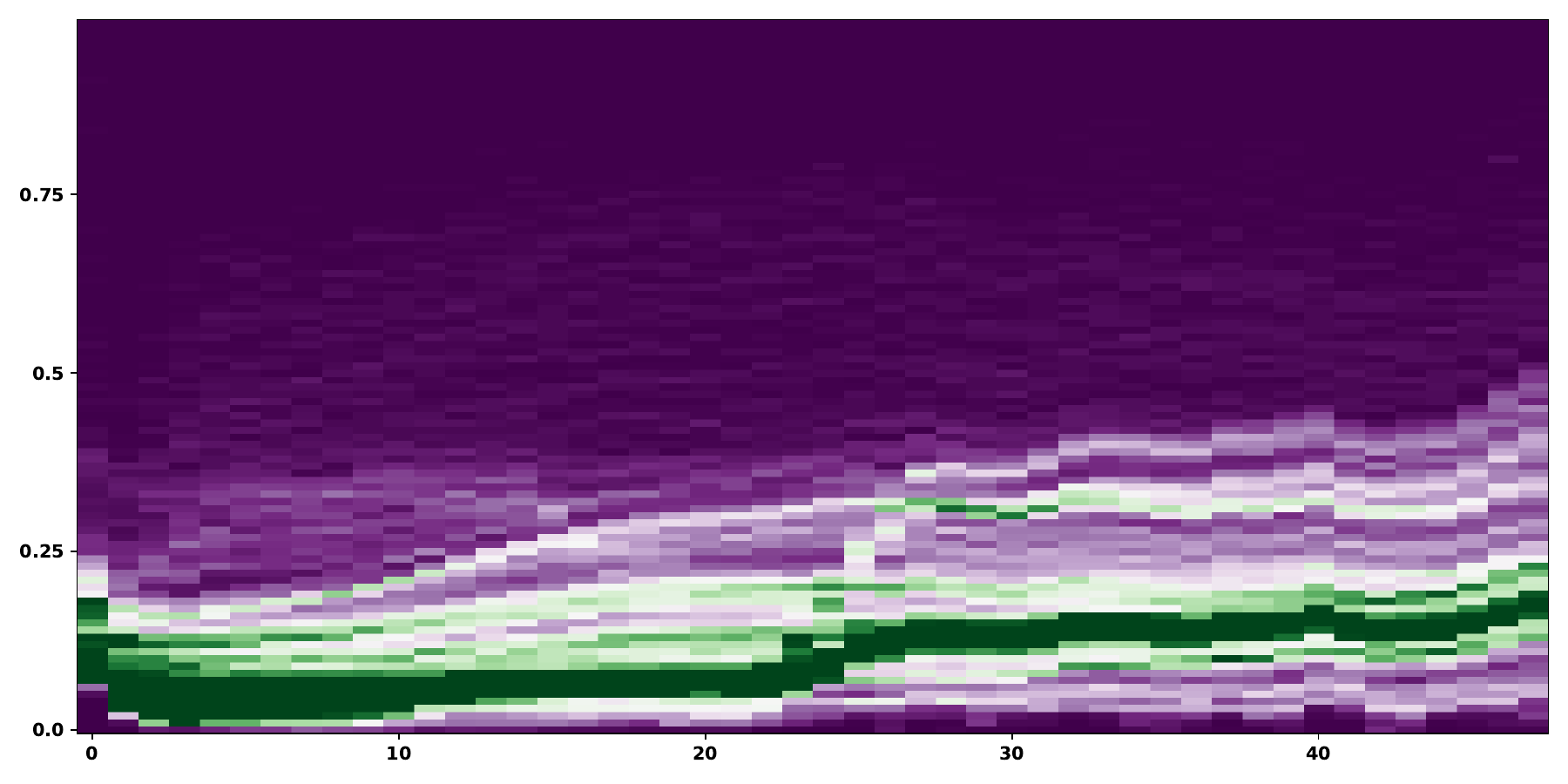}};
\draw[red, very thick, rounded corners](-1.5,-0.5) rectangle (0.5,0.8);
\end{tikzpicture}
\end{minipage}
}

\caption{{Spectral curve distributions of real HSI and generated HSI. 
All labeled data are plotted. (a)--(b) IN dataset, (c)--(d) PU dataset, and (e)--(f) Houston 2018 dataset. 
Colored boxes highlight characteristic wavelength regions.}}
\label{Spectral curve distributions}
\end{figure} 

\begin{figure}[h]
\centering
\subfigure[Real]{\includegraphics[width=0.225\textwidth]{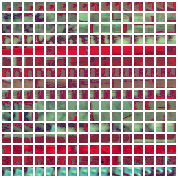}}
\subfigure[Txt2HSI-LDM(VAE)]
{\includegraphics[width=0.225\textwidth]{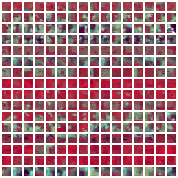}}

\subfigure[Real]{\includegraphics[width=0.225\textwidth]{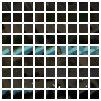}}
\subfigure[Txt2HSI-LDM(VAE)]
{\includegraphics[width=0.225\textwidth]{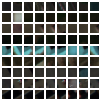}}

\caption{{Pseudo-color visualization of real HSI (left) and generated HSI (right) patches with size of \(9 \times 9\). Each row represents specific ground cover. R: 50, G: 30, B: 10.}}
\label{real_syn_images}
\end{figure}

\subsection{Classification Stage}
\subsubsection{\textbf{Expanding the Training Data}}
In this section, we combine the limited labeled data with synthetic data to train our previously proposed HSIC model SS-ConvNeXt \cite{zhuym2023}. Unlike using linear interpolation or random oversampling methods to expand the training data \cite{Weifeng2019, OZDEMIR2021114986}, in our work, the sampling process of Txt2HSI-LDM(VAE) is nonlinear and considers the varying degrees of mixing from the perspective of latent feature level and prior language description (i.e., spatial mixup), which confirm that the samples we generated will be more diverse. To better solve the unsatisfactory HSIC performance due to ISSD problem in the real world of remote sensing, one easy but effective way is generating more data for that small sample, while less data for that larger sample to balance the distribution of training data while fully considering the proportion of different categories in the labeled data. So, we introduce a scale factor named \textbf{Sample Balance Rate (SBR)} \(\lambda\). The number of generated sample \(\tilde{N}\) for specific category is as follow:
\begin{align}
    \begin{aligned} \label{EQ21}
        r(i) &= \lceil \lambda \times \frac{\text{Max}(N)}{N(i)}\rceil, i=1,..,c \\
        \tilde{N}(i) &= r(i) \times N(i)
    \end{aligned}
\end{align}
where \(N\) represents the number of each labeled data, \(c\) is the number of category, \(\lceil * \rceil\) represents rounding up to an integer. In our experiment, we set \(\lambda = 0.4\).

\subsubsection{\textbf{Effectiveness of Generated Samples}}
To prove the effectiveness when use the generated data to expand the training data, we analyze the characteristics of synthetic samples in two ways:
\subsubsubsection{\textbf{Statistical Characteristics of Generated Samples}} The error bar is used to depict the changes in mean value and standard deviation of generated samples along the spectral dimension. It should be noted that all the results are calculated according to the central pixel of the generated hyperspectral patch. As we can see from \autoref{error_IN}, with only limited labeled data when training the language-informed diffusion model, Txt2HSI-LDM(VAE) can fully take advantage of abundant unlabeled data and generate realistic samples, which can be proved from the error bar plot and the change of standard deviation along the band dimension. The changes in standard deviation on generated samples depict the band-by-band variance characteristics. The error bars of generated samples are more similar to the real change of all labeled data. {Also, in \autoref{Spectral curve distributions}, which shows the spectral curve distributions of the real HSI and the generated HSI for all the labeled data. We stack all the labeled spectral curves of a particular HSI to show
the spectral distributions. For the IN dataset, we generate the same number of (10249 samples) spectral curves with this dataset, as well as the PU dataset (42776 samples), while for the Houston dataset, since the labeled data is more than these two datasets, we fractionally generate a total of 6256 samples. The boundary shows whether the generated spectral curves are consistent with the real HSI, and the density (color) infers the distribution of spectral curves. As shown in \autoref{Spectral curve distributions}, the proposed Txt2HSI-LDM(VAE) generates high-quality spectral curves that have a similar distribution to the real one, also the local pattern, for example, the middle dark green part on the IN dataset. \autoref{point_fidelity} shows the maximum and minimum value of cosine similarity between real patches and generated patches \cite{shen2025hyperspectral}. Txt2HSI-LDM(VAE) achieves the high point fidelity,
indicating that its generated images better preserve the spectral
content of real images. Some point fidelity of landcover nearly up to 1. Meanwhile, the minimum value of some landcover types around 0.7, meaning that the spectral diversity raised by spatial mixing contributes to more diverous synthestic samples.}

\begin{figure*}[]
\centering
\subfigure[]{\includegraphics[width=0.16\textwidth]{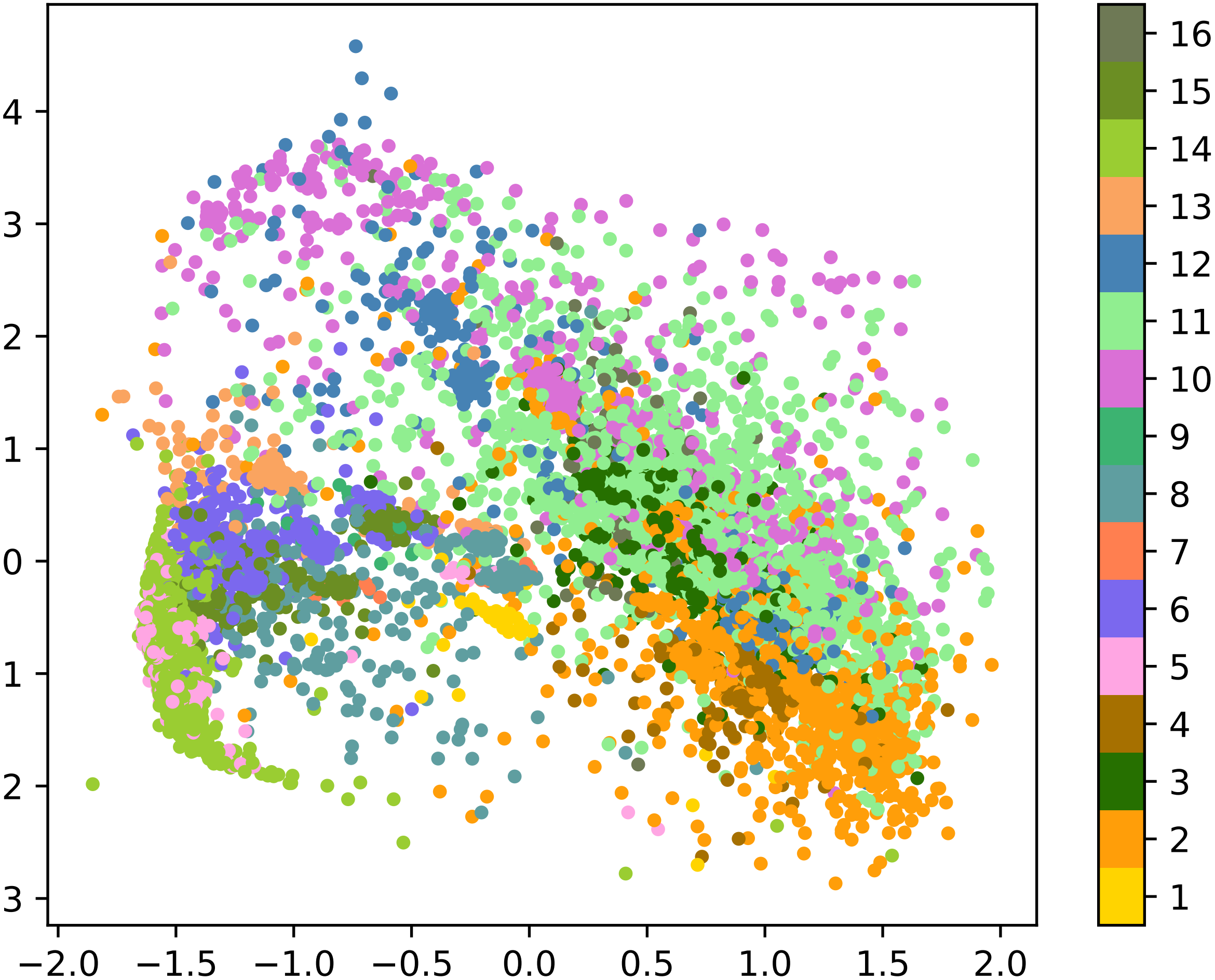}}
\subfigure[]
{\includegraphics[width=0.16\textwidth]{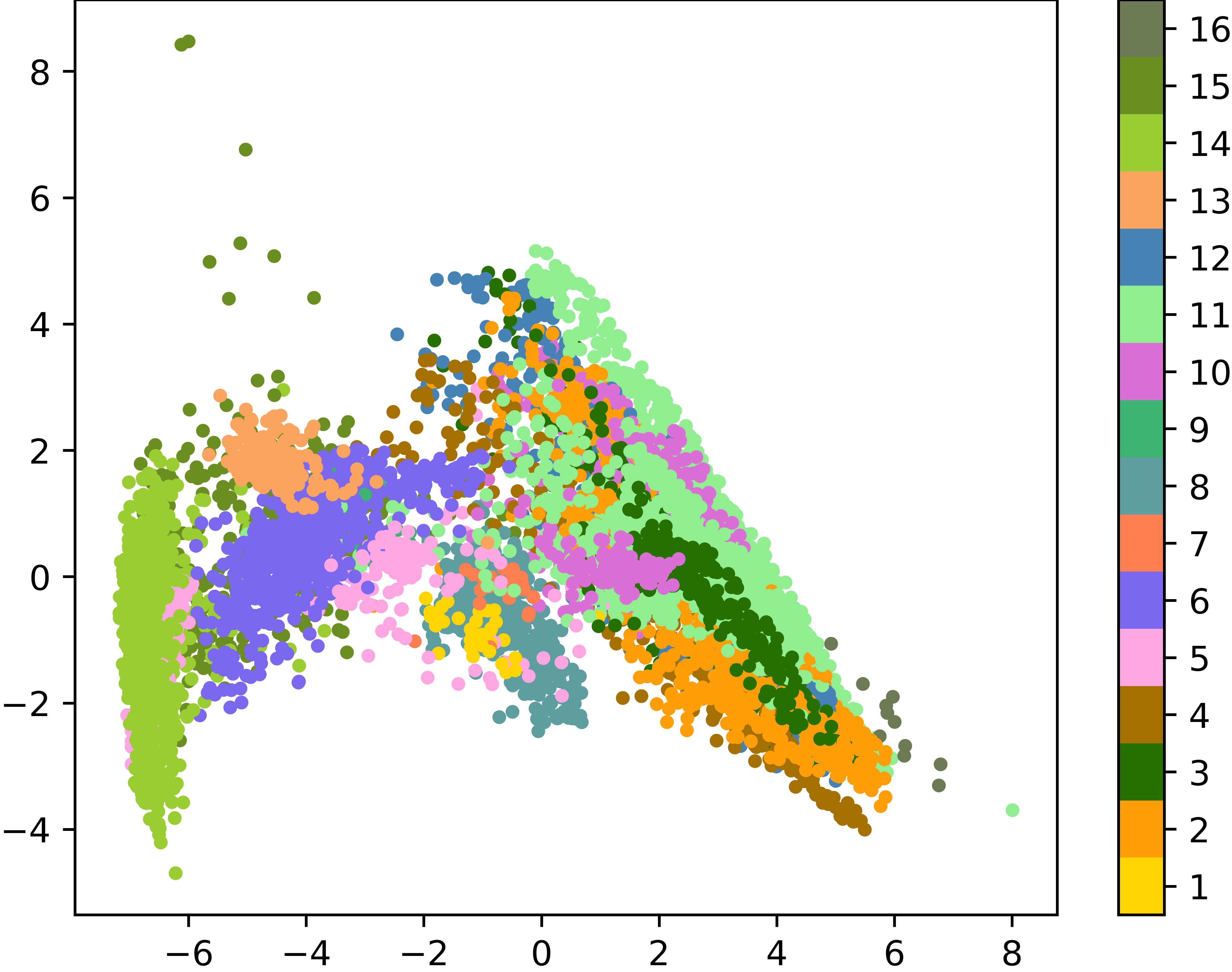}}
\subfigure[]{\includegraphics[width=0.16\textwidth]{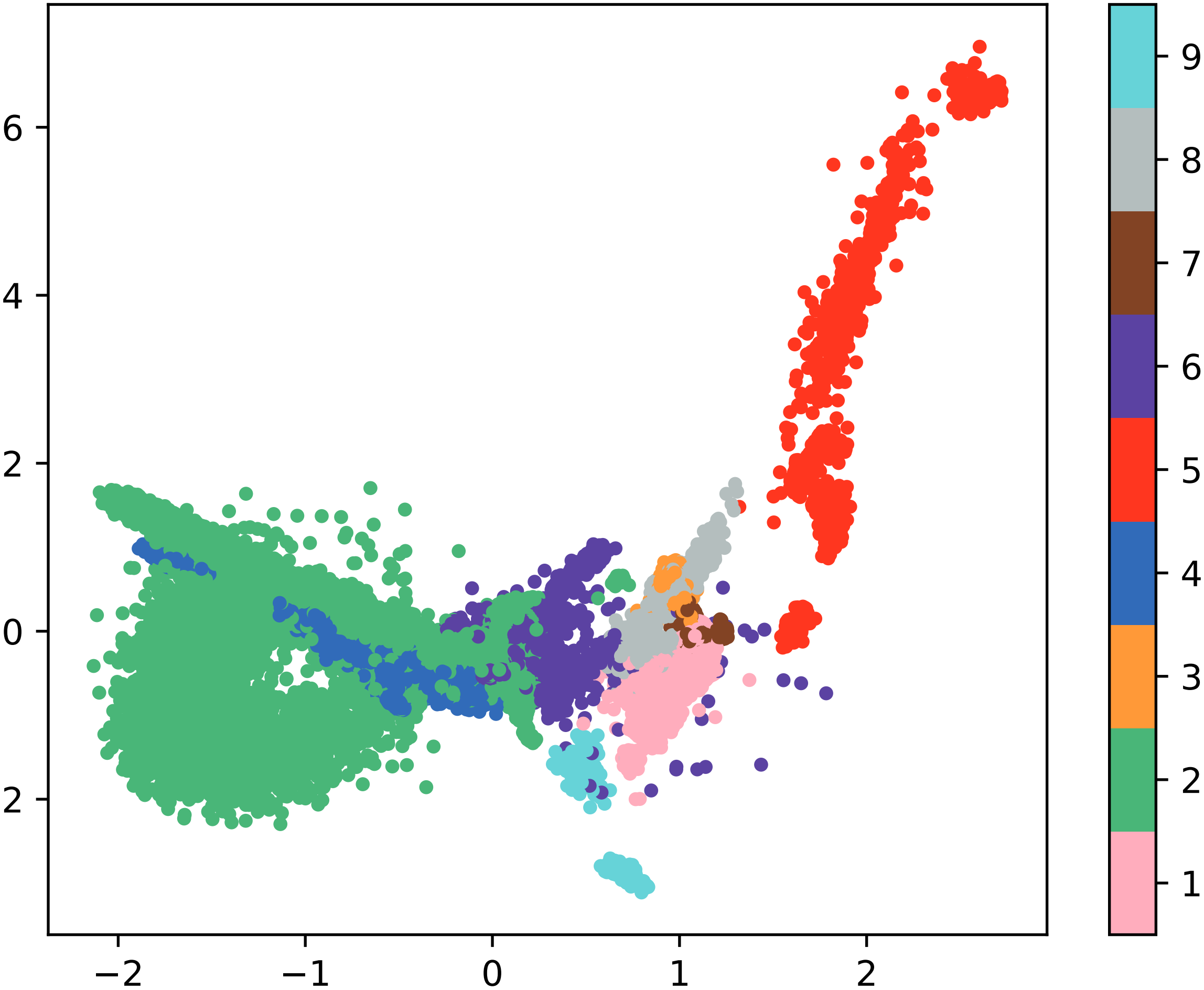}}
\subfigure[]
{\includegraphics[width=0.16\textwidth]{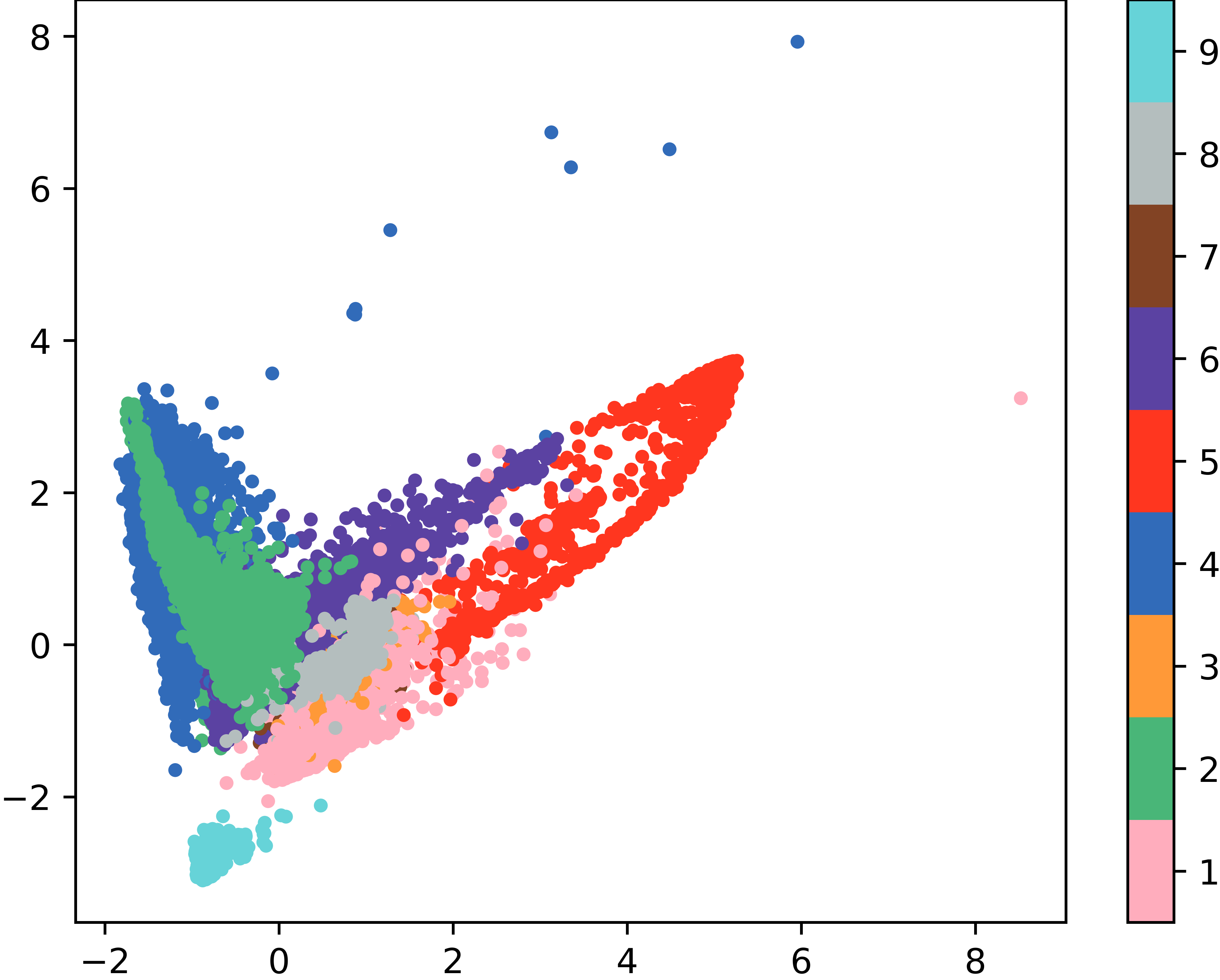}}
\subfigure[]
{\includegraphics[width=0.16\textwidth]{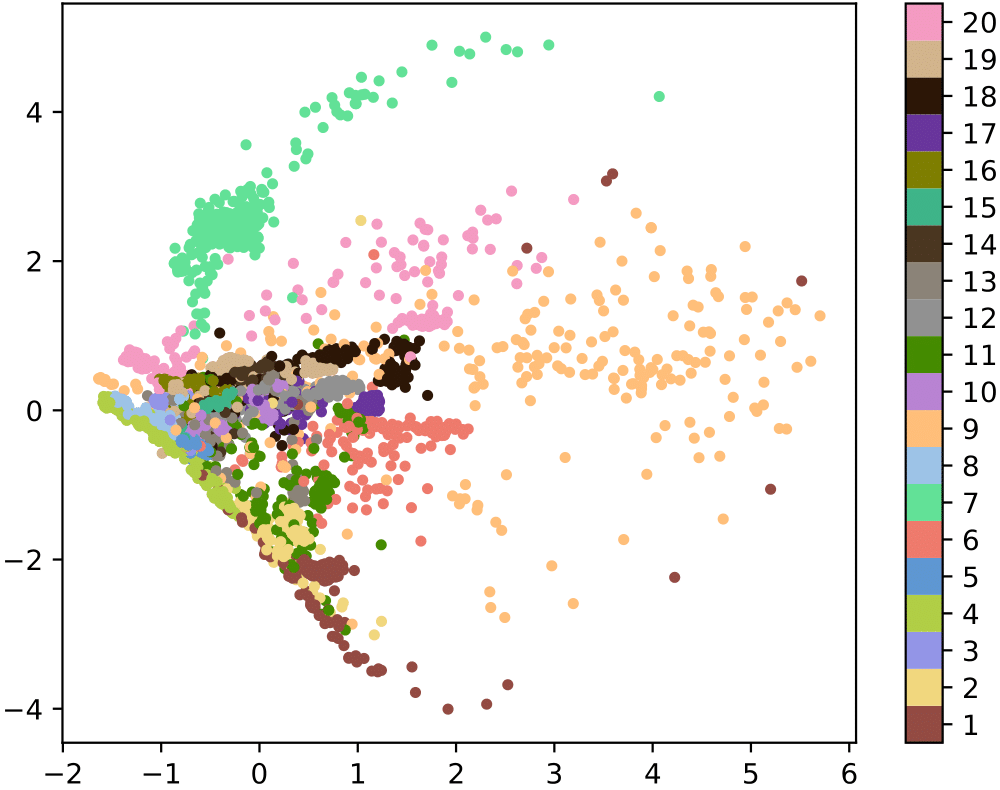}}
\subfigure[]
{\includegraphics[width=0.16\textwidth]{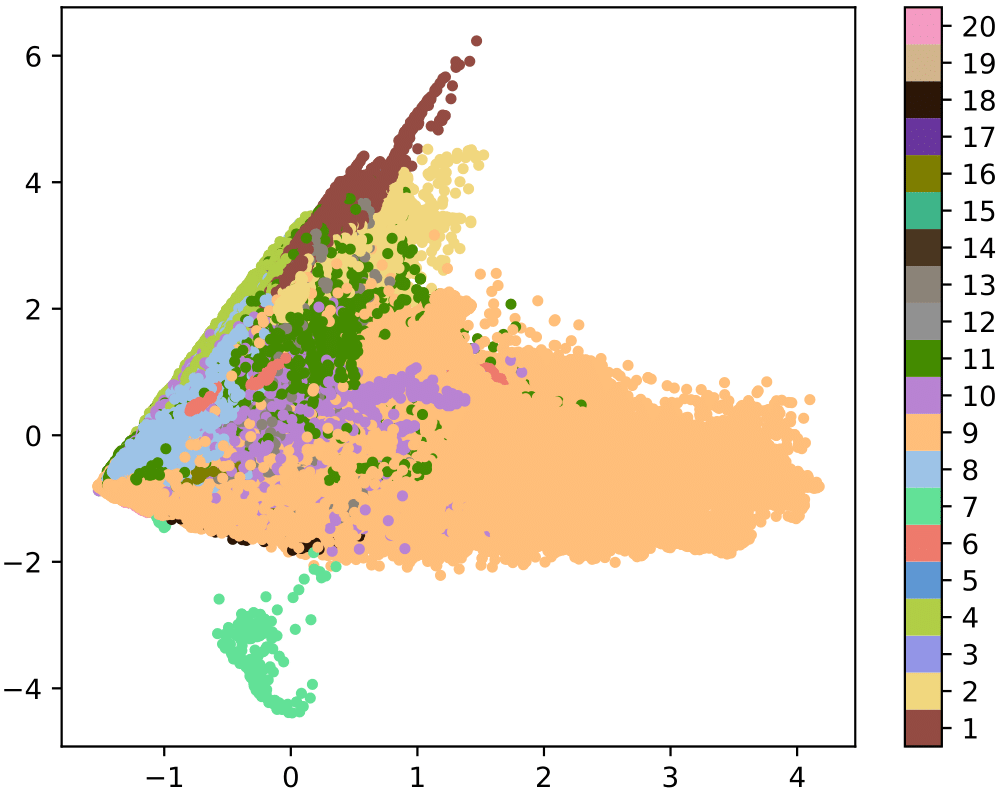}}

\caption{{Data Distribution of the central pixel of the generated sample in PCA space on the Indian Pines and Pavia University dataset. Different colors represent different categories. (a) PCA feature on the generated samples on Indian Pines (10249 samples). (b) PCA feature on all labeled data on Indian Pines. (c) PCA feature on the generated samples at the Pavia University (42766 samples). (d) PCA feature on all labeled data at Pavia University. (e) PCA feature on the generated samples on Houston (6256 samples). (f) PCA feature on all labeled data at the Houston dataset.}}
\label{PCA_IN}
\end{figure*}

\subsubsubsection{\textbf{RGB Visualization of Synthetic Images:}} {Some randomly selected real patch samples and the generated images are shown in \autoref{real_syn_images}. Since Txt2HSI-LDM(VAE) is transformer-based model, there is no grids overlay on the synthetic image due to the downsampling. Meanwhile, the generated images not only have good spatial distribution with good color visualization, also with diversity.}
\subsubsubsection{\textbf{Data Distribution of Generated Samples in PCA Space}} {Data reduction is an effective way to analyze the data distribution. One of the popular methods is Principal Component Analysis (PCA). We reduce the central pixel of the generated patch data to 2D space. Predictably, as shown in  \autoref{PCA_IN}, when we reduce the spectrum data into 2D space using PCA, we can observe that the relative location of a specific category in latent space has a high consistency compared to the real data, indicating that the synthetic data generated by Txt2HSI-LDM(VAE) can better capture the data distribution status.}

\section{Experimental Results} \label{Results}
\subsection{Experimental Data}

\subsubsection{\textbf{Indian Pines Data}} This dataset was collected by the AVIRIS sensor over Northwestern Indiana, USA. This data consists of 145 × 145 pixels at a ground sampling distance (GSD) of 20 m and 220 spectral bands covering the wavelength range of 400–2500 nm with a 10-m spectral resolution. In the experiment, 24 water-absorption bands and noise bands were removed, and 200 bands were selected. There are 16 mainly investigated categories in this studied scene. \autoref{dataset} (a) shows the false-color map and ground-truth map. Training and testing numbers are shown in \autoref{training_num}. The fine-grained text description is shown in \autoref{language_IN}.

\subsubsection{\textbf{Pavia University Data}} 
\begin{figure*}[]
\centering
\subfigure[]{\includegraphics[width=0.32\textwidth]{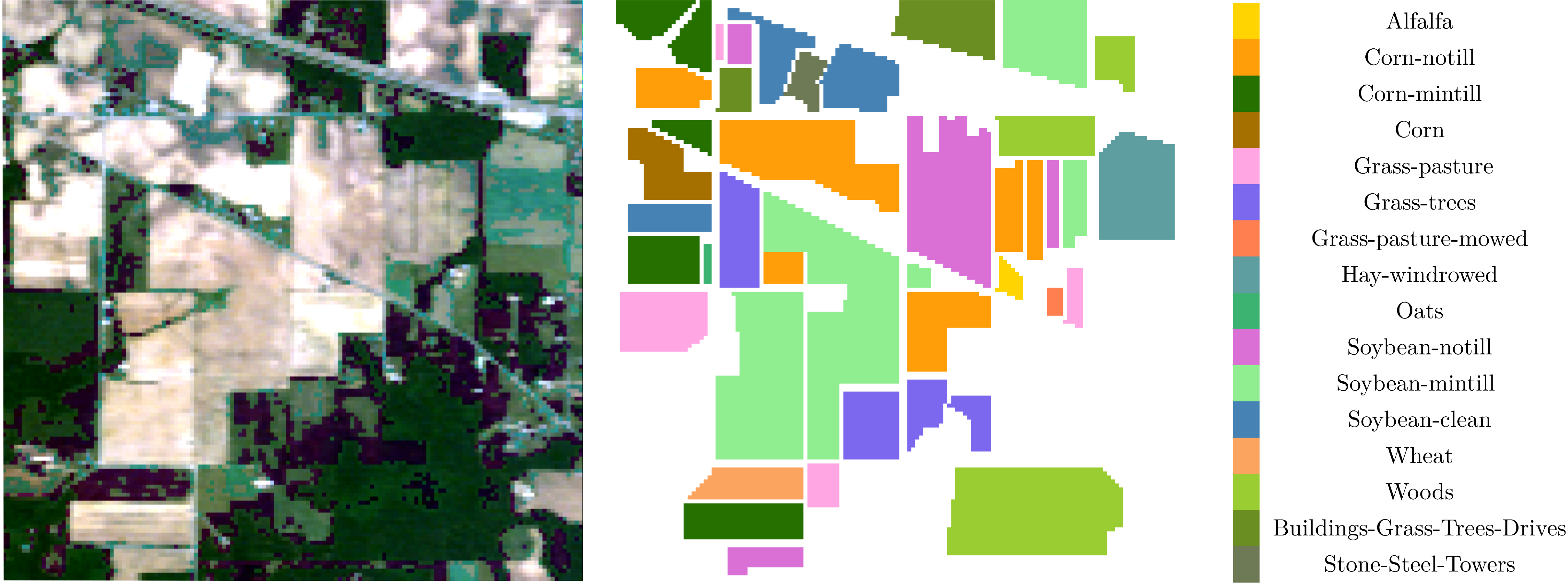}}
\subfigure[]
{\includegraphics[width=0.32\textwidth]{figures/PU_brief.pdf}}
\subfigure[]
{\includegraphics[width=0.32\textwidth]{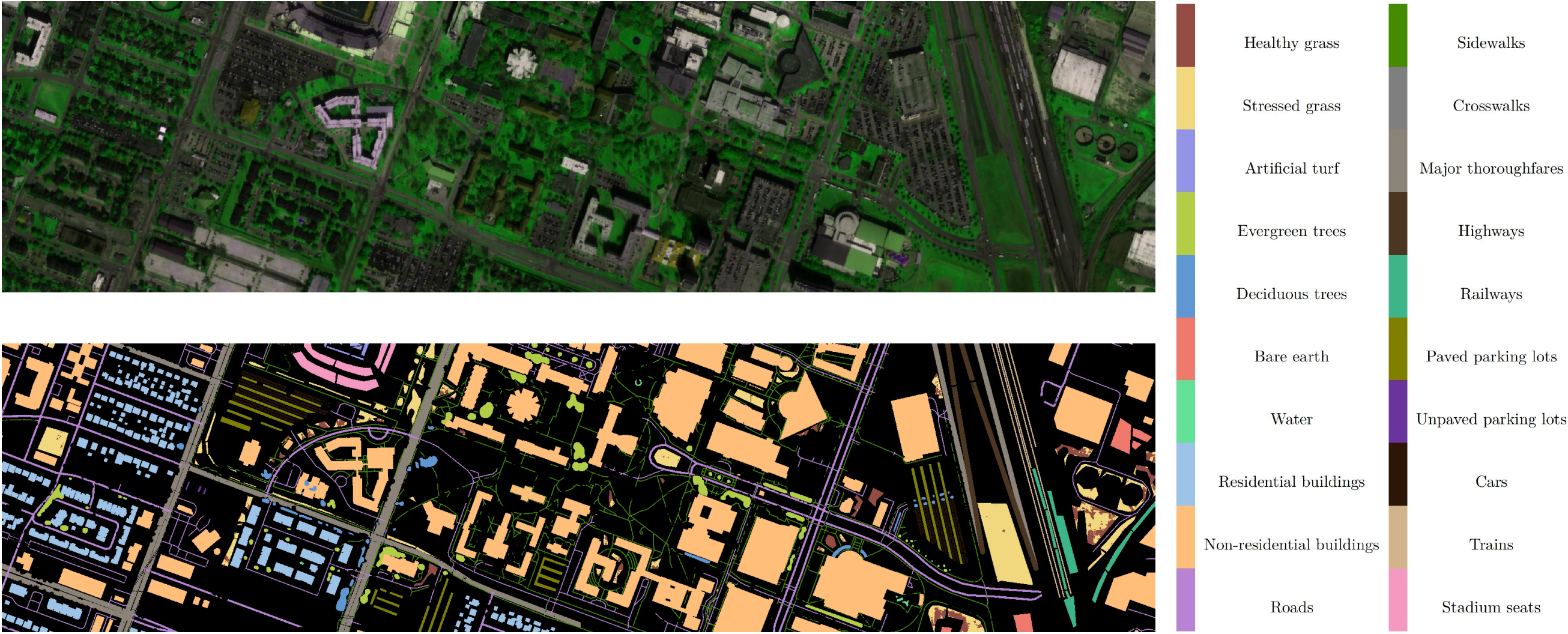}}
\caption{{The pseudocolor image with corresponding ground-truth
map and the legend of the datasets. (a) Indian Pines. (b) Pavia University. (c) Houston18.}}
\label{dataset}
\end{figure*}

It was acquired by the ROSIS sensor over Pavia University and its surroundings, Pavia, Italy. This dataset has 103 spectral bands ranging from 430 to 860 nm. Its spatial resolution is 1.3 m, and its image size is 610 × 340. Nine land-cover categories are covered.  \autoref{dataset} (b) shows the false-color map and ground-truth map. Training and testing numbers are shown in \autoref{training_num}. The fine-grained text description is shown in \autoref{language_PU}.

\subsubsubsection{\textbf{Houston 2018 Data}} {The Houston 2018 dataset was provided by the 2018 IEEE GRSS Data Fusion Contest and acquired by the National Center for Airborne Laser Mapping over the University of Houston campus and its neighborhood. The HSI covers a 380–1050 nm spectral wavelength range with 48 bands at a 1 m ground sampling distance of size 601 × 2384. The ground truth samples contained 20 classes. \autoref{dataset} (c) shows the false-color map and ground-truth map. Training and testing numbers are shown in \autoref{training_num}. The fine-grained text description is shown in \autoref{language_Houston}.}

\begin{table}[]
\centering
\caption{{Point Fidelity on each generated ground cover type.}}
\resizebox{0.49\textwidth}{!}{
\begin{tabular}{c|ccccccc}
\hline
\multirow{3}{*}{Class   No.} & \multicolumn{7}{c}{PF (max/min) $\uparrow$}                                                                                                                   \\ \cline{2-8} 
                             & \multicolumn{3}{c|}{Dataset}                                            & \multicolumn{3}{c}{Dataset}                                 &  \\ \cline{2-8} 
                             & \multicolumn{1}{c}{IN} & \multicolumn{1}{c}{PU} & \multicolumn{1}{l|}{Houston} & Class No. & \multicolumn{1}{c}{IN} & \multicolumn{1}{c}{PU} &  Houston\\ \hline
1                            & 0.9865/0.8279          &    0.9984/0.5734            & \multicolumn{1}{l|}{0.9978/0.5128} & 11        &  0.9585/0.9585        &           -             & 0.8961/0.8961 \\
2                            & 0.9975/0.6190          &    0.9994/0.7207            & \multicolumn{1}{l|}{0.9996/0.6823} & 12        &  0.9543/0.9543        &           -             & 0.9794/0.9794 \\
3                            & 0.9974/0.7478          &    0.9812/0.9812            & \multicolumn{1}{l|}{0.9989/0.9110} & 13        &  0.8968/0.8968        &           -             & 0.9582/0.9582 \\
4                            & 0.9949/0.8200          &    0.7361/0.7361            & \multicolumn{1}{l|}{0.9991/0.6924} & 14        &  0.7652/0.7652        &           -             & 0.9634/0.9634 \\
5                            & 0.8137/0.8137          &    0.9552/0.9552            & \multicolumn{1}{l|}{0.9994/0.7343} & 15        &  0.8499/0.8499        &           -             & 0.9587/0.9587 \\
6                            & 0.8831/0.8831          &    0.8994/0.8994            & \multicolumn{1}{l|}{0.9994/0.7972} & 16        &  0.9620/0.9620        &           -             & 0.9796/0.9796 \\
7                            & 0.8643/0.8643          &    0.9781/0.9781            & \multicolumn{1}{l|}{0.9503/0.4887} & 17        &        -               &          -              & 0.9821/0.9821 \\
8                            & 0.8815/0.8815          &    0.9796/0.9796            & \multicolumn{1}{l|}{0.9995/0.7588} & 18        &        -               &          -              & 0.9737/0.9737 \\ 
9                            & 0.9250/0.9250          &    0.9364/0.9364            & \multicolumn{1}{l|}{0.9994/0.4242} & 19        &        -               &           -             & 0.9780/0.9780 \\ 
10                           & 0.9697/0.9697          &          -                   & \multicolumn{1}{l|}{0.9571/0.9571} & 20        &       -                &         -               & 0.9670/0.9670 \\ \hline

\end{tabular}
}
\label{point_fidelity}
\end{table}

\begin{table}[]
\centering
\caption{{LAND-COVER TYPES, TRAINING AND TESTING SAMPLE NUMBERS OF THREE USED DATASETS.}}
\resizebox{0.49\textwidth}{!}{
\begin{threeparttable}
\begin{tabular}{cccc|ccc|ccc}
\hline
\multirow{2}{*}{Class No.} & \multicolumn{2}{c}{Indian Pines} & \multicolumn{2}{c}{Pavia University} & \multicolumn{2}{c}{Houston 2018}\\ \cline{2-10} 
                           & Color  & Training Sample  & Testing Sample & Color & Training Sample   & Testing Sample   & Color& Training Sample      & Testing Sample\\ \hline
1                   &  \cellcolor[RGB]{255, 212, 0}     & 3               & 43          & \cellcolor[RGB]{255, 173, 189}  & 15                & 6616      &  \cellcolor[RGB]{147,75,67}     & 3     & 9555 \\
2                   & {\cellcolor[RGB]{255, 158, 9}}      & 12              & 1416        & \cellcolor[RGB]{73, 182, 120}  & 43                & 18606     &    \cellcolor[RGB]{241,215,126}   &  3    & 32101 \\
3                   &  {\cellcolor[RGB]{38, 112, 0}}     & 8               & 822         &  \cellcolor[RGB]{255, 153, 56} & 5                 & 2094      &     \cellcolor[RGB]{147,148,231}  & 3     & 681 \\
4                   &  {\cellcolor[RGB]{166, 112, 0}}     & 2               & 235         &  \cellcolor[RGB]{49, 107, 185}   & 7                 & 3057      &    \cellcolor[RGB]{177,206,70}   & 3     & 13585 \\
5                   & {\cellcolor[RGB]{255, 166, 227}}      & 4               & 479         &  \cellcolor[RGB]{255, 54, 31}   & 4                 & 1341      &   \cellcolor[RGB]{95,151,210}    & 3     & 5045\\
6                   & {\cellcolor[RGB]{123, 104, 238}}      & 7               & 723         & \cellcolor[RGB]{91, 66, 162}  & 11                & 508       &    \cellcolor[RGB]{238,122,109}   &  3    & 4281 \\
7                   & {\cellcolor[RGB]{255, 127, 80}}      & 3               & 25          & \cellcolor[RGB]{130, 67, 36}  & 4                 & 1326       &   \cellcolor[RGB]{98,225,151}  &  3    &263 \\
8                   &  {\cellcolor[RGB]{95, 158, 160}}     & 4               & 474         &  \cellcolor[RGB]{180, 190, 190}   & 8                 & 3674       &   \cellcolor[RGB]{157,195,231}   & 3      & 39041 \\
9                   &  {\cellcolor[RGB]{60, 179, 113}}     & 3               & 17          &  \cellcolor[RGB]{102, 211, 216} & 3                 & 944       &  \cellcolor[RGB]{255,190,122}     &  39    & 220966 \\
10                  & {\cellcolor[RGB]{218, 112, 214}}      & 8               & 964         &   &                   &           &    \cellcolor[RGB]{184,131,211}   &  5    & 45242 \\
11                  & {\cellcolor[RGB]{144, 238, 144}}      & 22              & 2433        &   &                   &           &    \cellcolor[RGB]{69,139,0}   &  3    & 33684 \\
12                  &  {\cellcolor[RGB]{70, 130, 180}}     & 5               & 588         &   &                   &           &    \cellcolor[RGB]{127,127,127}   &  3    & 1513\\
13                  &  {\cellcolor[RGB]{250, 164, 96}}     & 2               & 203         &   &                   &           &   \cellcolor[RGB]{139,131,120}    &  5    & 46055 \\
14                  & {\cellcolor[RGB]{154, 205, 50}}      & 11              & 1254        &   &                   &           &   \cellcolor[RGB]{75,54,33}    &  3    & 9846 \\
15                  &  {\cellcolor[RGB]{107, 142, 35}}        & 3               & 383         &   &                   &           &  \cellcolor[RGB]{62,180,137}     &  3    & 6844 \\
16                  &  {\cellcolor[RGB]{110, 121, 85}}
      & 3               & 90          &   &                   &           &   \cellcolor[RGB]{128,128,0}    & 3     & 11396 \\
17                  &      &                 &              &  &                   &            & \cellcolor[RGB]{105,53,156}     & 3     & 146 \\
18                  &       &                 &             &   &                   &           &  \cellcolor[RGB]{44,22,8}     &  3    & 6575 \\
19                  &       &                 &             &   &                   &           &   \cellcolor[RGB]{210,180,140}    &  3    & 5362 \\
20                  &       &                 &             &   &                   &           &    \cellcolor[RGB]{244,154,194}   & 3      & 6821 \\ \hline
Total               &       & 100             & 10149       &   & 100               & 42676       &     & 100      & 499102 \\ \hline
\end{tabular}
\end{threeparttable}
}
\label{training_num}
\end{table}
% \usepackage{multirow}

\begin{table}[]
\centering
\caption{FINE GRAINED TEXT DESCRIPTIONS FOR INDIAN PINES DATASET. RED WORDS REPRESENT THE CATEGORY.}
\resizebox{0.49\textwidth}{!}{
\begin{threeparttable}
\begin{tabular}{cccc}
\hline
Class No. & Fine Grained Text                                                                                             & Class No. & Fine-Grained Text                                                                                                      \\ \hline
1         & {Alfalfa} occupies a small area and appears light green                                                         & 9         & \begin{tabular}[c]{@{}c@{}}The shape of {oats} field is rectangle and \\ has small planting areas\end{tabular}           \\
2         & \begin{tabular}[c]{@{}c@{}} {Corn notill} appears as a large block and has \\ three farming modes\end{tabular}     & 10        & The {soybean notill} fields have not been ploughed                                                                              \\
3         & The {corn mintill} is next to corn notill                                                                   & 11        & \begin{tabular}[c]{@{}c@{}}The {soybean mintill} occupies the largest area \\ and has been lightly ploughed\end{tabular} \\
4         & \begin{tabular}[c]{@{}c@{}}The {corn} is next to corn mintill and \\ appears irregular in shape\end{tabular}       & 12        & \begin{tabular}[c]{@{}c@{}}The {soybean clean} is next to stone steel tower \\ and has been harvested\end{tabular}       \\
5         & \begin{tabular}[c]{@{}c@{}}The location of {grass pasture} is relatively \\ dispersed\end{tabular}              & 13        & {Wheat} is next to soybean mintill                                                                                       \\
6         & \begin{tabular}[c]{@{}c@{}}The area of {grass trees} is relatively \\ concentrated\end{tabular}                 & 14        & The {woods} appears dark green                                                                                           \\
7         & \begin{tabular}[c]{@{}c@{}} {Grass pasture mowed} is next to alfalfa \\ and occupies a small part\end{tabular} & 15        & \begin{tabular}[c]{@{}c@{}}The {building grass tress} drives is one of the \\ constructions\end{tabular}                 \\
8         & \begin{tabular}[c]{@{}c@{}}The {hay windrowed} is used as the food of \\ livestock\end{tabular}                 & 16        & \begin{tabular}[c]{@{}c@{}}The {stone steel towers} are next to soybean clean \\ and appears white in image\end{tabular}  \\ \hline
\end{tabular}
\end{threeparttable}
}
\label{language_IN}
\end{table}

\begin{table}[]
\centering
\caption{FINE GRAINED TEXT DESCRIPTIONS FOR PAVIA UNIVERSITY DATASET. RED WORDS REPRESENT THE CATEGORY}
\resizebox{0.49\textwidth}{!}{
\begin{threeparttable}
\begin{tabular}{cccc}
\hline
Class No. & Fine Grained Text                                                                                                                   & Class No. & Fine-Grained Text                                                                                      \\ \hline
1         & \begin{tabular}[c]{@{}c@{}}The {asphalt} road appears grey and black and there \\ were many trees along the asphalt road\end{tabular} & 6         & \begin{tabular}[c]{@{}c@{}}There is no grass growing on the \\ surface of {bare soil} \end{tabular}      \\
2         & {Meadows} grows sparsely covered with grass                                                                                                   & 7         & \begin{tabular}[c]{@{}c@{}}{Bitumen} is a widely used waterproof material\end{tabular}      \\
3         & {Gravel} is a material used in buildings and roads                                                                                    & 8         & \begin{tabular}[c]{@{}c@{}} {Self-blocking bricks} are usually laid \\ flat on the road\end{tabular}      \\
4         & \begin{tabular}[c]{@{}c@{}}The {trees} appear as small circular regions and grow next to \\ the road\end{tabular}                              & 9         & \begin{tabular}[c]{@{}c@{}}The {shadows} is next to the buildings \\ and appear black in color\end{tabular} \\
5         & {Painted metal sheets} are often used as roofing surfaces                                                                                &           &                                                                                                        \\ \hline
\end{tabular}
\end{threeparttable}
}
\label{language_PU}
\end{table}

% \usepackage{multirow}

\begin{table*}[]
\centering
\caption{{FINE GRAINED TEXT DESCRIPTIONS FOR Houston18 DATASET. RED WORDS REPRESENT THE CATEGORY}}
\resizebox{\textwidth}{!}{
\begin{tabular}{cccc}
\hline
Class No. & Fine Grained Text                                                                                                                                                                                          & Class No. & Fine Grained Text                                                                                                                                                                                \\ \hline
1         & \begin{tabular}[c]{@{}c@{}}{Healthy grass} is evenly distributed near the building \\ and appears green, and it   is often adjacent to the sidewalk\end{tabular}                                           & 11        & {Sidewalks} are kind of road and are narrower than roads                                                                                                                                        \\
2         & \begin{tabular}[c]{@{}c@{}}The  {stressed grass} is not healthy grass and \\ appears light yellow, growing   sparsely on the soil.\end{tabular}                                                             & 12        & \begin{tabular}[c]{@{}c@{}} {Crosswalks}   are often in the shape of straight lines \\ and connected with roads and major   throughfares\end{tabular}                                               \\
3         & \begin{tabular}[c]{@{}c@{}} {Artificial turf} is a kind of material that is \\ used for building decoration and   stadiums, occupying large areas\end{tabular}                                             & 13        & \begin{tabular}[c]{@{}c@{}} {Major throughfares} are much wider than roads and run through the entire area,   \\ forming a vertically intersecting shape\end{tabular}                             \\
4         & \begin{tabular}[c]{@{}c@{}} {Evergreen trees} usually grow near the buildings and sidewalks, \\ appearing dark green with circular tree canopies\end{tabular}                                           & 14        & {Highways} are often next to the major throughfares and in the shape of straight lines                                                                                                           \\
5         & \begin{tabular}[c]{@{}c@{}} {Deciduous   trees} have a similar planting distribution as evergreen trees, \\ but leaves will   show yellow and green seasonally\end{tabular}                                    & 15        & \begin{tabular}[c]{@{}c@{}} {Railways}   are usually adjacent to the major throughfares \\ and non-residential areas,   appearing white and grey color\end{tabular}                                 \\
6         & {Bare earth} shows brown color and may grows sparse grass                                                                                                                                                    & 16        & \begin{tabular}[c]{@{}c@{}} {Paved   parking lots} are in the shape of straight lines, and many cars are   \\ parked above and near the lots, appearing darker color than unpaved lots\end{tabular} \\
7         & {Water} body   has an irregular shape and is located around the vegetation and buildings                                                                                                                     & 17        & \begin{tabular}[c]{@{}c@{}} {Unpaved   parking lots} are under construction and \\ located near the throughfares,   showing brighter color\end{tabular}                                             \\
8         & \begin{tabular}[c]{@{}c@{}} {Residential   buildings} are often in the form of regular polygons \\ and are densely clustered   together, usually close to sidewalks and roads\end{tabular}                    & 18        & \begin{tabular}[c]{@{}c@{}} {Cars}   are often in the shape of rectangle and located near the parking lots,   \\ appearing in a densely distributed form\end{tabular}                               \\
9         & \begin{tabular}[c]{@{}c@{}} {Non-residential   buildings} are usually surrounded by trees and have a similar distribution   \\ as residential areas, the roof material usually has a bright color\end{tabular} & 19        & {Trains} are   on the railways with the shape of lines, consisting of many rectangular   carriages                                                                                                 \\
10        & \begin{tabular}[c]{@{}c@{}}The {roads}   are usually in the shape of straight lines or curves, \\ connecting the   buildings and having trees on both sides\end{tabular}                                     & 20        & {Stadium   seats} are arranged in an overall curved pattern and are closely packed                                                                                                                \\ \hline
\end{tabular}
}
\label{language_Houston}
\end{table*}

\subsection{Experimental Setting}
\subsubsection{\textbf{Implementation and Training Details}}
Our proposed Txt2HSI-LDM(VAE) is implemented on the PyTorch 1.10.2 platform using a NVIDIA RTX A6000 Ada Generation with 48GB of VRAM. In the pre-training of VAE, we set epoch 4000, batch size 64, and patch size 9 for all datasets. The epoch and batch size of pre-training of semi-supervised language-informed diffusion model is 2000 and 64 respectively. The AdamW optimizer is adopted with a learning rate of 1e-4. In the classification stage, we choose our previous model SS-ConvNeXt \cite{zhuym2023} with 400 epochs.  We calculate the results by averaging the results of ten repeated experiments with 100 training samples selected randomly. 
\subsubsection{\textbf{Comparing with Other Methods}}
To compare the effectiveness of the proposed Txt2HSI-LDM(VAE), several representative algorithms are selected for the control experiments, including conventional classifier: SVM, CNNs-based networks: 2D-CNN \cite{chenys2016}, SSRN \cite{zhongzl2018}, SS-ConvNeXt \cite{zhuym2023}, semi-supervised model: DFRes-CR \cite{wangyx2021}, and self-ensembling network: RSEN \cite{Xuyh2024}. {Three diffuison-model-based methods, DKDMN \cite{DKDMN}, DEMAE \cite{DEMAE}, and DiffusionAAE \cite{DiffusionAAE}. A linguistic-visual feature alignment method LDGNet \cite{zhang-LDGNet2023} is also included. Since LDGNet is propsosed for cross-domain classification, we test excluded the target domain dataset, train and test on the source dataset instead.}

\begin{itemize} 
    \item [a)] \textbf{DFRes-CR:} {An end-to-end semi-supervised full-convolutional classification model is developed using probability bars for class prediction. The component random field selects similar pixels in the neighborhood of the center pixel to generate pseudo-labels, which are then used to expand the training dataset. }

    \item [b)] \textbf{RSEN:} {A semi-supervised learning model consists of a base network and an ensemble network. The model jointly constrains the supervised loss of labeled data and the consistency loss of unlabeled data to update model parameters. Consistency filters are also employed to filter high-quality unlabeled data.}

    \item [c)]\textbf{DEMAE:} {A two-stage model, combines Masked autoencoder (MAE) and diffusion model to capture the spatial correlation as well as the data distribution under the condition of category label.}

    \item [d)] \textbf{DKDMN:} {This model establishes two views, one this the original hypersepctral patches, another is the reconstructed land-cover distribution knowledge, and uses contrastive learning to enhance the network's cross-sample awareness ability.}

    \item [e)] \textbf{DiffusionAAE:} {This model incorporates spectral similarity constraints and class label guidance into the diffusion process, ensuring the generation of physically realistic synthetic samples, which shows remarkable performance on minority classes. The model is trained in GAN-style.}

    \item [f)] \textbf{LDGNet:} {This model uses a dual-stream architecture to extract two levels (coarse-grained and fine-grained) of linguistic features, which are used as cross-domain shared semantic space. The visual–linguistic alignment is completed by supervised contrastive learning in semantic space. Since this method is proposed for Cross-Scene HSIC, we set the source and target scenes as the same scene.}
\end{itemize}

To quantitatively evaluate the proposed method and other compared methods, we choose the following commonly used metrics, i.e., Overall Classification Accuracy (OA \(\uparrow\)), Average Classification Accuracy (AA \(\uparrow\)), Category Accuracy (CA \(\uparrow\)), and Kappa Coefficient (\(\kappa\) \(\uparrow\)).

\begin{figure}[]
\centering
\subfigure[]{\includegraphics[width=0.15\textwidth]{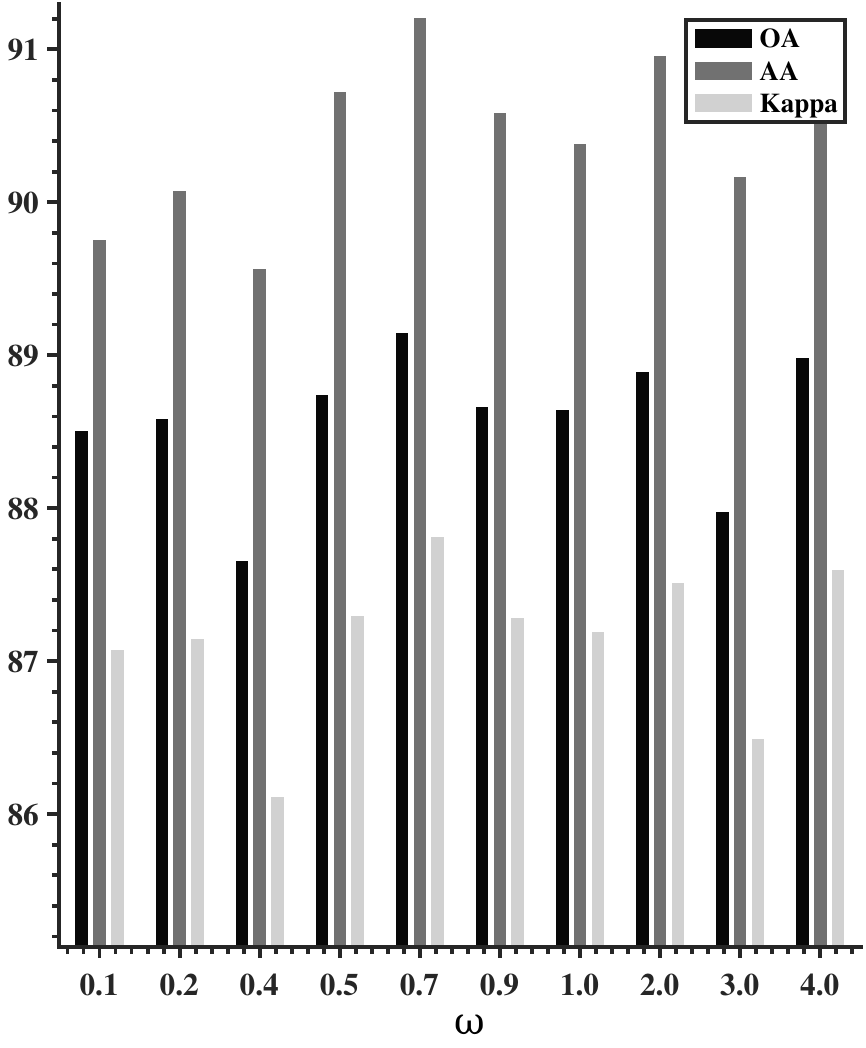}}
\subfigure[]
{\includegraphics[width=0.15\textwidth]{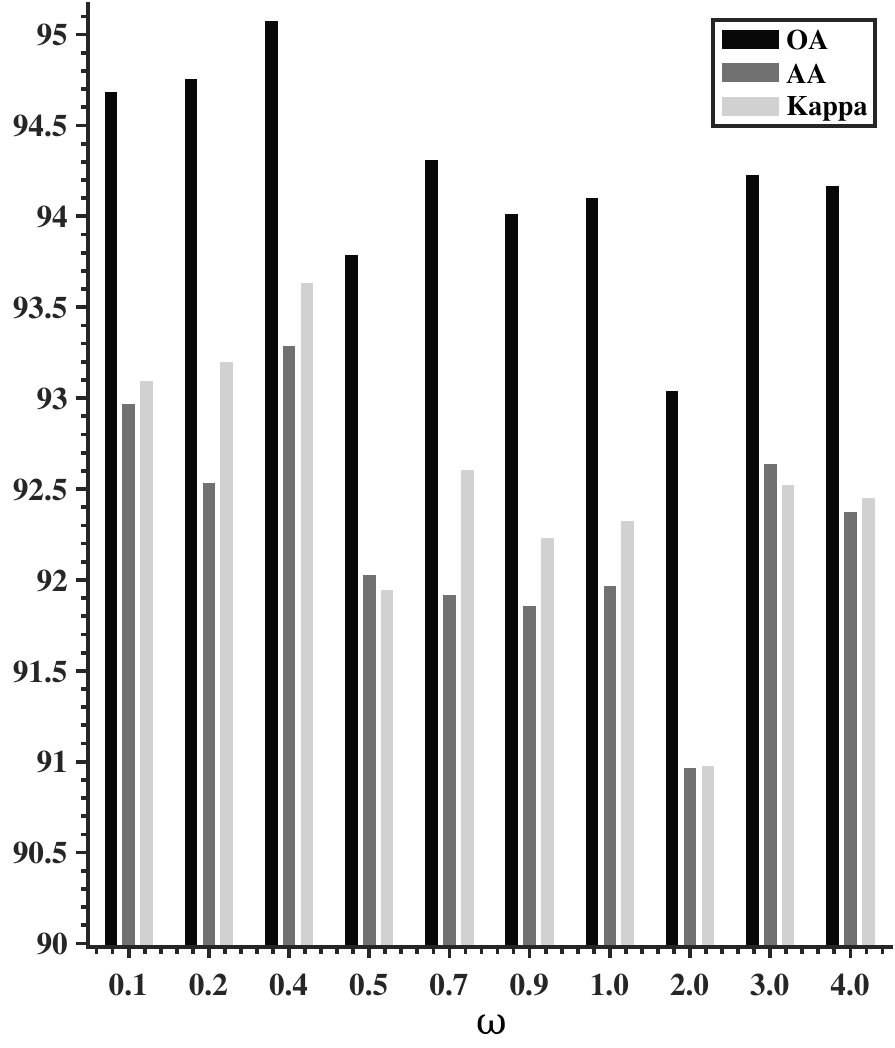}}
\subfigure[]{\includegraphics[width=0.15\textwidth]{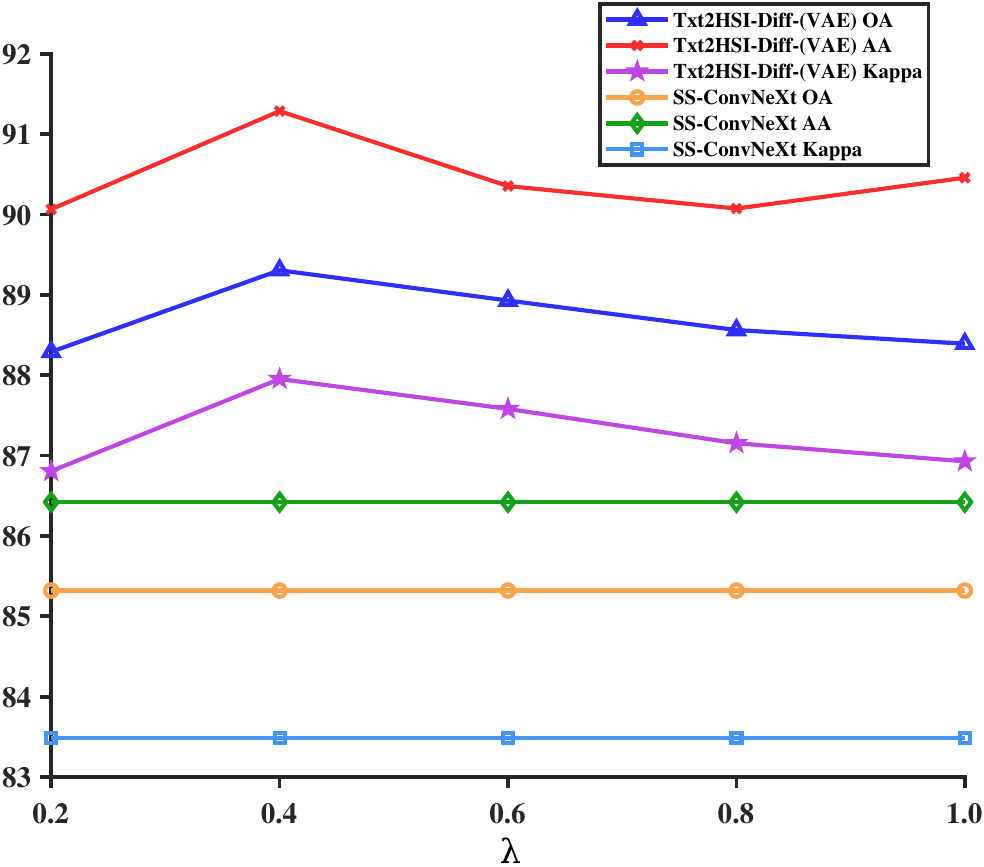}}
\caption{(a) Parameter experiment of \(\omega\) on Indian Pines dataset. (b) Parameter experiment of \(\omega\) on Pavia University dataset. (c) Parameter experiment on Sample Balance Rate \(\lambda\), results on Indian Pines dataset.}
\label{ablation_study}
\end{figure}

\subsection{{Experiments on Parameter Setting and Synthetic Samples Effectiveness}}
\subsubsection{\textbf{Parameter Selection on Guidance Coefficient \(\omega\)}}
For the generative model, some related researchers find that more diverse and realistic samples used to expand the limited training sample will boost the classification performance \cite{lomurno2024stablediffusiondatasetgeneration, chenxq2023, azizi2023syntheticdatadiffusionmodels}. To confirm this and get better \(\omega\), we search for the most suitable guide coefficient from 0.1 to 4.0. As shown in \autoref{ablation_study} (a), (b), generally, the precision index shows the trend of first rising and then decreasing. For the Indian Pines dataset, when \(\omega=0.7\) achieves the best results, while \(\omega=2.0\) ranks the second one. For the Pavia Univesity dataset, the best classification performance appears at \(\omega=0.4\), and then \(\omega=0.2\).

\subsubsection{\textbf{Parameter Selection on Sample Balance Rate \(\lambda\)}}
Imbalanced data annotation is supposed to be a prevalent problem in the remote sensing community. Imbalanced-Small Sample Data (ISSD) HSIC is the main focus of this work. For better expanding and balancing the training samples, the Sample Balance Rate \(\lambda\) is viewed as a simple rule to control the generated number of specific categories based on the sampling frequency. Some related results are shown in \autoref{ablation_study} (c), compared to the baseline model SS-ConvNeXt, classification performance after expanding training samples gets a big promotion. Even with little \(\lambda\), in other words, adding fewer generated data will also boost the performance. When \(\lambda=0.4\), three precision indexes show the highest level. The accuracy obtained by expanding samples outperforms the SS-ConvNeXt by 3.98\%, 4.86\%, and 4.46\% in terms of OA, AA, and \(\kappa\) respectively. Samples after balancing contribute more to the Average Classification Accuracy, alleviating the ISSD problem in HSIC.

\subsubsection{\textbf{Improvement on other classification methods with generated samples}} To verify that the generated data can also enhance the classification performance of other methods, we perform experiments on SVM, 2D-CNN, SSRN, and SS-ConvNeXt using 100 real training samples with samples generated by Txt2HSI-LDM(VAE) with \(\lambda\) of 0.4 and \(\omega\) of 0.7. The improvement in classification performance is presented in \autoref{adding2other} on the Indian Pines dataset. The red font in \autoref{adding2other} indicates the improvement. {By adding more generated samples, alleviating the data scarcity brought by the ISSD problem, making \(4.79\%\) improvement in AA on SS-ConvNeXt.} 2D-CNN and SSRN model enhance the richness of each category's spatial-spectral features due to the addition of generated data. This allows the data to better represent the entire data distribution, thereby improving classification performance. The improvement of 2D-CNN by 30.17\% in terms of OA demonstrates that the generated data possesses both authenticity and diversity. For the traditional method SVM, the classification accuracy was not significantly improved, with only a 2.43\% increase in terms of OA. It is likely that the generated data do not satisfactorily capture spectral details and variations, making it difficult to improve a classifier that relies solely on spectral information.

\subsubsection{\textbf{Impact of different granularity of text descriptions}} In this section, we discuss the level of classification performance contribution on fine-grained and coarse-grained text descriptions. There are two kinds of text designs listed in \autoref{txt}. The 'No text' means directly using SS-ConvNeXt to conduct HSIC without any expanded training sample, while 'Coarse-grained' and 'Fine-grained' represent using template and the manually designed text description to generate hyperspectral patches. As we can see from the \autoref{text_grained}, not only 'Fine-grained' but also 'Coarse-grained' will boost the classification performance, and outperform the baseline model by 3.06\%, 3.84\% in terms of OA, suggesting that linguistic modality enhances the learning of visual representations, and
both coarse and fine-grained visual–linguistic similarity calculation via cross attention can help improve the authenticity of synthetic samples. Additionally, well-designed text improves by around 1\% than coarse-grained one, indicating that fine-grained language boosts visual–linguistic alignment.

\begin{table}[]
\centering
\caption{{Increase of classification accuracy after adding Txt2HSI-LDM(VAE) samples to different baseline models. Results on Indian Pines with 100 raw labeled data}}
\resizebox{0.49\textwidth}{!}{
\begin{threeparttable}
\begin{tabular}{ccccc}
\hline
\multicolumn{1}{c}{\multirow{2}{*}{}} & \multicolumn{4}{c}{Methods}                  \\ \cline{2-5} 
\multicolumn{1}{c}{}                  & SVM          & 2D-CNN        & SSRN    & SS-ConvNeXt       \\ \hline
OA                                    & 54.51({+2.43}) & 83.66({+30.17}) & 80.63({+15.29}) & 89.16({+3.84})\\
AA                                    & 57.36({+4.65}) & 87.92({+32.81}) & 84.01({+15.72})  & 91.21({+4.79}) \\
\(\kappa\)                                 & 48.65({+3.43}) & 81.63({+44.72}) & 78.05({+17.53}) & 87.83({+4.34}) \\ \hline
\end{tabular}
\end{threeparttable}
}
\label{adding2other}
\end{table}

\begin{table}[]
\centering
\caption{The influence of different granularity text guidance signals on classification accuracy, taking Indian Pines as an example}
\resizebox{0.49\textwidth}{!}{
\begin{threeparttable}
\begin{tabular}{ccccc}
\hline
      & No text    & Machine generated & Coarse-grained & Fine-grained \\ \hline
OA    & 85.32±3.10 & 87.84±0.93 & 88.38±1.73     & \cellcolor[RGB]{251, 228, 213}89.16±\textbf{0.91}   \\
AA    & 86.42±7.58 & 90.33±1.66  & 90.25±1.39     & \cellcolor[RGB]{251, 228, 213}91.21±\textbf{1.06}   \\
\(\kappa\) & 83.49±3.48 & 86.18±1.06 & 86.97±1.94     & \cellcolor[RGB]{251, 228, 213}87.83±\textbf{1.02}   \\ \hline
\end{tabular}
\end{threeparttable}
}
\label{text_grained}
\end{table}

% \usepackage{multirow}

\subsection{Numerical Evaluation on HSIC}

% \usepackage{multirow}
\begin{table*}[]
\centering
\caption{QUANTITATIVE PERFORMANCE OF DIFFERENT CLASSIFICATION METHODS IN TERMS OF OA, AA, \(\kappa\), AS WELL AS THE ACCURACIES FOR EACH CLASS ON THE INDIAN PINES DATASET. THE BEST RESULTS ARE COLORED SHADOW AND OPTIMAL STANDARD DEVIATION IN BOLD}
\resizebox{\textwidth}{!}{
\begin{threeparttable}
\begin{tabular}{c|c|cc|cc|cc|ccccc}
\hline
\multirow{2}{*}{\makecell[c]{Class \\ No.}} &   
\multirow{2}{*}{Color} & 
\multicolumn{2}{c|}{Conventional Classifier} &
\multicolumn{2}{c|}{CNNs Based Networks} &
\multicolumn{2}{c|}{Semi-supervised Based Networks} &
\multicolumn{3}{c|}{Diffusion Model Based Networks} &
\multicolumn{2}{c}{Language-informed Networks}                                    \\ \cline{3-13}
\multicolumn{1}{c|}{}                           & \multicolumn{1}{c|}{}                       & SVM         & 2D-CNN      & SSRN        & SS-ConvNeXt & DFRes-CR    & RSEN    &  DKDMN   & DEMAE   &  DiffusionAAE   & LDGNet    & \textbf{Txt2HSI-LDM(VAE)} \\ \hline \hline
\multicolumn{1}{c|}{1}                          & \multicolumn{1}{c|}{\cellcolor[RGB]{255, 212, 0}}                       & 52.09±17.86 & 62.95±21.68 & 84.59±16.16 & 86.50±16.88 & 90.91±30.15 & 85.05±20.13 &  99.53±0.93   &   99.30±1.56    &   92.09±13.56    &    11.74±14.39   & \cellcolor[RGB]{251, 228, 213}100.00±\textbf{0.00}      \\
\multicolumn{1}{c|}{2}                          & \multicolumn{1}{c|}{\cellcolor[RGB]{255, 158, 9}}                       & 42.32±5.17  & 44.12±9.33  & 54.74±12.12 & 83.87±6.77  & \cellcolor[RGB]{251, 228, 213}79.89±9.42  & 60.40±9.72   &  77.45±7.60     &    71.71±10.07   &    69.29±7.41   &  34.67±3.56    & 79.87±\textbf{2.81}       \\
\multicolumn{1}{c|}{3}                          & \multicolumn{1}{c|}{\cellcolor[RGB]{38, 112, 0}}                       & 40.01±9.12  & 24.80±9.62  & 39.73±9.98  & 76.80±10.65 & 69.49±16.64 & 42.63±13.85  &   74.08±15.47    &    71.53±9.18   &      54.01±8.81  &   38.64±7.26  & \cellcolor[RGB]{251, 228, 213}84.09±\textbf{7.46}      \\
\multicolumn{1}{c|}{4}                          & \multicolumn{1}{c|}{\cellcolor[RGB]{166, 112, 0}}                       & 14.13±5.86  & 20.46±10.15 & 39.21±16.72 & 84.11±7.08  & 74.45±18.35 & 11.25±\textbf{3.84}   &  78.63±3.08     &   64.93±19.81    &   44.89±13.96    &    47.24±7.50   & \cellcolor[RGB]{251, 228, 213}85.02±9.43       \\
\multicolumn{1}{c|}{5}                          & \multicolumn{1}{c|}{\cellcolor[RGB]{255, 166, 227}}                       & 41.21±14.81 & 23.83±14.71 & 51.36±16.80 & 70.38±6.28  & 50.07±30.99 & 45.66±15.87  &  73.77±8.87     &    77.64±7.89   &   54.09±13.10    &   52.13±5.29   & \cellcolor[RGB]{251, 228, 213}74.27±\textbf{5.34}       \\
\multicolumn{1}{c|}{6}                          & \multicolumn{1}{c|}{\cellcolor[RGB]{123, 104, 238}}                       & 84.98±8.21  & 60.61±20.06 & 87.31±7.64  & 95.72±3.34  & 97.56±3.45  & 95.30±5.07   &    95.92±2.49   &   92.83±5.91    &    70.38±6.13   &   76.72±4.75   & 95.71±1.51       \\
\multicolumn{1}{c|}{7}                          & \multicolumn{1}{c|}{\cellcolor[RGB]{255, 127, 80}}                       & 89.20±7.55  & 80.53±19.41 & 96.50±6.56  & \cellcolor[RGB]{251, 228, 213}100.00±\textbf{0.00} & 98.76±2.12  & 99.42±1.51   &   98.40±3.20    &   100.00±0.00   &    97.60±6.31   &   0.00±0.00     & \cellcolor[RGB]{251, 228, 213}100.00±\textbf{0.00}      \\
\multicolumn{1}{c|}{8}                          & \multicolumn{1}{c|}{\cellcolor[RGB]{95, 158, 160}}                       & 49.14±14.59 & 77.36±15.22 & 73.89±14.30 & 82.21±13.74 & \cellcolor[RGB]{251, 228, 213}98.31±\textbf{5.59}  & 95.54±5.69   &   88.43±12.07    &   90.71±7.74    &   82.15±18.38    &   93.05±1.48     & 91.58±13.86      \\
\multicolumn{1}{c|}{9}                          & \multicolumn{1}{c|}{\cellcolor[RGB]{60, 179, 113}}                       & 91.76±4.96  & 94.51±13.43 & 98.52±2.72  & 99.25±0.23  & 81.81±4.05  & 86.55±19.43  &  100.00±0.00     &   99.41±1.86    &    97.64±7.44   &    0.00±0.00  & \cellcolor[RGB]{251, 228, 213}100.00±\textbf{0.00}     \\
\multicolumn{1}{c|}{10}                         & \multicolumn{1}{c|}{\cellcolor[RGB]{218, 112, 214}}                       & 38.43±9.88  & 42.72±11.58 & 51.93±10.94 & 79.50±\textbf{0.57}  & 79.64±7.28  & 48.53±19.43  &   85.22±5.94    &   67.76±4.41    &    68.79±6.87   &   59.03±4.47  & \cellcolor[RGB]{251, 228, 213}88.76±7.09       \\
\multicolumn{1}{c|}{11}                         & \multicolumn{1}{c|}{\cellcolor[RGB]{144, 238, 144}}                       & 61.61±3.51  & 71.30±6.63  & 78.87±4.55  & 89.71±4.01  & \cellcolor[RGB]{251, 228, 213}94.60±3.50  & 86.87±6.19   &  85.28±4.05     &  82.52±5.75     &   74.40±2.52    &   59.03±4.47     & 93.43±\textbf{3.35}       \\
\multicolumn{1}{c|}{12}                         & \multicolumn{1}{c|}{\cellcolor[RGB]{70, 130, 180}}                       & 22.52±7.88  & 18.21±8.58  & 25.57±\textbf{5.53}  & 75.56±15.15 & 68.23±19.96 & 20.07±8.51   &    73.07±10.52   &  57.19±15.77    &    51.29±11.41   &  40.19±5.51    & \cellcolor[RGB]{251, 228, 213}86.62±9.36       \\
\multicolumn{1}{c|}{13}                         & \multicolumn{1}{c|}{\cellcolor[RGB]{250, 164, 96}}                       & 77.04±17.75 & 65.94±25.91 & 81.12±19.18 & 95.23±5.74  & \cellcolor[RGB]{251, 228, 213}99.86±0.23  & 81.02±25.51  &   96.15±4.54    &   98.07±3.40   &     86.99±10.46  &  92.18±7.33   & 99.90±\textbf{0.22}       \\
\multicolumn{1}{c|}{14}                         & \multicolumn{1}{c|}{\cellcolor[RGB]{154, 205, 50}}                       & 71.88±4.46  & 79.97±8.24  & 91.31±6.66  & 95.42±7.03  & \cellcolor[RGB]{251, 228, 213}99.51±\textbf{1.32}  & 92.47±10.50  &    98.69±6.28   &   98.11±1.35    &   86.99±4.62    &   91.98±0.35     & 96.46±1.98       \\
\multicolumn{1}{c|}{15}                         & \multicolumn{1}{c|}{\cellcolor[RGB]{107, 142, 35}}                       & 17.00±6.92  & 26.23±16.27 & 43.92±8.13  & 64.24±11.11 & 36.43±15.89 & 46.44±25.38  &   79.26±6.56    &    56.21±15.18   &    58.45±12.12   &    39.88±2.82   & \cellcolor[RGB]{251, 228, 213}83.74±\textbf{2.83}      \\
\multicolumn{1}{c|}{16}                         & \multicolumn{1}{c|}{\cellcolor[RGB]{110, 121, 85}}                       & 50.11±21.69 & 82.22±6.77  & 94.02±7.10  & 97.13±5.64  & 93.83±3.01  & 72.06±34.26  &    97.77±2.81   &   97.56±6.57    &    96.66±6.43   &    76.85±4.44   & \cellcolor[RGB]{251, 228, 213}100.00±\textbf{0.00}      \\ \hline \hline
\multicolumn{2}{c|}{OA(\%)}                                                                    & 52.08±1.91  & 53.49±2.11  & 65.34±3.91  & 85.32±3.10  & 83.86±2.39  & 68.69±1.75   &  84.58±2.90     &   79.25±2.44    &   70.51±3.24    &   63.34±1.31    & \cellcolor[RGB]{251, 228, 213}89.16±\textbf{0.91}       \\
\multicolumn{2}{c|}{AA(\%)}                                                                    & 52.71±2.12  & 55.11±29.88 & 68.29±10.30 & 86.42±7.58  & 82.09±13.02 & 67.23±2.78  &   87.60±2.60    &   82.85±1.48    &    74.20±3.52   &     51.98±30.24    & \cellcolor[RGB]{251, 228, 213}91.21±\textbf{1.06}       \\
\multicolumn{2}{c|}{\(\kappa\)(\%)}                                                                 & 45.22±2.38  & 46.91±2.69  & 60.52±4.59  & 83.49±3.48  & 81.57±2.81  & 65.53±1.97  &      82.50±3.32 &    76.34±2.76   &     66.59±3.89  &    57.82±1.61    & \cellcolor[RGB]{251, 228, 213}87.83±\textbf{1.02}       \\ \hline
\end{tabular}
\end{threeparttable}
}
\label{exp_IN}
\end{table*}

% \usepackage{multirow}
\begin{table*}[]
\centering
\caption{QUANTITATIVE PERFORMANCE OF DIFFERENT CLASSIFICATION METHODS IN TERMS OF OA, AA, \(\kappa\), AS WELL AS THE ACCURACIES FOR EACH CLASS ON THE PAVIA UNIVERSITY DATASET. THE BEST RESULTS ARE COLORED SHADOW AND OPTIMAL STANDARD DEVIATION IN BOLD}
\resizebox{\textwidth}{!}{
\begin{threeparttable}
\begin{tabular}{c|c|cc|cc|cc| cccc c}
\hline
\multirow{2}{*}{\makecell[c]{Class \\ No.}} &   
\multirow{2}{*}{Color} & 
\multicolumn{2}{c|}{Conventional Classifier} &
\multicolumn{2}{c|}{CNNs Based Networks} &
\multicolumn{2}{c|}{Semi-supervised Based Networks} &
\multicolumn{3}{c|}{Diffusion Model Based Networks} &
\multicolumn{2}{c}{Language-informed Networks}                                    \\ \cline{3-13}
\multicolumn{1}{c|}{}                           & \multicolumn{1}{c|}{}                        
& SVM         & 2D-CNN      & SSRN        & SS-ConvNeXt & DFRes-CR    & RSEN      &  DKDMN   & DEMAE   &  DiffusionAAE   & LDGNet    & \textbf{Txt2HSI-LDM(VAE)} \\ \hline \hline
\multicolumn{1}{c|}{1}                          &    \cellcolor[RGB]{255, 173, 189}                  & 74.87±5.49  & 85.24±9.82  & 89.69±4.23  & 94.68±3.37  & 84.60±8.93  & \cellcolor[RGB]{251, 228, 213}95.42±\textbf{2.72}  &   94.05±2.63    &  70.65±13.08     &  78.79±3.72     & 80.81±3.54  & 93.34±5.77       \\
\multicolumn{1}{c|}{2}                          &    \cellcolor[RGB]{73, 182, 120}                    & 88.81±3.65  & 95.45±2.50  & 96.12±2.24  & 97.00±2.00  & 95.67±3.73  & 98.74±1.34 &   99.85±0.19    &   94.65±9.43    &   97.39±1.30    &   96.32±1.33    & \cellcolor[RGB]{251, 228, 213}99.08±\textbf{0.77}       \\
\multicolumn{1}{c|}{3}                          &  \cellcolor[RGB]{255, 153, 56}                      & 45.15±12.12 & 24.92±16.98 & 54.83±17.92 & 70.43±14.32 & 52.12±20.72 & 48.06±24.63 &   68.53±5.94    &    89.81±8.36   &    70.97±4.50   &  68.61±11.68   & \cellcolor[RGB]{251, 228, 213}81.98±\textbf{11.09}      \\
\multicolumn{1}{c|}{4}                          &   \cellcolor[RGB]{49, 107, 185}                     & 55.57±9.91  & 82.58±11.98 & 81.44±5.38  & 92.48±\textbf{4.32}  & 83.54±4.84  & \cellcolor[RGB]{251, 228, 213}94.06±10.95 &   77.87±5.13    &   74.51±7.04    &    74.99±11.40   &   89.85±5.72   & 91.07±4.52       \\
\multicolumn{1}{c|}{5}                          &     \cellcolor[RGB]{255, 54, 31}                   & 48.59±20.42 & 99.66±0.34  & 96.85±5.30  & \cellcolor[RGB]{251, 228, 213}99.85±0.31  & 97.03±2.15  & 82.57±11.42 &   100.00±0.00    &    99.72±0.60   &    93.25±5.81   &    98.81±0.78    & 99.80±\textbf{0.16}       \\
\multicolumn{1}{c|}{6}                          &   \cellcolor[RGB]{91, 66, 162}                     & 48.42±8.54  & 31.76±11.02 & 81.90±10.03 & 82.74±7.60  & 40.50±3.40  & 61.58±7.32  & 94.63±3.13      &  96.35±3.22    &   99.05±1.03    &   63.10±10.10   & \cellcolor[RGB]{251, 228, 213}94.61±\textbf{2.69}       \\
\multicolumn{1}{c|}{7}                          &    \cellcolor[RGB]{130, 67, 36}                    & 56.37±18.04 & 55.45±21.79 & 86.00±12.26 & 91.09±6.76  & 58.71±41.48 & 24.10±10.72 &  99.03±1.18     & 93.79±5.96      &   96.01±0.80    &  59.23±9.95    & \cellcolor[RGB]{251, 228, 213}91.43±\textbf{4.88}       \\
\multicolumn{1}{c|}{8}                          &   \cellcolor[RGB]{180, 190, 190}                     & 68.36±10.67 & 77.49±11.06 & 87.27±7.60  & \cellcolor[RGB]{251, 228, 213}93.43±\textbf{6.25}  & 76.38±21.33 & 85.18±16.16 &   92.93±2.96  &   66.66±12.26    &   80.22±4.36    &    64.64±8.67   & 90.48±7.76       \\
\multicolumn{1}{c|}{9}                          &      \cellcolor[RGB]{102, 211, 216}                   & 33.31±20.07 & 93.53±4.17  & 97.86±1.37  & 94.28±3.12  & \cellcolor[RGB]{251, 228, 213}99.79±\textbf{0.17}  & 61.20±38.41 &    88.29±4.07   &    84.35±9.61   &    52.86±12.78   & 90.09±4.38   & 97.84±1.23       \\ \hline \hline
\multicolumn{2}{c|}{OA(\%)}                                              & 72.18±2.68  & 79.29±1.80  & 89.36±1.99  & 92.88±1.46  & 81.78±2.38  & 86.20±1.97  &   94.32±0.74    &   86.94±5.07    &    89.16±0.49   &   84.40±1.59   & \cellcolor[RGB]{251, 228, 213}95.09±\textbf{1.02}       \\
\multicolumn{2}{c|}{AA(\%)}                                              & 57.71±3.48  & 71.79±9.96  & 85.78±3.22  & 90.66±\textbf{1.44}  & 76.48±2.41  & 72.36±4.61  &  90.57±1.40     &  85.61±2.67     &    82.61±2.28   &  79.06±2.33   & \cellcolor[RGB]{251, 228, 213}93.30±4.32       \\
\multicolumn{2}{c|}{\(\kappa\)(\%)}                                           & 62.40±3.50  & 71.71±2.45  & 85.87±2.65  & 90.54±1.95  & 75.30±3.08  & 81.31±2.65  &  92.42±1.01     &  82.99±6.13     &   85.68±0.67    &  78.89±2.08   & \cellcolor[RGB]{251, 228, 213}93.65±\textbf{1.32}       \\ \hline
\end{tabular}
\end{threeparttable}
}
\label{exp_PU}
\end{table*}

% \usepackage{multirow}
% \usepackage{multirow}
\begin{table*}[]
\centering
\caption{{QUANTITATIVE PERFORMANCE OF DIFFERENT CLASSIFICATION METHODS IN TERMS OF OA, AA, \(\kappa\), AS WELL AS THE ACCURACIES FOR EACH CLASS ON THE Houston DATASET. THE BEST RESULTS ARE COLORED SHADOW AND OPTIMAL STANDARD DEVIATION IN BOLD}}
\resizebox{\textwidth}{!}{
\begin{threeparttable}
\begin{tabular}{c|c|cc|cc|cc| cccc c}
\hline
\multirow{2}{*}{\makecell[c]{Class \\ No.}} &   
\multirow{2}{*}{Color} & 
\multicolumn{2}{c|}{Conventional Classifier} &
\multicolumn{2}{c|}{CNNs Based Networks} &
\multicolumn{2}{c|}{Semi-supervised Based Networks} &
\multicolumn{3}{c|}{Diffusion Model Based Networks} &
\multicolumn{2}{c}{Languange-informed Networks}                                    \\ \cline{3-13}
\multicolumn{1}{c|}{}                           & \multicolumn{1}{c|}{}                        
& SVM         & 2D-CNN      & SSRN        & SS-ConvNeXt & DFRes-CR    & RSEN      &  DKDMN   & DEMAE   &  DiffusionAAE   & LDGNet    & \textbf{Txt2HSI-LDM(VAE)} \\ \hline \hline
\multicolumn{1}{c|}{1}                          &    \cellcolor[RGB]{147,75,67}                  & 84.39±14.81  & 79.46±13.27   & 79.45±22.99  & 65.22±7.14  & \cellcolor[RGB]{251, 228, 213}89.03±\textbf{0.83}  & 60.83±9.55  &   74.13±14.05    &  75.43±13.52     &  67.35±18.20     & 59.46±16.08  & 93.83±4.64       \\
\multicolumn{1}{c|}{2}                          &    \cellcolor[RGB]{241,215,126}                    & 58.60±20.56  & 58.82±16.97 &65.51±9.92     & 82.40±10.75 & 88.10±\textbf{1.25}  & 39.39±9.16 &   48.27±4.66    &   46.62±14.00    &   28.33±9.00   &   \cellcolor[RGB]{251, 228, 213}90.56±2.75    & 62.56±11.52       \\
\multicolumn{1}{c|}{3}                          &  \cellcolor[RGB]{147,148,231}                      & 98.74±1.29 & 96.10±4.42 & 92.38±4.35  & 94.89±2.65 & \cellcolor[RGB]{251, 228, 213}100.00±\textbf{0.00} & 62.15±16.96 &   96.87±1.86    &    99.98±0.04   &    73.30±13.89   &  24.93±43.19   & 99.19±0.45      \\
\multicolumn{1}{c|}{4}                          &   \cellcolor[RGB]{177,206,70}                     & 65.90±9.49  & 79.13±7.43  & \cellcolor[RGB]{251, 228, 213}87.01±7.99 & 77.57±16.71  & 97.69±\textbf{0.36}  & 85.26±6.01 &   81.17±1.94    &   86.21±8.58    &    72.77±11.78   &   75.62±7.83  & 84.61±6.71       \\
\multicolumn{1}{c|}{5}                          &     \cellcolor[RGB]{95,151,210}                   & 51.08±9.88 &  45.91±13.18 & 39.61±9.86 & 56.87±10.01  & 84.28±\textbf{0.38}  & 18.37±5.88 &   \cellcolor[RGB]{251, 228, 213}69.13±4.81    &    66.21±10.03   &    38.45±6.65   &    11.42±9.55    & 60.96±5.22      \\
\multicolumn{1}{c|}{6}                          &   \cellcolor[RGB]{238,122,109}                     & 56.94±\textbf{3.59}  & 52.54±14.01 & 70.84±9.38  & 71.75±18.97  & 87.16±13.99  & 84.48±8.55  & 65.90±18.16      &  69.57±20.00    &   \cellcolor[RGB]{251, 228, 213}86.79±7.71   &   7.27±5.72   & 73.81±5.76       \\
\multicolumn{1}{c|}{7}                          &    \cellcolor[RGB]{98,225,151}                    & 70.19±20.47 & 96.69±1.92 & 55.15±22.66  & 98.92±1.01  & \cellcolor[RGB]{251, 228, 213}100.00±\textbf{0.00} & 45.96±9.82 &  \cellcolor[RGB]{251, 228, 213}100.00±\textbf{0.00}     & 86.00±17.38      &   53.76±11.37    &  0.00±0.00    & \cellcolor[RGB]{251, 228, 213}100.00±\textbf{0.00}       \\
\multicolumn{1}{c|}{8}                          &   \cellcolor[RGB]{157,195,231}                     & 55.01±6.26 & 19.73±10.77  & 35.90±17.50  & 37.40±14.16  & \cellcolor[RGB]{251, 228, 213}80.70±3.50 & 42.65±5.92 &   69.00±10.09  &   47.35±15.52   &   44.16±6.39   &    74.76±\textbf{2.35}   & 66.71±11.22       \\
\multicolumn{1}{c|}{9}                          &   \cellcolor[RGB]{255,190,122}                     & 46.66±8.22 & 83.57±4.17 & 84.88±3.86   & 37.40±14.16  & 42.97±7.42 & 94.54±4.12 &   89.16±3.01  &   \cellcolor[RGB]{251, 228, 213}93.32±2.60    &   83.49±3.95    &    79.403±5.05   & 83.46±\textbf{2.08}       \\
\multicolumn{1}{c|}{10}                          &   \cellcolor[RGB]{184,131,211}                     & 26.31±6.49 & 19.75±10.73 & 23.36±6.01   & 88.84±2.05  & 28.02±7.50 & 1.4±\textbf{0.67} &   31.13±9.53  &   26.18±7.62    &   19.99±9.32    &    \cellcolor[RGB]{251, 228, 213}36.88±6.54   & 15.53±3.73       \\

\multicolumn{1}{c|}{11}                          &   \cellcolor[RGB]{69,139,0}                     & 12.80±5.10 & 8.19±3.12 & 7.64±2.07  & 27.05±14.55  & 34.68±13.26 & 2.91±\textbf{0.60} &   16.87±1.10  &   \cellcolor[RGB]{251, 228, 213}66.66±12.26    &   10.59±4.05    &    26.79±4.68   & 43.70±3.79       \\

\multicolumn{1}{c|}{12}                          &   \cellcolor[RGB]{127,127,127}                     & 26.31±6.49 & 12.94±8.25 & 13.68±6.16  & 13.72±5.43  & 40.21±6.59 & 3.54±\textbf{1.05} &   31.82±10.4  &   26.89±10.13    &   14.11±3.14    &    0.00±0.00  & \cellcolor[RGB]{251, 228, 213}41.87±4.07       \\

\multicolumn{1}{c|}{13}                          &   \cellcolor[RGB]{139,131,120}                     & 19.97±8.28 & 33.64±9.72 & 33.24±5.35  & 5.50±4.17  & 48.72±5.36 & 7.61±\textbf{1.21} &   37.43±8.07  &  23.10±7.86    &   27.47±4.57   &    \cellcolor[RGB]{251, 228, 213}53.46±8.31   & 42.32±5.95       \\

\multicolumn{1}{c|}{14}                          &   \cellcolor[RGB]{75,54,33}                     & 62.78±13.46 &40.91±11.59 &  63.25±9.25   & 53.82±15.60  & 94.44±\textbf{1.34} & 68.51±9.49 &   70.51±9.91  &   76.48±10.51    &   69.08±14.40    &    6.37±3.92   & \cellcolor[RGB]{251, 228, 213}76.50±9.58       \\

\multicolumn{1}{c|}{15}                          &   \cellcolor[RGB]{62,180,137}                     & 69.48±11.12 & 40.19±9.69 & 58.00±15.12   & 61.81±13.31  & \cellcolor[RGB]{251, 228, 213}99.59±\textbf{0.20} & 96.47±0.68 &   69.98±15.07  &   76.76±15.81    &   76.43±18.61    &    47.16±28.11  & 90.54±5.75       \\

\multicolumn{1}{c|}{16}                          &   \cellcolor[RGB]{128,128,0}                     & 51.66±18.09 & 39.08±18.33 & 38.78±9.32  & 66.01±18.16  & \cellcolor[RGB]{251, 228, 213}84.81±8.57 & 43.31±\textbf{4.01} &   60.09±9.96  &   60.63±17.12    &   42.56±7.96    &    41.04±22.84   & 72.19±15.49       \\

\multicolumn{1}{c|}{17}                          &   \cellcolor[RGB]{105,53,156}                     & 96.96±2.51 & 98.60±1.92 & 77.76±18.65  & 67.02±11.48  &\cellcolor[RGB]{251, 228, 213}100.00±\textbf{0.00} & 37.60±5.29 &   \cellcolor[RGB]{251, 228, 213}100.00±\textbf{0.00}  &   97.73±6.55    &   92.05±14.20   &    0.00±0.00 & \cellcolor[RGB]{251, 228, 213}100.00±\textbf{0.00 }     \\

\multicolumn{1}{c|}{18}                          &   \cellcolor[RGB]{44,22,8}                     & 23.78±12.60 & 36.86±14.03 & 15.42±4.01   & \cellcolor[RGB]{251, 228, 213}98.18±\textbf{2.98}  & 88.04±3.24 & 34.59±4.09 &   49.52±20.58  &   66.01±20.46    &    54.12±7.93   &    11.48±9.28   & 72.55±6.16       \\

\multicolumn{1}{c|}{19}                          &   \cellcolor[RGB]{210,180,140}                     & 35.49±14.74 & 45.57±17.17 & 41.01±10.74   & 16.85±10.84  & \cellcolor[RGB]{251, 228, 213}88.38±7.33 & 48.44±5.45 &   88.31±\textbf{0.24}  &   72.22±14.85    &   68.86±19.82    &   9.88±6.03   & 76.96±8.33       \\

\multicolumn{1}{c|}{20}                          &      \cellcolor[RGB]{244,154,194}                   & 67.76±13.49 & 50.16±19.64    & 55.57±11.78 & 45.61±18.58  & \cellcolor[RGB]{251, 228, 213}98.73±\textbf{0.84}  & 61.20±38.41 &    85.85±6.48   &    85.42±12.32   &    73.10±7.52  & 1.55±0.16   & 89.53±7.16       \\ \hline \hline
\multicolumn{2}{c|}{OA(\%)}                                              & 43.54±3.69  & 56.55±2.91  & 59.82±0.71   & 66.17±1.65  & 55.21±5.01  & 58.16±2.33  &   67.11±1.19 &   65.80±1.59    &    57.77±2.85  &   62.27±1.48   & \cellcolor[RGB]{251, 228, 213}67.74±\textbf{1.12}       \\
\multicolumn{2}{c|}{AA(\%)}                                              & 54.03±1.96  & 51.89±10.52  & 52.02±10.35   & 60.51±10.52  & \cellcolor[RGB]{251, 228, 213}78.77±4.23  & 72.36±4.61  &  66.76±1.07     &  65.00±3.05     &    54.84±3.61  &  32.90±21.18   & 71.39±\textbf{1.50}       \\
\multicolumn{2}{c|}{\(\kappa\)(\%)}                                           & 35.12±3.03  & 43.66±3.83  & 47.44±0.60   & 55.79±2.70  & 49.13±4.83  & 41.08±4.23  &  57.78±0.99     &  55.10±2.66     &   46.96±3.26    &  32.90±4.05   & \cellcolor[RGB]{251, 228, 213}58.26±\textbf{1.57}       \\ \hline
\end{tabular}
\end{threeparttable}
}
\label{exp_Houston}
\end{table*}

{Quantitative assessments of classification performance are presented in \autoref{exp_IN} \autoref{exp_PU}, and \autoref{exp_Houston}.} The best classification values are colored in shadow and optimal standard deviation is displayed in bold. Our proposed Txt2HSI-LDM(VAE) expands the training data by generating samples via language-informed conditional diffusion model to solve ISSD in HSIC. The results show that the proposed Txt2HSI-LDM(VAE) is superior to all other techniques in terms of OA, AA, \(\kappa\), and represents better performance on classwise accuracy simultaneously.

\subsubsection{\textbf{Performance on Indian Pines}}
\autoref{exp_IN} shows the quantitative comparison results of the Indian Pines dataset. Benefiting from the realistic and diverse generated samples by Txt2HSI-LDM(VAE), the performance of our proposed method is higher than other conventional classifiers, CNNs backbone, semi-supervised networks. For instance, Txt2HSI-LDM(VAE) is 35.67\%, 23.82\%, 3.84\%, 5.3\%, 20.47\% better than 2D-CNN, SSRN, SS-ConvNeXt, DFRes-CR, RSEN in terms of OA. The results show that Txt2HSI-LDM(VAE) can obtain label domain classification results with higher precision masks that have better raster reversibility under very limited labeled data. In terms of AA, The proposed Txt2HSI-LDM(VAE) exhibits a performance that is superior to DFRes-CR, with an increase of approximately 10.12\%. For small category, the proposed Txt2HSI-LDM(VAE) outperforms the semi-supervised RSEN by 14.95\%, 0.58\%, 13.45\%, 27.94\% on class 1, 7, 9, and 16. DFRes-CR only considers the conditional random field model in local domain, and the pseudo label generated by DFRes-CR relies heavily on baseline performance. Although RSEN performs data reduction before performing classification task contributing to less training time, but the information loss is also an issue worth considering. After expanding the sample according to Sample Balance Rate \(\lambda\), there is little fluctuation in accuracy between the 16 categories. The maximum fluctuation amplitude has been reduced from 35.76\% (class 7 and class 15) in the baseline model SS-ConvNeXt to 25.73\% (class 6 and class 5). While, the fluctuation amplitude of SVM, 2D-CNN, SSRN, DFRes-CR, RSEN is 77.63\%, 74.05\%, 70.93\%, 63.43\%, 88.17\%, respectively. So, the language-informed diffusion model can better capture the linguistic-visual similarity on pixel-level, generating more diverse samples to alleviate the ISSD problem in HSIC. {DKDMN use the contrastive learning method to enhance the discriminative capacility of model between diffusion knowledge and real samples, making this network achieves 87.60\% in terms of AA. Although DEMAE uses maksed aotoencoder to learn the spatial coherence, the diemnsion reduction PCA lead to information loss. DiffusionAAE is trained in GAN-style, which is time-consuming during model pretraining, see \autoref{model_size}, doesnt provide more model generalization ability, only achieves 70.51\% in terms of OA, 18.65\% lower than our model. LDGNet is proposed to solve the cross-scene hyperspectral image classification. In order to adpat to single scene, we test LDGNet on the same image. Although this model used language of two particle sizes of thickness, the unlabeled data haven't been fully used, leading to lower generalization, 39.23 lower than ours in terms of AA.}

\subsubsection{\textbf{Performance on Pavia University}}
\autoref{exp_PU} compares the experimental results to various classes on the Pavia University dataset. From \autoref{exp_PU}, Txt2HSI-LDM(VAE) achieves the best accuracy (OA of 95.09\%, AA of 93.30\%, and \(\kappa\) of 93.65\%) among the methods. The AA obtained by
Txt2HSI-LDM(VAE) on the Pavia University dataset outperforms SSRN, SS-ConvNeXt, DFRes-CR, RSEN by 7.52\%, 2.64\%, 16.82\% and 20.49\%, respectively. The maximum fluctuation amplitude has been reduced from 29.426\% (class 5 and class 2) in SS-ConvNeXt to 17.82\% (class 5 and class 3). While, the fluctuation amplitude of SVM, 2D-CNN, SSRN, DFRes-CR, RSEN is 55.5\%, 74.74\%, 43.03\%, 59.29\%, 74.64\%, respectively.

\subsubsection{\textbf{Performance on Houstom}}
{\autoref{exp_Houston} compares the experimental results to various classes on the Houston dataset. From \autoref{exp_Houston}, Txt2HSI-LDM(VAE) achieves the best accuracy (OA of 67.74\%, and \(\kappa\) of 58.26\%) among the methods. The OA obtained by
Txt2HSI-LDM(VAE) on the Houston dataset outperforms DKDMN, DEMAE, DiffusionAAE, LDGNet, by 0.63\%, 1.94\%, 9.97\% and 5.47\%, respectively.}

\begin{figure*}[htbp]
\centering
\includegraphics[width=\textwidth]{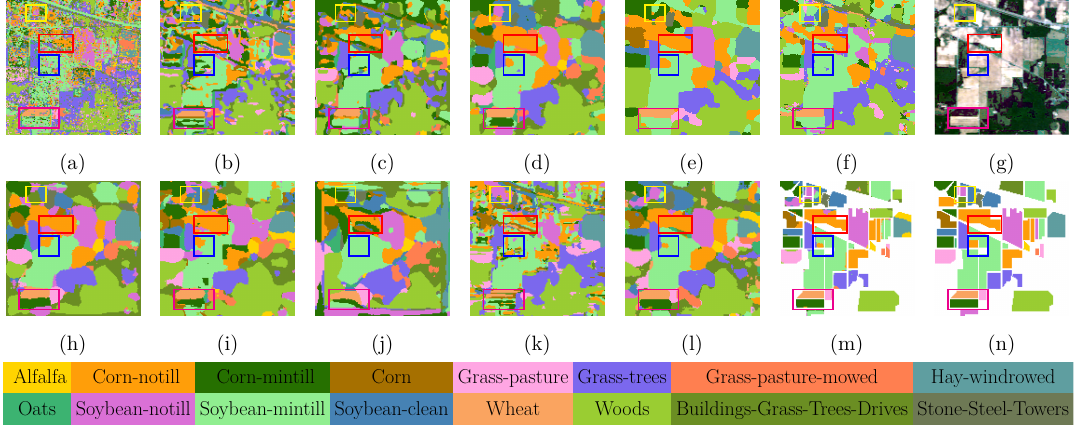}
\caption{{Classification maps of the Indian Pines dataset. (a) SVM (OA=52.08\%). (b) 2D-CNN (OA=53.49\%). (c) SSRN (OA=65.34\%). (d) SS-ConvNeXt (OA=85.32\%). (e) DFRes-CR (OA=83.86\%). (f) RSEN (OA=68.69\%). (g) RGB Composite Image. (h) DKDMN (OA=84.58\%). (i) DEMAE (OA=79.25\%). (j) DiffusionAAE (OA=70.51\%). (k) LDGNet (OA=61.34\%). (l) Txt2HSI-LDM(VAE) (OA=89.16\%). (m) Label-domain classification map of Txt2HSI-LDM(VAE). (n) Ground Truth Map. The box in red, blue, yellow, and magenta shows the region of interest, which represents the category boundary preserving capability of different methods.}}
\label{map_IN}
\end{figure*}

\begin{figure*}[htbp]
\centering
\includegraphics[width=\textwidth]{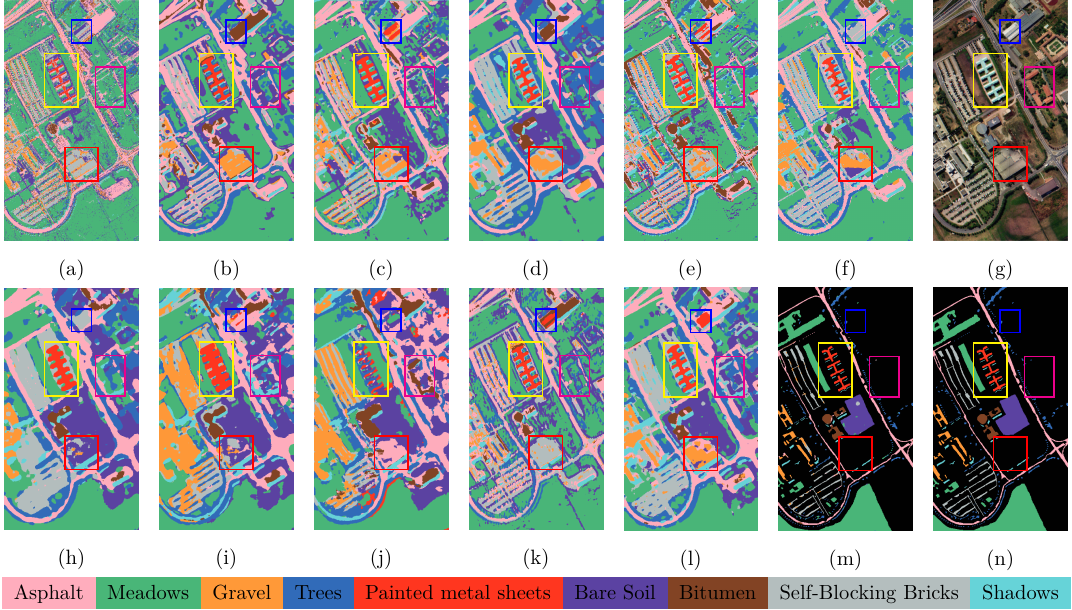}
\caption{{Classification maps of the Pavia University dataset. (a) SVM (OA=72.18\%). (b) 2D-CNN (OA=79.29\%). (c) SSRN (OA=89.36\%). (d) SS-ConvNeXt (OA=92.88\%). (e) DFRes-CR (OA=81.78\%). (f) RSEN (OA=86.20\%). (g) RGB Composite Image. (h) DKDMN (OA=94.32\%). (i) DEMAE (OA=86.94\%). (j) DiffusionAAE (OA=89.16\%). (k) LDGNet (OA=84.40\%). (l) Txt2HSI-LDM(VAE) (OA=95.09\%). (m) Label-domain classification map of Txt2HSI-LDM(VAE). (n) Ground Truth Map. The box in red, blue, yellow, and magenta shows the region of interest, which represents the category boundary preserving capability of different methods.}}
\label{map_PU}
\end{figure*}

\begin{figure*}[htbp]
\centering
\includegraphics[width=\textwidth]{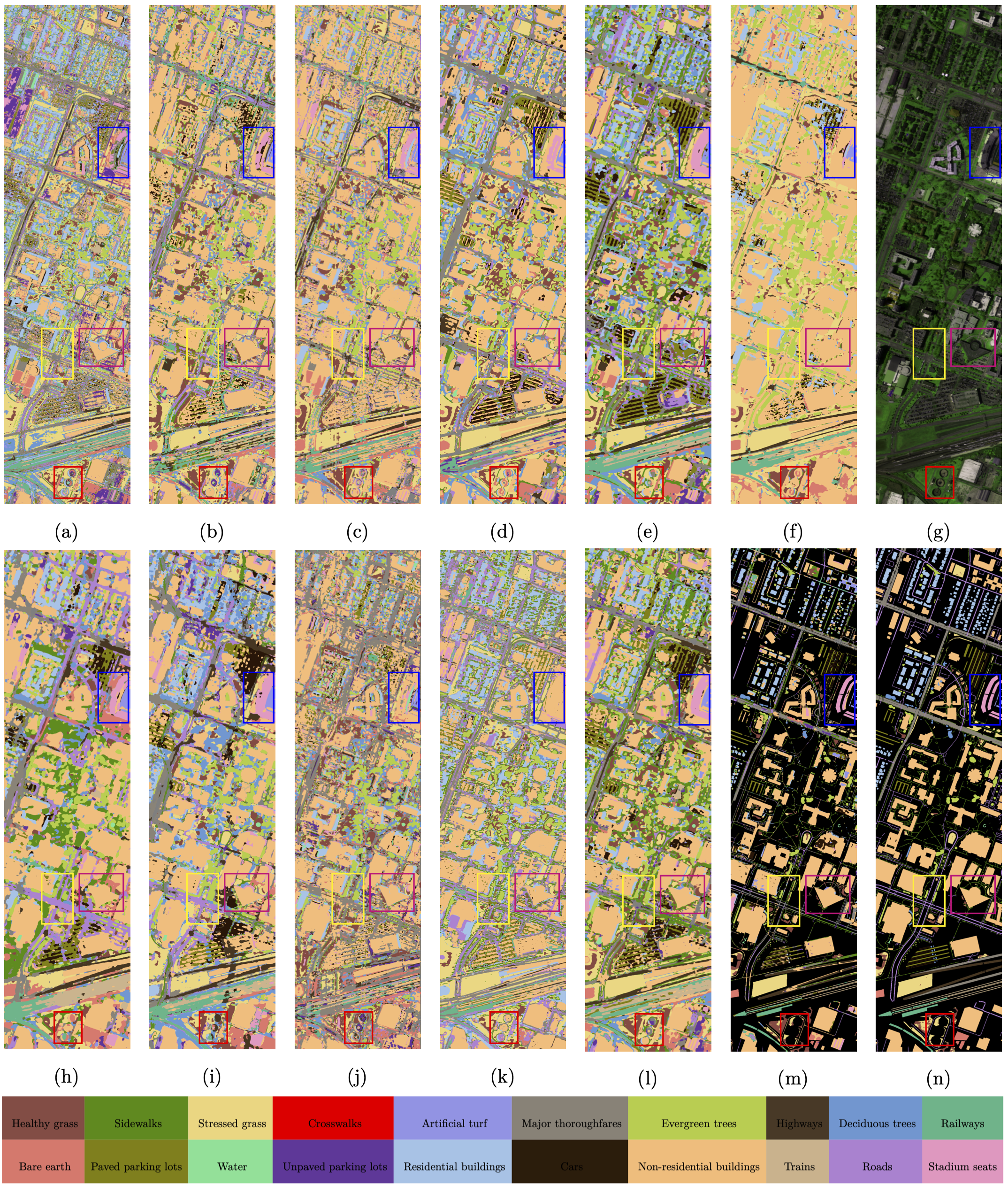}
\caption{{Classification maps of the Houston dataset. (a) SVM (OA=43.54\%). (b) 2D-CNN (OA=56.55\%). (c) SSRN (OA=59.82\%). (d) SS-ConvNeXt (OA=66.17\%). (e) DFRes-CR (OA=55.21\%). (f) RSEN (OA=58.16\%). (g) RGB Composite Image. (h) DKDMN (OA=67.11\%). (i) DEMAE (OA=65.80\%). (j) DiffusionAAE (OA=57.77\%). (k) LDGNet (OA=62.27\%). (l) Txt2HSI-LDM(VAE) (OA=67.74\%). (m) Label-domain classification map of Txt2HSI-LDM(VAE). (n) Ground Truth Map. The box in red, blue, yellow, and magenta shows the region of interest, which represents the category boundary preserving capability of different methods.}}
\label{map_Houston}
\end{figure*}
\begin{figure*}[htbp]
    \centering
    \includegraphics[width=\textwidth]{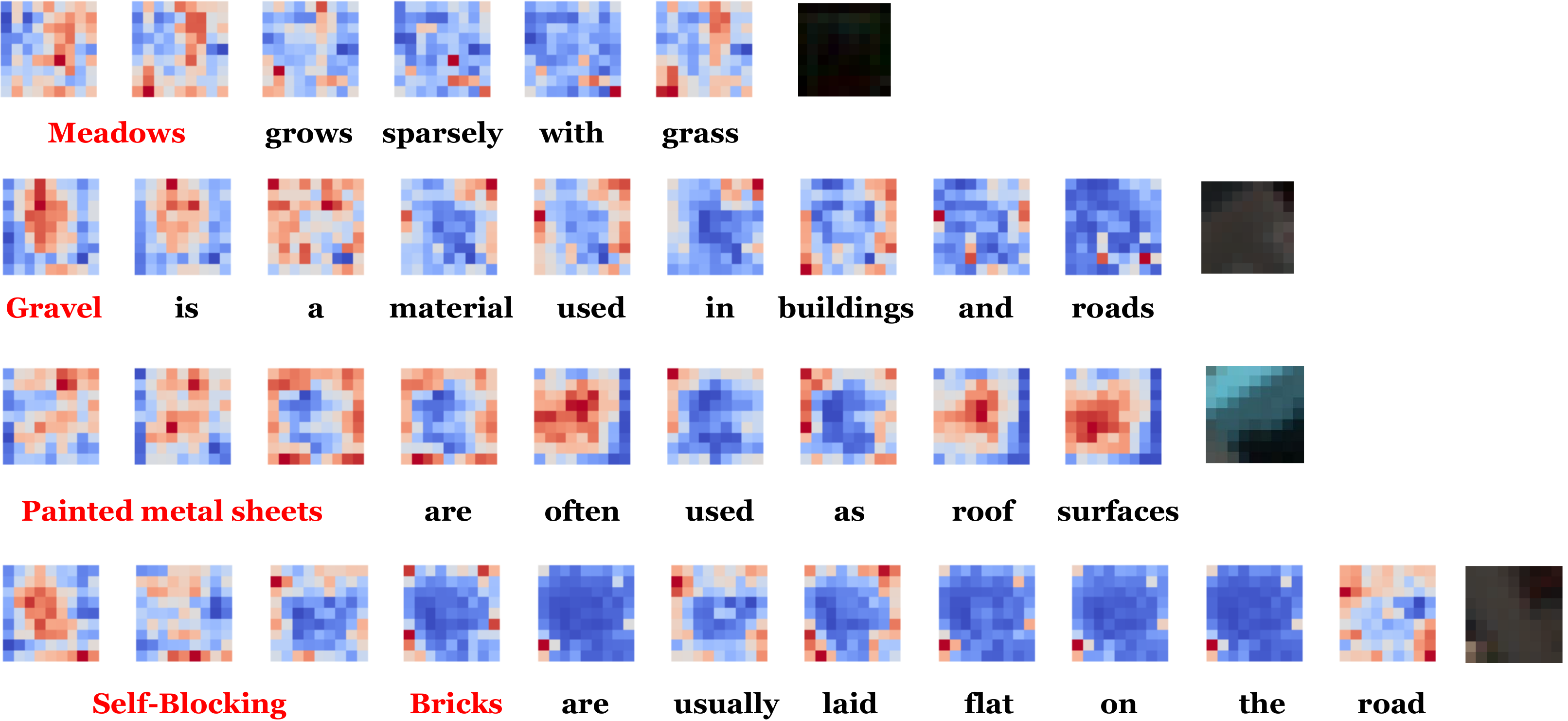}
    \caption{The plot displays Average attention masks for each word in the text which was used to synthesize the right image, showing the similarity of text-image features based on the cross-attention mechanism. The redder the color, the higher the similarity between language and spatial pixel. Take Pavia University data as an example. The category names are indicated in red font.}
    \label{Visual–Linguistic Cross Attention}
\end{figure*}
\subsection{Visual Evaluation on HSIC}
The classification maps are illustrated in \autoref{map_IN}, \autoref{map_PU}, and \autoref{map_Houston}. In \autoref{map_IN}, the proposed Txt2HSI-LDM(VAE) obtains less noisy and more accurate results in some areas of the classification maps, such as the category in blue block (Corn-notill) and Soybean-notill in yellow block. It is obvious that the boundary of land cover in the red block is better predicted, while other methods blur the boundaries, such as DKDMN, DEMAE. Some related results can also be seen in \autoref{map_PU}, such as the details in the yellow block. Due to the combination of conditional random fields and region growing, DFRes-CR can obtain high-confidence samples and get better classification maps in \autoref{map_IN} (e). {On more large scale dataset, Houston, our proposed models can well capture the building or artifical structure boundary (the blue, magenta color box). Overall, the classification map keep consistency wth the ground truth, representing the the scalability of Txt2HSI-LDM(VAE) on large size dataset is promissing, which can be explained by the uasge of unlabled data as well as the spatial mixing operation during model pre-training, i.e., LF-UE nad RPSC.}

\subsection{Visualization of Visual–Linguistic Cross Attention}

We return to our key observation: the spatial layout and geometry of the generated image depend on the cross-attention maps. The interaction between pixel and text is illustrated in \autoref{Visual–Linguistic Cross Attention}, where the average attention maps are plotted. Note that averaging is done for visualization purposes, and attention maps are kept separate for each head. As we can see, pixels show a stronger correlation with the words that describe them, e.g., pixels of the \textbf{Gravel} are correlated with the word "\textbf{Gravel}", as well as "Painted metal sheets". Additionally, some words related to the category are also captured, such as "\textbf{roof surface}", "\textbf{road}", while attention maps of prepositions and conjunctions, such as "\textbf{with}", "\textbf{on}", seem to fail to be captured. These words contain much information of spatial layout and topological relation. In the future, the ability to model prepositions and conjunctions needs to be further strengthened.

\subsection{{Computational Complexity}}

{To compare with the computational cost and model size of the diffusion training part, we set the batch size equal to 2 for Spectraldiff, DKDMN, DEMAE, DiffusionAAE, and our proposed method Txt2HSI-LDM(VAE) to calculate the number of parameters as well as the floating point operations (FLOPs). As we can see from \autoref{model_size}, spectraldiff has a huge amount of FLOPs, for example, with the shape of [2, 1, 200, 64, 64] as model input, the FLOPs will reach 1.557T, leading to slow model training and inference. During model training, spectraldiff will cost around 7 days to pretrain, which is much longer than Txt2HSI-LDM(VAE), so we exclude this method in comparative experiments and do not include the training time and generation speed here. DKDMN is also implemented based on 3D convolution, so the computational cost is also higher than Txt2HSI-LDM(VAE). Additionally, due to the PCA reduction, DEMAE reduces all four datasets into 36 dimensions, which hugely reduces the model parameter as well as FLOPs. Our model also has relatively lower FLOPs and model parameters. DiffusionAAE is trained in GAV-style, which needs more pretraining time than other models. 1s/16IMGs means that generating 16 images will cost 1 second.}

\begin{table}[htbp]
\centering
\caption{{Computational Cost and Model Size Across Different Architectures.}}
\resizebox{0.49\textwidth}{!}{
\begin{tabular}{c|c|cccccc}
\hline
\multirow{2}{*}{Dataset}          & \multirow{2}{*}{Metrics} & \multicolumn{6}{c}{Methods}                                                      \\ \cline{3-8} 
                                  &                          & Spectraldiff & DKDMN     & DEMAE    & DiffusionAAE & LDGNet   & Txt2HSI-LDM(VAE) \\ \hline
\multirow{4}{*}{Indian Pines}     & FLOPS                    & 1.55T        & 88.157G   & 68.595M  & 36.02G       & 4.538G   & 6.39G            \\
                                  & Params                   & 4.300M       & 4.300M    & 783.288K & 68.77M       & 36.68M   & 115.32M          \\
                                  & Training Time            & -            & 2s/epoch  & 1s/epoch & 1h/epoch     & 1s/epoch & 8s/epoch         \\
                                  & Generation Speed         & -            & -         & -        & 1s/35 IMGs   & -        & 1s/16 IMGs       \\ \hline
\multirow{4}{*}{Pavia University} & FLOPS                    & 809.604G     & 45.841G   & 68.595M  & 17.35G       & 4.456G   & 6.39G            \\
                                  & Params                   & 4.300M       & 4.300M    & 783.288K & 68.77M       & 33.87M   & 72.19M           \\
                                  & Training Time            & -            & 9s/epoch  & 1s/epoch & 1h/epoch     & 1s/epoch & 6s/epoch         \\
                                  & Generation Speed         & -            & -         & -        & 1s/35 IMGs   & -        & 1s/16 IMGs       \\ \hline
\multirow{4}{*}{Houston}          & FLOPS                    & 4.110T       & 21.143G   & 68.595M  & 10.78G       & 4.41G    & 6.39G            \\
                                  & Params                   & 4.300M       & 4.300M    & 783.288K & 68.77M       & 33.65M   & 115.33M          \\
                                  & Training Time            & -            & 17s/epoch & 1s/epoch & 2.4h/epoch   & 3s/epoch & 6s/epoch         \\
                                  & Generation Speed         & -            & -         & -        & 1s/35 IMGs   & -        & 1s/16 IMGs       \\ \hline
\end{tabular}
}
\label{model_size}
\end{table}

\section{Conclusion and Discussion} \label{conclusion}
The Txt2HSI-LDM(VAE), which uses condition of language to generate diverse samples for expanding training data, is proposed. The visual-linguistic alignment is established by the cross-attention mechanism. For better learning the spatial-spectral information, we design a semi-supervised conditional diffusion model framework to realize the information interaction between limited labeled data and a bunch of unlabeled data via knowledge distillation.
Experiments show that whether it is a fine-grained text or a coarse-grained text, it will both promote the classification performance. The spatial-level attention maps show that cross attention can capture the spatial layout and location of a specific category.

While in this work, the text description is designed by hand according to prior knowledge or based on templates. It may be possible to build a specialized text library using ChatGPT, DeepSeek, or other models to realize it, even a large model in the remote sensing field. Additionally, prepositions and conjunctions in text description are especially important to establish reasonable topological relationships of land cover, which need further research. {Also, in the cross-attention module part, how to emphasize the importance of prepositions and conjunctions, which will affect the geometric relationship modeling. The most relevant descriptions and significant content shouldn't be submerged in redundant text when building the linguistic descriptions \cite{ye2025differential}. Moreover, the VAE training objective function is often on the pixel level, lacking constraints on the objective level and explicit geometric consistency. How to avoid or reduce the suppression of high-frequency information, like the boundary and linear feature, also deserves to be studied.}

\bibliographystyle{IEEEtran}
\bibliography{IEEEabrv,ref}
\end{document}